%% file: archive.tex
\documentclass[10pt,twocolumn,letterpaper]{article}

\usepackage{cvpr}
\usepackage{times}
\usepackage{epsfig}
\usepackage{graphicx}
\usepackage{amsmath}
\usepackage{amssymb}

\usepackage{subcaption}
\usepackage{enumerate}
\usepackage[shortlabels]{enumitem}
\usepackage{textcomp}
\usepackage{gensymb}
\usepackage{listings}
\usepackage{xcolor}
\usepackage{wrapfig}

\setlist[itemize]{leftmargin=*,nosep,nolistsep}

\usepackage[pagebackref=true,breaklinks=true,colorlinks,bookmarks=false]{hyperref}

\cvprfinalcopy 

\def\cvprPaperID{3386} 

\graphicspath{{figs/}} 

\newcommand{\R}{\mathbb{R}}
\newcommand{\yh}{\widehat{y}}
\newcommand{\ch}{\widehat{c}}
\newcommand{\twodots}{\mathinner {\ldotp \ldotp}}
\newcommand{\epsi}{\varepsilon}
\newcommand{\p}{\mathbf{p}}
\newcommand{\e}{\mathbf{e}}

\newcommand{\savefootnote}[2]{\footnote{\label{#1}#2}}
\newcommand{\repeatfootnote}[1]{\textsuperscript{\ref{#1}}}

\newcommand{\beginsupplement}{%
        \setcounter{section}{0}
        \setcounter{table}{0}
        \renewcommand{\thetable}{S\arabic{table}}%
        \setcounter{figure}{0}
        \renewcommand{\thefigure}{S\arabic{figure}}%
     }

\begin{document}
	
	\title{Polygonal Building Extraction by Frame Field Learning}
	
	\author{Nicolas Girard$^1$ \quad Dmitriy Smirnov$^2$ \quad Justin Solomon$^2$ \quad Yuliya Tarabalka$^3$\\
	$^1$Universit\'e C\^{o}te d'Azur, Inria \quad $^2$Massachusetts Institute of Technology \quad $^3$LuxCarta Technology
 	}
	
	\maketitle
	
	\addtocontents{toc}{\protect\setcounter{tocdepth}{0}}
	\input{main_content}
	
	{\small
		\bibliographystyle{ieee_fullname}
		\bibliography{biblio}
	}
	
	\newpage
    \section*{
        \centering
        Supplementary Materials
    }
    \beginsupplement
    
    \addtocontents{toc}{\protect\setcounter{tocdepth}{2}}
    \input{supp_mat_content}
	
\end{document}

%% file: main_content.tex
\begin{abstract}
	While state of the art image segmentation models typically output segmentations in raster format, applications in geographic information systems often require vector polygons. To help bridge the gap between deep network output and the format used in downstream tasks, we add a frame field output to a deep segmentation model for extracting buildings from remote sensing images. We train a deep neural network that aligns a predicted frame field to ground truth contours. This additional objective improves segmentation quality by leveraging multi-task learning and provides structural information that later facilitates polygonization;  we also introduce a polygonization algorithm that utilizes the frame field along with the raster segmentation. Our code is available at \url{https://github.com/Lydorn/Polygonization-by-Frame-Field-Learning}. 
\end{abstract}

\section{Introduction}

\begin{wrapfigure}{R}{0.4\linewidth}
	\vspace{-1em}
	\centering
	\includegraphics[width=\linewidth]{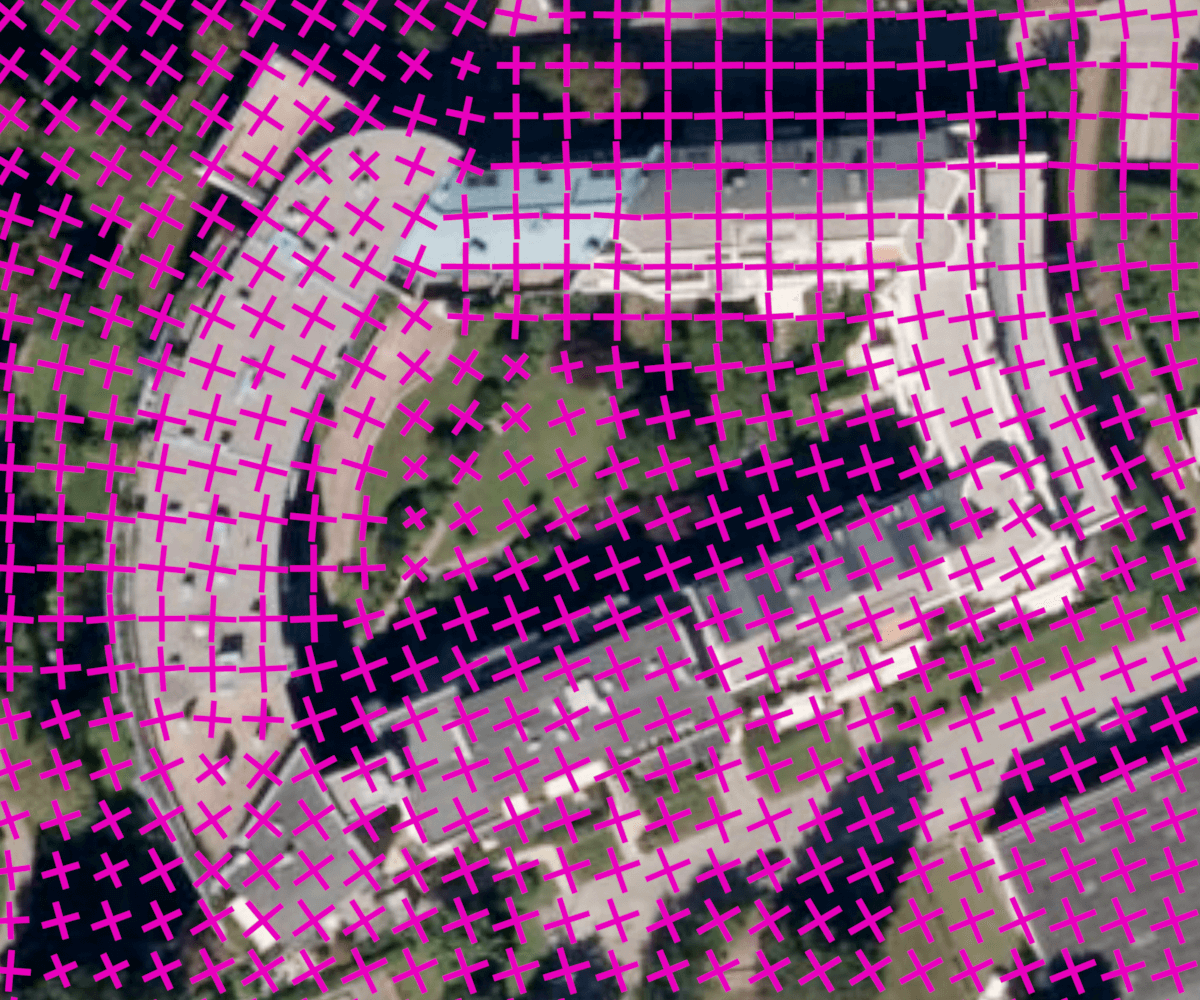}
	\vspace{-1.5em}
	\caption[Frame field overlaid on sample image.]{A frame field output by our network.}
	\vspace{-1em}
	\label{fig:frame_field_sample}
\end{wrapfigure}

Due to their success in processing large collections of noisy images, deep convolutional neural networks (CNNs) have achieved state-of-the-art in remote sensing segmentation. Geographic information systems like Open Street Map (OSM)~\cite{osm}, however, require segmentation data in \emph{vector} format (e.g., polygons and curves) rather than raster format, which is generated by segmentation networks. Additionally, methods that extract objects from remote sensing images require especially high throughput to handle the volume of high-resolution aerial images captured daily over large territories of land. Thus, modifications to the conventional CNN pipeline are necessary.

Existing work on deep building segmentation generally falls into one of two general categories. The first vectorizes the probability map produced by a network a posteriori, e.g., by using contour detection (marching squares~\cite{marching_cubes}) followed by polygon simplification (Ramer–Douglas–Peucker~\cite{ramer-d-p, r-douglas-peucker}). Such approaches suffer when the classification maps contain imperfections such as smoothed out corners, a common artifact of conventional deep segmentation methods. Moreover, as we show in Fig.~\ref{fig:frame_field_motivation}, even perfect probability maps are challenging to polygonize due to shape information being lost from the discretization of the raster output. To improve the final polygons, these methods employ expensive and complex post-processing procedures. ASIP polygonization~\cite{cvpr2020li} uses polygonal partition refinement to approximate shapes from the output probability map based on a tunable parameter controlling the trade-off between complexity and fidelity. In~\cite{building_reg_gan}, a decoder and a discriminator  regularize output probability maps adversarially. This requires computing large matrices of pairwise discontinuity costs between pixels and involves adversarial training, which is less stable than conventional supervised learning. 

Another category of deep segmentation methods learns a vector representation directly. For example, Curve-GCN~\cite{CurveGCN2019} trains a graph convolutional network (GCN) to deform polygons iteratively, and PolyMapper~\cite{PolyMapper} uses a recurrent neural network (RNN) to predict vertices one at a time. While these approaches directly predict polygon parameters, GCNs and RNNs suffer from several disadvantages. Not only are they more difficult to train than CNNs, but also their output topology is restricted to simple polygons without holes---a serious limitation in segmenting complex buildings. Additionally, adjoining buildings with common walls are common, especially in city centers. Curve-GCN and PolyMapper are unable to reuse the same polyline in adjoining buildings, yielding overlaps and gaps.

We introduce a building extraction algorithm that avoids the challenges above by adding a frame field output to a fully-convolutional network (see Fig.~\ref{fig:frame_field_sample}). While this has imperceptible effect on training or inference time, the frame field not only increases segmentation performance, e.g., yielding sharper corners, but also provides useful information for vectorization. Additional losses learn a valid frame field that is consistent with the segmentation. These losses regularize the segmentation, similar to \cite{MengTang2018b}, which includes MRF/CRF regularization terms in the loss function to avoid extra MRF/CRF inference steps.

The frame field allows us to devise a straightforward polygonization method extending the Active Contours Model (ACM, or ``snakes'')~\cite{Kass88snakes:active}, which we call the Active Skeleton Model (ASM). Rather than fitting contours to image data, ASM fits a skeleton graph, where each edge connects two junction nodes with a chain of vertices (i.e., a polyline). This allows us to reuse shared walls between adjoining buildings. To our knowledge, no existing method handles this case (\cite{Tripodi2019} shows results with common walls but does not provide details). Our method naturally handles large buildings and buildings with inner holes, unlike end-to-end learning methods like PolyMapper~\cite{PolyMapper}. Lastly, our polygon extraction pipeline is highly GPU-parallelizable, making it faster than more complex methods. 

Our main contributions are:
\begin{enumerate}[(i),itemsep=-1pt,topsep=1pt]
	\item a learned frame field aligned to object tangents, which improves segmentation via multi-task learning;
	\item coupling losses between outputs for self-consistency, further leveraging multi-task learning; and
	\item a fast polygonization method leveraging the frame field, allowing complexity tuning of a corner-aware simplification step and handling non-trivial topology.
\end{enumerate}

\section{Related work}

ASIP polygonization~\cite{cvpr2020li} inputs an RGB image and a probability map of objects (e.g., buildings) detected in the image (e.g., by a neural network). Then, starting from a polygonal partition that oversegments the image into convex cells, the algorithm refines the partition while labeling its cells by semantic class. The refinement process is an optimization with terms that balance fidelity to the input against complexity of the output polygons. The configuration space is explored by splitting and merging the polygonal cells. As the fidelity and complexity terms can be balanced with a coefficient, the fidelity-to-complexity ratio can be tuned. However, there does not exist a systematic approach for interpreting or determining this coefficient. While ASIP post-processes the output of a deep learning method, recent approaches aim for an end-to-end pipeline.

CNNs are successful at converting grid-based input to grid-based output for tasks where each output pixel depends on its local neighborhood in the input. In this setting, it is straightforward and efficient to train a network for supervised prediction of segmentation probability maps. The paragraphs below, however, detail major challenges when using such an approach to extract polygonal buildings.

First, the model needs to produce variable-sized outputs to capture varying numbers of objects, contours, and vertices. This requires complex architectures like recurrent neural networks (RNNs)~\cite{lstm}, which are not as efficiently trained as CNNs and need multiple iterations at inference time. Such is the case for PolyMapper~\cite{PolyMapper}, Polygon-RNN~\cite{PolygonRNN2017}, and Polygon-RNN++~\cite{PolygonRNN++2018}. Curve-GCN~\cite{CurveGCN2019} predicts a fixed number of vertices simultaneously.

A second challenge is that the model must make discrete decisions of whether to add a contour, whether to add a hole to an object, and with how many vertices to describe a contour. Adding a contour is solved by object detection: a contour is predicted for each detected object. Adding holes to an object is more challenging, but a few methods detect holes and predict their contours. One model, BSP-Net~\cite{chen2020bspnet}, circumvents this issue by combining predicted convex shapes for the final output, producing shapes in a compact format, with potential holes inside. To our knowledge, the number of vertices is not a variable that current deep learning models can optimize for; discrete decisions are difficult to pose differentiably without training techniques such as the straight-through estimator~\cite{bengio2013estimating} or reinforcement learning~\cite{Sutton20reinforcementlearning, mnih2015humanlevel, DBLP:journals/corr/MnihKSGAWR13}.

A third challenge is that, unlike probability maps, the output structure of polygonal building extraction is not grid-like. Within the network, the grid-like structure of the image input has to be transformed to a more general planar graph structure representing building outlines. City centers have the additional problem of adjoining buildings that share a wall. Ideally, the output geometry for such a case would be a collection of polygons, one for each individual building, which share polylines corresponding to common walls. Currently, no existing deep learning method tackles this case. Our method solves it but is not end-to-end. PolyMapper~\cite{PolyMapper} tackles the individual building and road network extraction tasks. As road networks are graphs, they propose a novel sequentialization method to reformulate graph structures as closed polygons. Their approach might work in the case of adjoining buildings with common walls. Their output structure, however, is less adapted to GPU computation, making it less efficient. RNNs such as PolyMapper~\cite{PolyMapper}, Polygon-RNN~\cite{PolygonRNN2017}, and Polygon-RNN++~\cite{PolygonRNN++2018} perform beam search at inference to prune off improbable sequences, which requires more vertex predictions than are used in the final output and is inefficient. The DefGrid~\cite{deformablegrid} module is a non-RNN approach where the network processes polygonal superpixels. It is more complex than our simple fully-convolutional network and is still subject to the rounded corner problem.

The challenges above demand a middle ground between learning a bitmap segmentation followed by a hand-crafted polygonization method and end-to-end methods, aiming to be easily-deployable, topologically flexible w.r.t.\ holes and common walls, and efficient. A step in this direction is the machine-learned building polygonization~\cite{zorzi2020machinelearned} that predicts building segmentations using a CNN, uses a generative adversarial network to regularize building boundaries, and learns a building corner probability map, from which vertices are extracted. In contrast, our model predicts a frame field both as additional geometric information (instead of a building corner probability map) and as a way to regularize building boundaries (instead of adversarial training). The addition of this frame field output is similar in spirit to DiResNet~\cite{Ding2020DiResNetDR}, a road extraction neural network that outputs road direction in addition to road segmentation, first introduced in~\cite{Batra_2019_CVPR}. The orientation is learned for each road pixel by a cross-entropy classification loss whose labels are orientation bins. This additional geometric feature learned by the network improves the overall geometric integrity of the extracted objects (in their case road connectivity). The differences to our method include the following:  (1) our frame fields encode two orientations instead of one (needed for corners), (2) we use a regression loss instead of a classification loss, and (3) we use coupling losses to promote coherence between segmentation and frame field.

\section{Method}

\begin{figure}[ht]
	\begin{subfigure}{0.24\linewidth}
		\includegraphics[width=\textwidth]{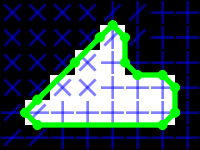}
		\caption{Iter. 0}
	\end{subfigure}
	\begin{subfigure}{0.24\linewidth}
		\includegraphics[width=\textwidth]{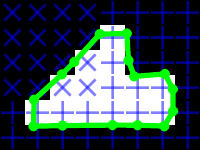}
		\caption{Iter. 50}
	\end{subfigure}
	\begin{subfigure}{0.24\linewidth}
		\includegraphics[width=\textwidth]{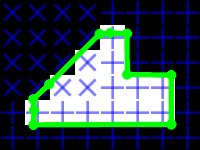}
		\caption{Iter. 250}
	\end{subfigure}
	\begin{subfigure}{0.24\linewidth}
		\includegraphics[width=\textwidth]{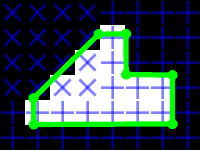}
		\caption{Result}
	\end{subfigure}
	\caption[An ambiguous segmentation map]{Even a perfect classification map can yield incorrect polygonization due to a locally ambiguous probability map, as shown in (a), the output of marching squares. Our polygonization method iteratively optimizes the contour (b-d) to align to a frame field, yielding better results as our frame field (blue) disambiguates between slanted walls and corners, preventing corners from being cut off.}
	\label{fig:frame_field_motivation}
\end{figure}

Our key idea is to help the polygonization method solve ambiguous cases caused by discrete probability maps by asking the neural network to output missing shape information in the form of a frame field (see Fig.~\ref{fig:frame_field_motivation}). This practically does not increase training and inference time, allows for simpler and faster polygonization, and regularizes the segmentation---solving the problem of small misalignments of the ground truth annotations that yield rounded corners if no regularization is used. 

\subsection{Frame fields}
\label{sec:frame_fields}

We provide the necessary background on frame fields, a key part of our method. Following \cite{vaxman2016directional,Diamanti2014}, a frame field is a \emph{4-PolyVector field}, which assigns four vectors to each point of the plane. In the case of a frame field, however, the first two vectors are constrained to be opposite to the other two, i.e., each point is assigned a set of vectors $\{u, -u, v, -v\}$.
At each point in the image, we consider the two directions that define the frame as two complex numbers $u, v \in \mathbb{C}$. We need two directions (rather than only one) because buildings, unlike organic shapes, are regular structures with sharp corners, and capturing directionality at these sharp corners requires two directions. To encode the directions in a way that is agnostic to relabeling and sign change, we represent them as coefficients of the following polynomial:
\begin{equation}
	\label{eq:frame_field_poly}
	f(z) = (z^2 - u^2)(z^2 - v^2) = z^4 + c_2 z^2 + c_0 \,.
\end{equation}
We denote \eqref{eq:frame_field_poly} above by $f(z;c_0,c_2)$. Given a $(c_0,c_2)$ pair, we can easily recover one pair of directions defining the corresponding frame:
{
	\footnotesize
	\begin{equation}
		\label{eq:frame_field_convert}
		\left\{\begin{aligned} 
			c_0 &= u^2v^2 \\
			c_2 &= -(u^2 + v^2)
		\end{aligned}\right. \iff 
		\left\{\begin{aligned}
			u^2 &= -\tfrac{1}{2}\left(c_2 + \sqrt{c_2^2 - 4c_0}\right)\\ 
			v^2 &= -\tfrac{1}{2}\left(c_2 - \sqrt{c_2^2 - 4c_0}\right) \,.
		\end{aligned}\right.
	\end{equation}
}

In our approach, inspired by \cite{Bessm2019}, we learn a smooth frame field with the property that, along building edges, at least one field direction is aligned to the polygon tangent direction. At polygon corners, the field aligns to \emph{both} tangent directions, motivating our use of PolyVector fields rather than vector fields. Away from polygon boundaries, the frame field does not have any alignment constraints but is encouraged to be smooth and not collapse to a line field. Like \cite{Bessm2019}, we formulate the field computation variationally, but, unlike their approach, we use a neural network to learn the field at each pixel, which is also explored in \cite{directional_field_learning}. To avoid sign and ordering ambiguity, we learn a $(c_0, c_2)$ pair per pixel rather than $(u, v)$.

\subsection{Frame field learning}

\begin{figure*}[ht]
	\centering
	\includegraphics[width=\textwidth]{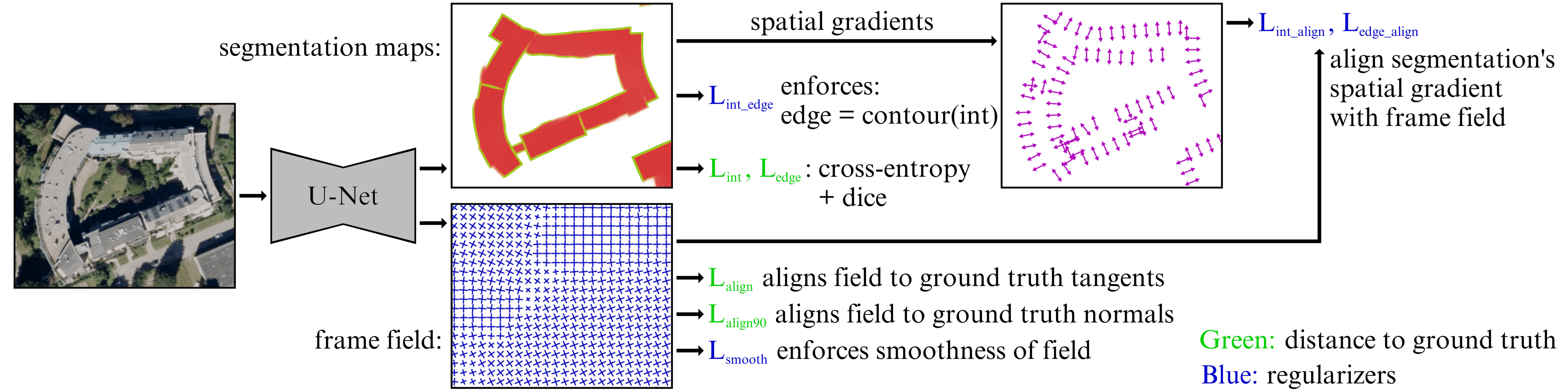}
	\caption[Frame field learning infographic.]{Given an overhead image, our model outputs an edge mask, interior mask, and frame field. The loss aligns the masks and field to ground truth data, enforces smoothness of the frame field, and ensures consistency between the outputs.}
	\label{fig:method_figure}
\end{figure*}

We describe our method, illustrated in Fig.~\ref{fig:method_figure}. Our network takes a $3\!\times\!H\!\times\!W$ image $I$ as input and outputs a pixel-wise classification map and a frame field. The classification map contains two channels, $\yh_\textit{int}$ corresponding to building interiors and $\yh_\textit{edge}$ to building boundaries. The frame field contains four channels corresponding to the two complex coefficients $\ch_0,\ch_2\in\mathbb{C}$, as in \S\ref{sec:frame_fields} above.

\paragraph{Segmentation losses.} Our method can be used with any deep segmentation model as a backbone; in our experiments, we use the U-Net~\cite{unet} and DeepLabV3 \cite{DeepLabV3} architectures. The backbone outputs an $F$-dimensional feature map $\yh_\textit{backbone} \in \R^{F \times H \times W}$. For the segmentation task, we append to the backbone a fully-convolutional block (taking $\yh_\textit{backbone}$ as input) consisting of a $3\!\times\!3$ convolutional layer, a batch normalization layer, an ELU nonlinearity, another $3\!\times\!3$ convolution, and a sigmoid nonlinearity. This segmentation head outputs a segmentation map $\yh_\textit{seg} \in \R^{2 \times H \times W}$. The first channel contains the object interior segmentation map $\yh_{\textit{int}}$ and the second contains the contour segmentation map $\yh_{\textit{edge}}$. Our training is supervised---each input image is labeled with ground truth $y_{\textit{int}}$ and $y_{\textit{edge}}$, corresponding to rasterized polygon interiors and edges, respectively. We then use a linear combination of the cross-entropy loss and Dice loss \cite{sudre2017generalised} for loss $L_{\textit{int}}$ applied on the interior output as well as loss $L_{\textit{edge}}$ applied on the contour (edge) output. 

\paragraph{Frame field losses.} In addition to the segmentation masks, our network outputs a frame field. We append another head to the backbone via a fully-convolutional block consisting of a $3\!\times\!3$ convolutional layer, a batch normalization layer, an ELU nonlinearity, another $3\!\times\!3$ convolution, and a tanh nonlinearity. This frame field block inputs the concatenation of the output features of the backbone and the segmentation output: $[\yh_\textit{backbone}, \yh_\textit{seg}] \in \R^{(F + 2) \times H \times W}$. It outputs the frame field with $\ch_0,\ch_2 \in \mathbb{C}^{H \times W}$. The corresponding ground truth label is an angle $\theta_\tau \in [0, \pi)$ of the unsigned tangent vector of the polygon contour. We use three losses to train the frame field:
\begin{align}
	L_{\textit{align}} &= \tfrac{1}{HW} \sum_{{x} \in I} y_{\textit{edge}}({x}) |f(e^{i\theta_{\tau}}; \ch_0({x}), \ch_2({x}))|^2, \label{alignloss}\\
	L_{\textit{align90}} &= \tfrac{1}{HW} \sum_{{x} \in I} y_{\textit{edge}}({x}) |f(e^{i\theta_{\tau^\bot}}; \ch_0({x}), \ch_2({x}))|^2,\\
	L_{\textit{smooth}} &= \tfrac{1}{HW} \sum_{{x} \in I} \left(\|\nabla \ch_0({x})\|^2 + \|\nabla \ch_2({x})\|^2\right),
\end{align}
where $\theta_w$ is the direction of $w$ ($w = \|w\|_2 e^{i\theta_w}$), and $\tau^\bot=\tau-\tfrac{\pi}2$. Each loss measures a different property of the field:
\begin{itemize}
	\item $L_{\textit{align}}$ enforces alignment of the frame field to the tangent directions. This term is small when the polynomial $f(\cdot;\ch_0,\ch_2)$ has a root near $e^{i\theta_\tau}$, implicitly implying that one of the field directions $\{\pm u, \pm v\}$ is aligned with the tangent direction $\tau$. Since \eqref{eq:frame_field_poly} has no odd-degree terms, this term has no dependence on the sign of $\tau$, as desired.
	\item $L_{\textit{align90}}$ prevents the frame field from collapsing to a line field by encouraging it to also align with $\tau^\bot$.
	\item $L_{\textit{smooth}}$ is a Dirichlet energy measuring the smoothness of $\ch_0({x})$ and $\ch_2({x})$ as functions of location ${x}$ in the image. Smoothly-varying $\ch_0$ and $\ch_2$ yield a smooth frame field.
\end{itemize}

\paragraph{Output coupling losses.} We add coupling losses to ensure mutual consistency between our network outputs:
{\allowdisplaybreaks
	\begin{align}
		L_{\textit{int align}} &= \frac{1}{HW} \sum_{{x} \in I} f(\nabla \yh_{\textit{int}}({x}); \ch_0({x}), \ch_2({x}))^2, \\
		L_{\textit{edge align}} &= \frac{1}{HW} \sum_{{x} \in I} f(\nabla \yh_{\textit{edge}}({x}); \ch_0({x}), \ch_2({x}))^2, \\
		\begin{split}
			L_{\textit{int edge}} &=
			\frac{1}{HW} \sum_{{x} \in I} \max\left(1 - \yh_{\textit{int}}({x}), \|\nabla \yh_{\textit{int}}({x})\|_2\right) \\
			&\cdot\left|\|\nabla \yh_{\textit{int}}({x})\|_2 - \yh_{\textit{edge}}({x})\right|. \end{split}
	\end{align}
}

\begin{itemize}
	\item $L_{\textit{int align}}$ aligns the spatial gradient of the predicted interior map $\yh_{\textit{int}}$ with the frame field (analogous to \eqref{alignloss}).
	\item $L_{\textit{edge align}}$ aligns the spatial gradient of the predicted edge map $\yh_{\textit{edge}}$ with the frame field (analogous to \eqref{alignloss}).
	\item $L_{\textit{int edge}}$ makes the predicted edge map be equal to the norm of the spatial gradient of the predicted interior map. This loss is applied outside of buildings (hence the $1 - \yh_{\textit{int}}({x})$ term) and along building contours (hence the $\|\nabla \yh_{\textit{int}}({x})\|_2$ term) and is not applied inside buildings, so that common walls between adjoining buildings can still be detected by the edge map.
\end{itemize}

\paragraph{Final loss.}
Because the losses ($L_{\textit{int}}$, $L_{\textit{edge}}$, $L_{\textit{align}}$, $L_{\textit{align90}}$, $L_{\textit{smooth}}$, $L_{\textit{int align}}$, $L_{\textit{edge align}}$, and $L_{\textit{int edge}}$) have distinct units, we compute a normalization coefficient for each loss by averaging its value over a random subset of the training dataset using a randomly-initialized network. Losses are then normalized by this coefficient before being linearly combined. This normalization aims to rescale losses such that they are easier to balance. More details are in the supplementary materials.

\subsection{Frame field polygonization}

\begin{figure*}[ht]
	\centering
	\includegraphics[width=\textwidth]{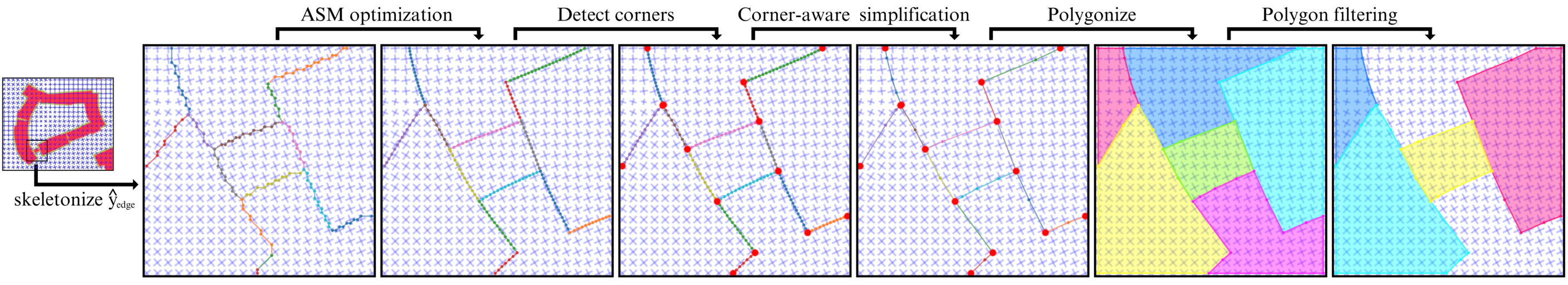}
	\caption{Overview of our post-processing polygonization algorithm. Given an interior classification map and frame field (Fig.~\ref{fig:method_figure}) as input, we optimize the contour to align to the frame field using an Active Skeleton Model (ASM) and detect corners using the frame field, simplifying non-corner vertices.}
	\label{fig:polygonization_method_figure}
\end{figure*}

The main steps of our polygonization method are shown in Fig.~\ref{fig:polygonization_method_figure}. It is inspired by the Active Contour Model (ACM)~\cite{Kass88snakes:active}. ACM is initialized with a given contour and minimizes an energy function $E_{\textit{contour}}^*$, which moves the contour points toward an optimal position. Usually this energy is composed of a term to fit the contour to the image and additional terms to limit the amount of stretch and/or curvature. The optimization is performed by gradient descent. Overall the ACM lends itself perfectly for parallelized execution on the GPU, and the optimization can be performed using an automatic differentiation module included in deep learning frameworks. We adapt ACM so that the optimization is performed on a skeleton graph instead of contours, giving us the Active Skeleton Model (ASM). We call the skeleton graph the graph of connected pixels of the skeleton image obtained by the thinning method~\cite{Zha84} applied on the building wall probability map $y_{\textit{edge}}$. The following energy terms are used:
\begin{itemize}
	\item $E_{\textit{probability}}$ fits the skeleton paths to the contour of the building interior probability map $y_{\textit{int}}({v})$ at a certain probability threshold $\ell$ (set to 0.5 in practice).
	\item $E_{\textit{frame field align}}$ aligns each edge of the skeleton graph to the frame field.
	\item $E_{\textit{length}}$ ensures that the node distribution along paths remains homogeneous as well as tight.
\end{itemize}

Details about our data structure (designed for GPU computation), definition and computation of our energy terms, and explanation of our corner-aware simplification step can be found in the supplementary materials.

\section{Experimental setup}

\subsection{Datasets}

Our method requires ground truth polygonal building annotations (rather than raster binary masks) so that the ground truth angle for the frame field can be computed by rasterizing separately each polygon edge and taking the edge's angle. Thus, for each pixel we get a $\theta_\tau$ value, which is used in $L_{\textit{align}}$.

We perform experiments on these datasets (more details in the supplementary material):
\begin{itemize}
	\item CrowdAI Mapping Challenge dataset~\cite{CrowdAI} (\emph{CrowdAI dataset}): 341438 aerial images of size $300\!\times\!300$~pixels with associated ground truth polygonal annotations.
	\item Inria Aerial Image Labeling dataset~\cite{maggiori2017dataset} (\emph{Inria dataset}): 360 aerial images of size $5000\!\times\!5000$~pixels. Ten cities are represented, making it more varied than the \emph{CrowdAI dataset}. However, the ground truth is in the form of raster binary masks. We thus create the \emph{Inria OSM dataset} by taking OSM polygon annotations and correcting their misalignment using~\cite{Girard_2019_IGARSS}. We also create the \emph{Inria Polygonized dataset} by converting the original ground truth binary masks to polygon annotations with our polygonization method (see supplementary materials).
	\item \emph{Private dataset}: 57 satellite images for training with sizes varying from $2000\!\times\!2000$ pixels to $20000\!\times\!20000$ pixels, captured over 30 different cities from all continents with three different types of satellites. This is our most varied and challenging dataset. However, the building outline polygons were manually labeled precisely by an expert, ensuring the best possible ground truth. Results for this private dataset are in the supplementary material.
\end{itemize}

\subsection{Backbones}

The first backbone we use is \emph{U-Net16}, a small U-Net~\cite{unet} with 16 starting hidden features (instead of 64 in the original). We also use \emph{DeepLab101}, a DeepLabV3~\cite{DeepLabV3} model that utilizes a ResNet-101~\cite{ResNet} encoder. Our best performing model is \emph{UResNet101}---a U-Net with a ResNet-101~\cite{ResNet} encoder (pre-trained on ImageNet~\cite{imagenet}). We observed that the pre-trained ResNet-101 encoder achieves better final performance than random initialization. For the UResNet101, we additionally use distance weighting for the cross-entropy loss, as done for the original U-Net~\cite{unet}.

\subsection{Ablation study and additional experiments}

We perform an ablation study to validate various components of our method (results in Tables~\ref{tab:max_tangent_angle_error_results}, ~\ref{tab:coco_metrics_main_results}, and ~\ref{tab:inria_results}):
\begin{itemize}
	\item ``No field'' removes the frame field output for comparison to pure segmentation. Only interior segmentation $L_{\textit{int}}$, edge segmentation $L_{\textit{edge}}$ and interior/edge coupling $L_{\textit{int edge}}$ losses remain.
	\item ``Simple poly.'' uses a baseline polygonization algorithm (marching-squares contour detection followed by the Ramer–Douglas–Peucker simplification) on the interior classification map learned by our full method. This allows us to study the improvement of our polygonization method from leveraging the frame field.
\end{itemize}
Additional experiments in the supplementary material include: ``no coupling losses'' removes all coupling losses ($L_{\textit{int align}}$, $L_{\textit{edge align}}$, $L_{\textit{int edge}}$) to determine whether enforcing consistency between outputs has an impact; ``no $L_{\textit{align90}}$,'' ``no $L_{\textit{int edge}}$,'' ``no $L_{\textit{int align}}$ and $L_{\textit{edge align}}$,'' and ``no $L_{\textit{smooth}}$'' all remove the specified losses; ``complexity vs. fidelity'' varies the simplification tolerance parameter $\epsi$ to demonstrate the trade-off between complexity and fidelity of our corner-aware simplification procedure.

\subsection{Metrics}

The standard metric for image segmentation is Intersection over Union (IoU), which is then used to compute other metrics such as MS COCO~\cite{MS_COCO}, Average Precision (AP), and Average Recall (AR)---along with variants AP$_{50}$, AP$_{75}$, AR$_{50}$, AR$_{75}$. Since we aim to produce clean geometry, it is important to measure contour regularity, not captured by the area-based metrics IoU, AP, and AR. Moreover, as annotations are bound to have some alignment noise, only optimizing IoU will favor blurry segmentations with rounded corners over sharp segmentations, as the blurry ones correspond to the shape expectation of the noisy ground truth annotation; segmentation results with sharp corners may even yield a lower IoU than segmentations with rounded corners. We thus introduce the \emph{max tangent angle error} metric that compares the tangent angles between predicted polygons and ground truth annotations, penalizing contours not aligned with the ground truth. It is computed by uniformly sampling points along a predicted contour, computing the angle of the tangent for each point, and comparing it to the tangent angle of the closest point on the ground truth contour. The \emph{max tangent angle error} is the maximum tangent angle error over all sampled points.
More details about the computation of these metrics can be found in the supplementary material.

\section{Results and discussion}

\subsection{CrowdAI dataset}

\begin{figure*}[!htb]
	\rotatebox[origin=l]{90}{U-Net variant + ASIP \textcolor{white}{p}}
	\includegraphics[width=0.19\linewidth]{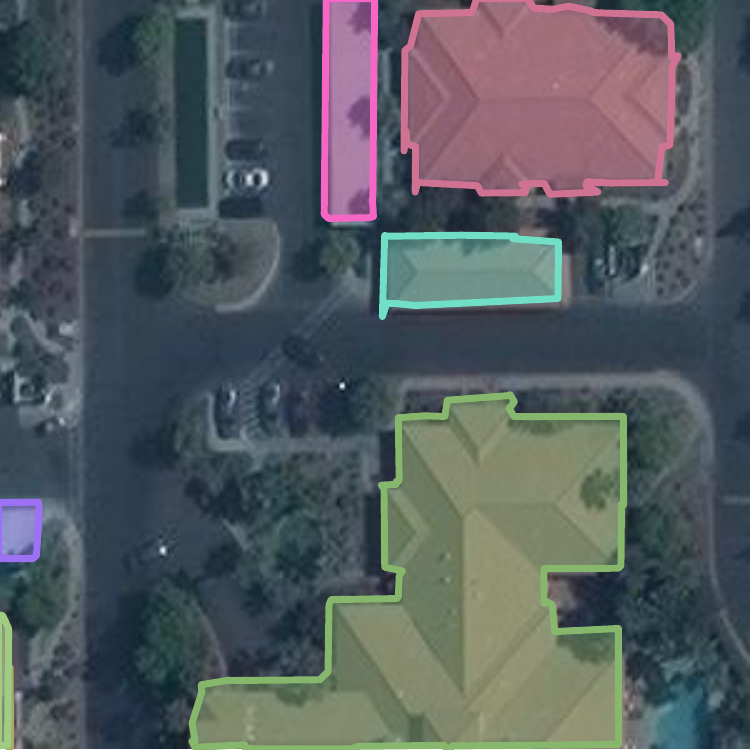}
	\includegraphics[width=0.19\linewidth]{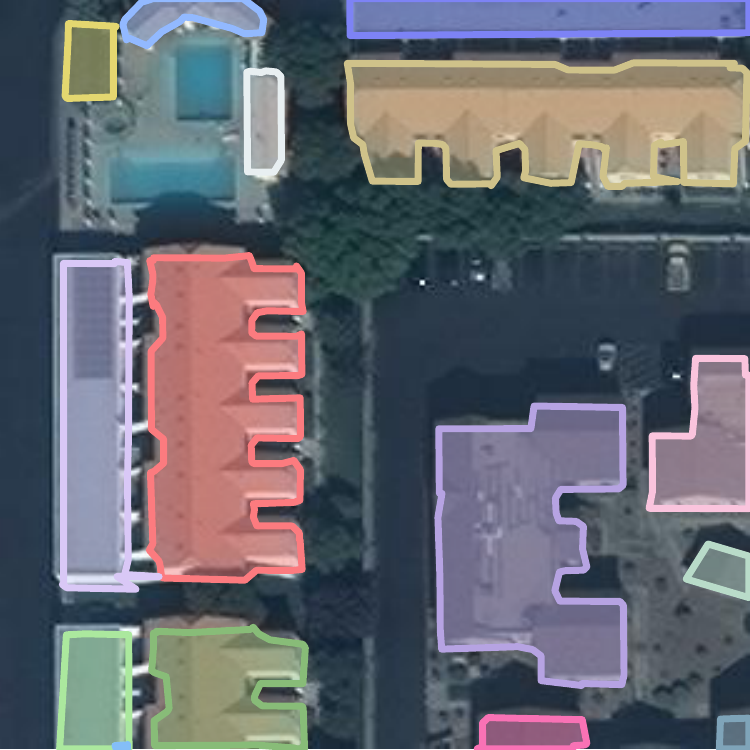}
	\includegraphics[width=0.19\linewidth]{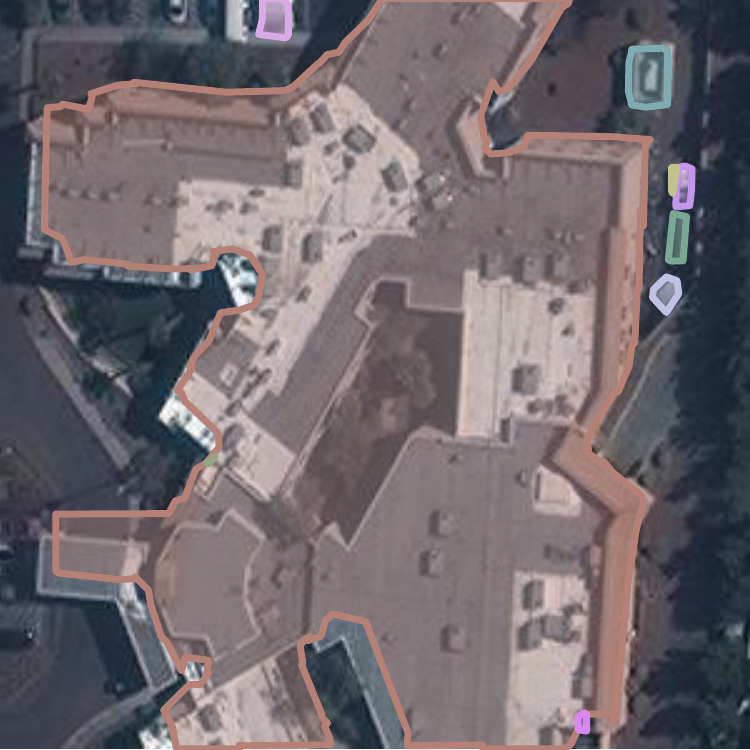}
	\includegraphics[width=0.19\linewidth]{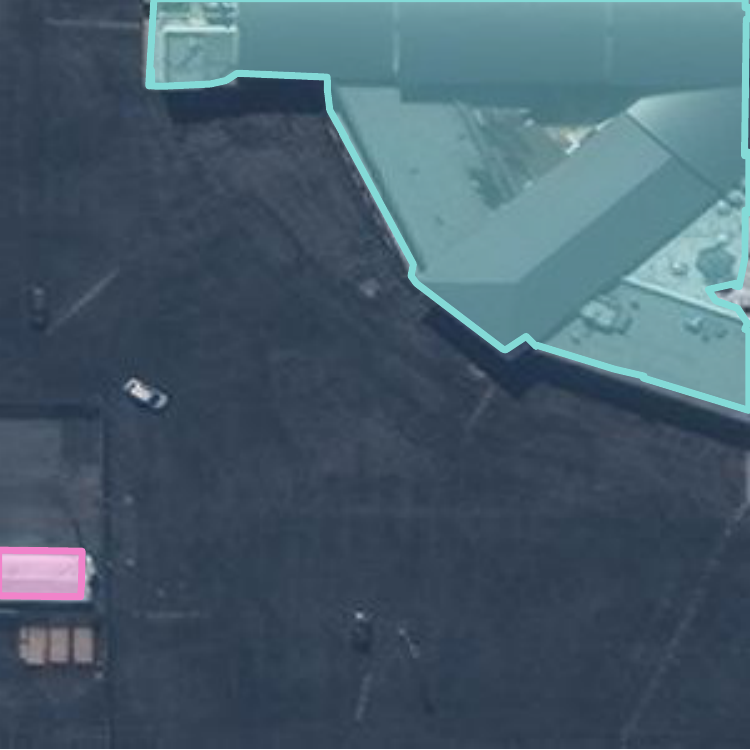}
	\includegraphics[width=0.19\linewidth]{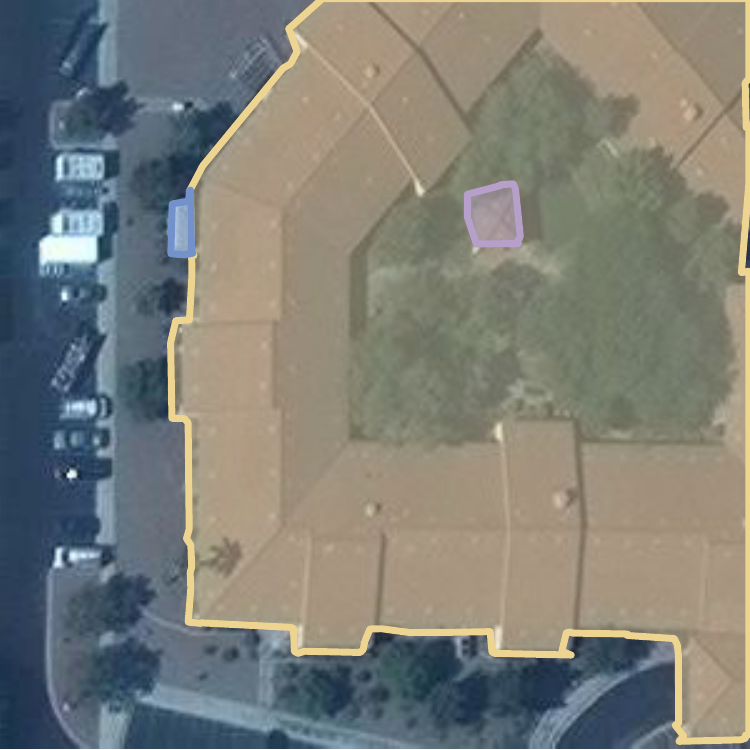}
	
	\rotatebox[origin=l]{90}{PolyMapper}
	\includegraphics[width=0.19\linewidth]{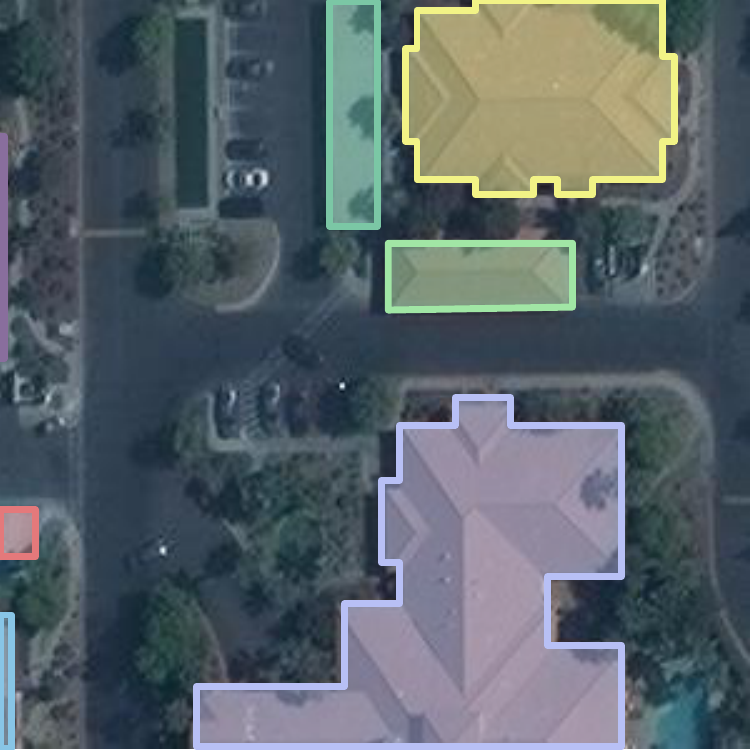}
	\includegraphics[width=0.19\linewidth]{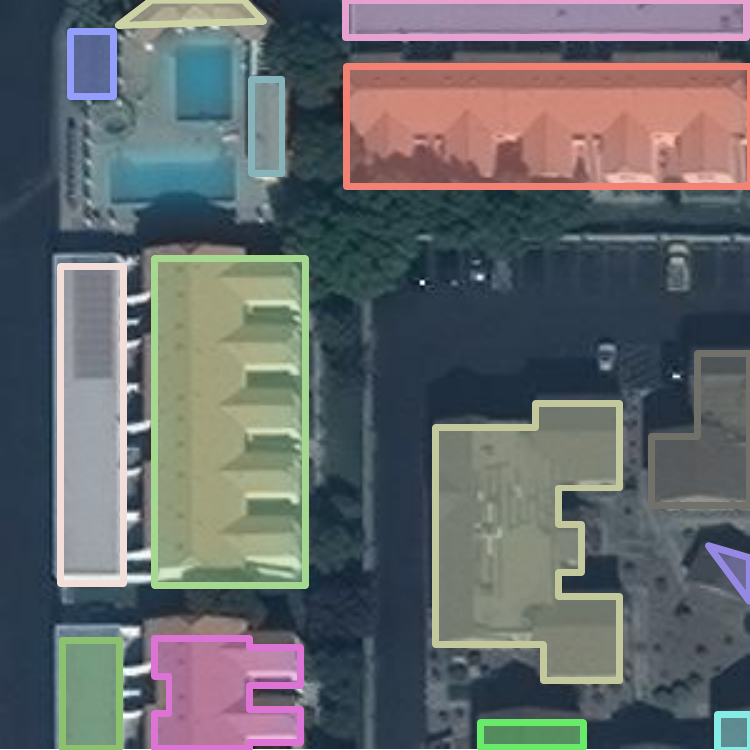}
	\includegraphics[width=0.19\linewidth]{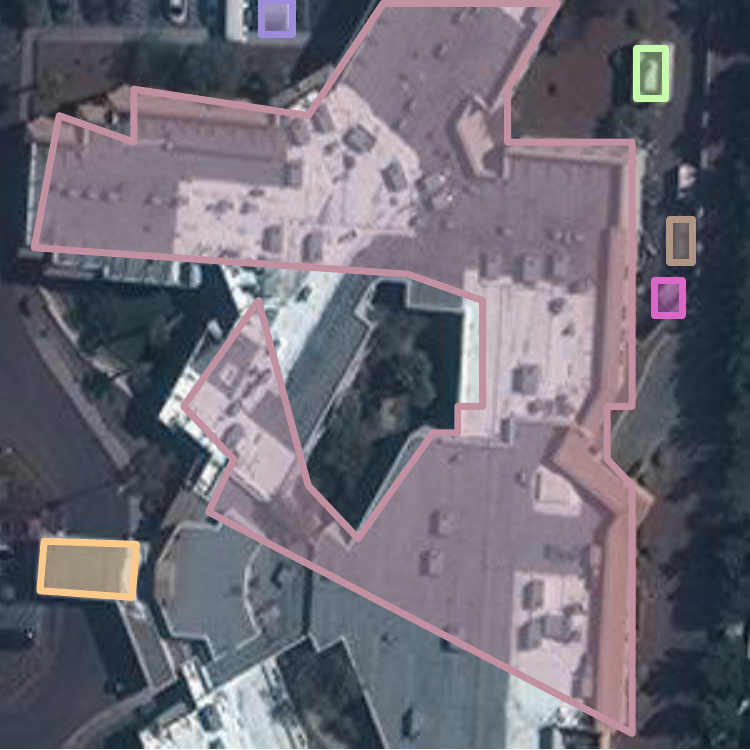}
	\includegraphics[width=0.19\linewidth]{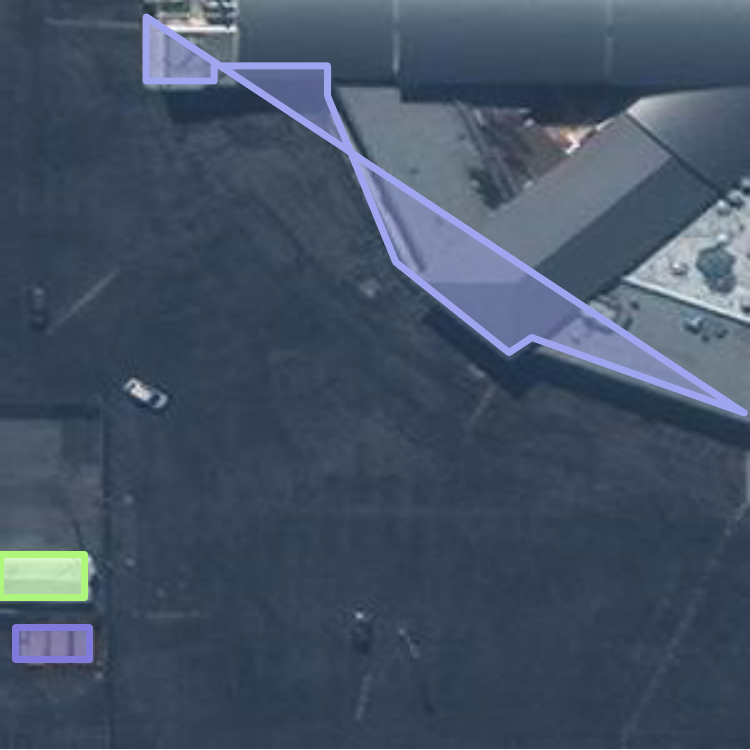}
	\includegraphics[width=0.19\linewidth]{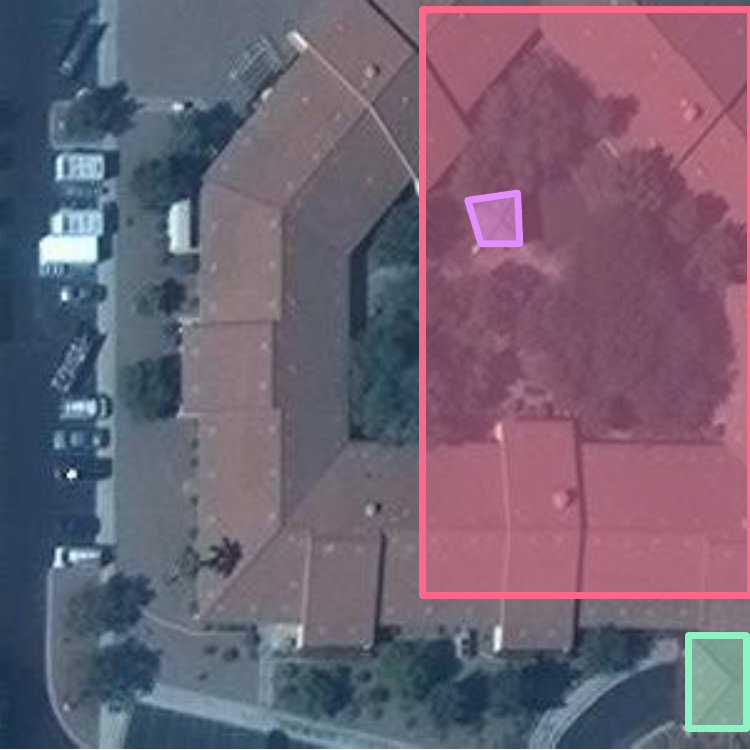}
	
	\rotatebox[origin=l]{90}{\textbf{Ours} \textcolor{white}{p}}
	\includegraphics[width=0.19\linewidth]{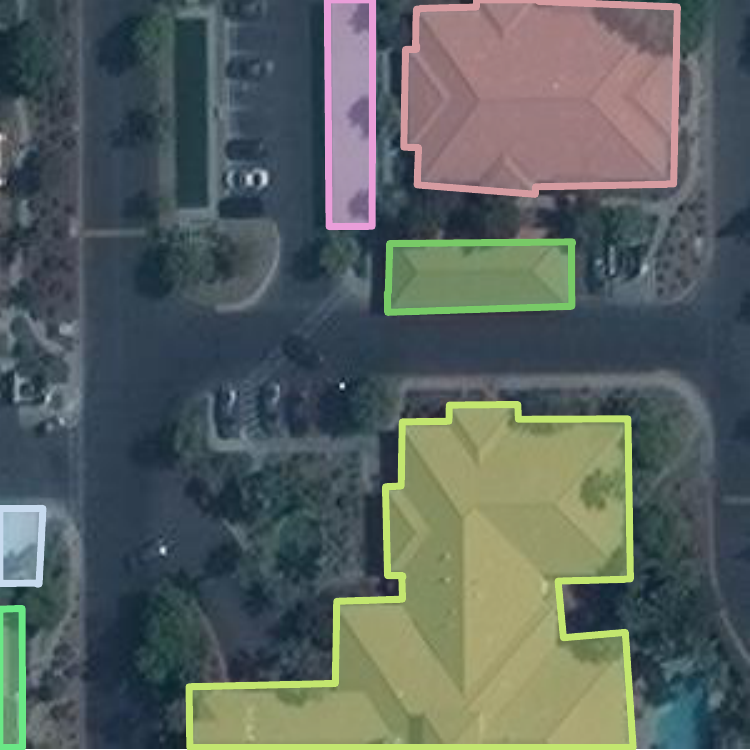}
	\includegraphics[width=0.19\linewidth]{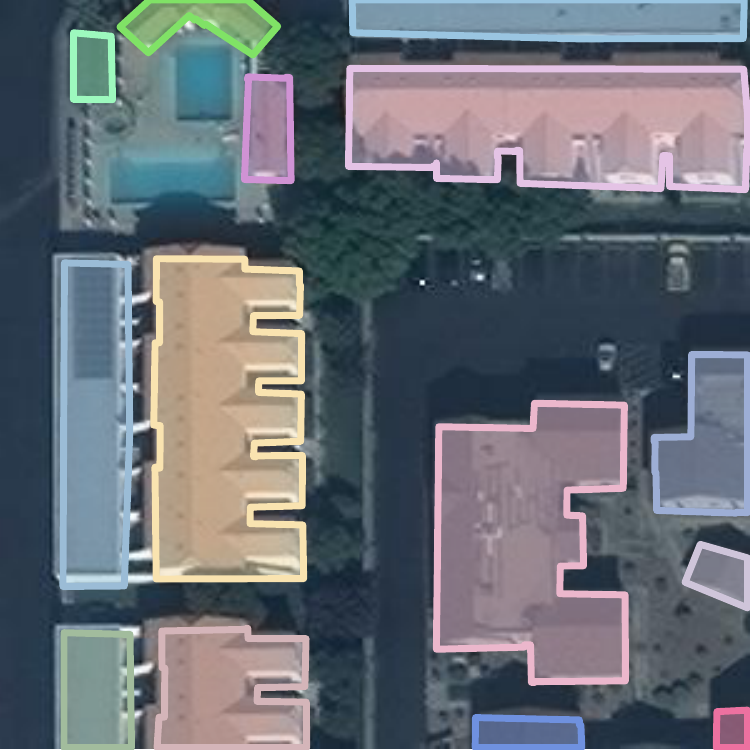}
	\includegraphics[width=0.19\linewidth]{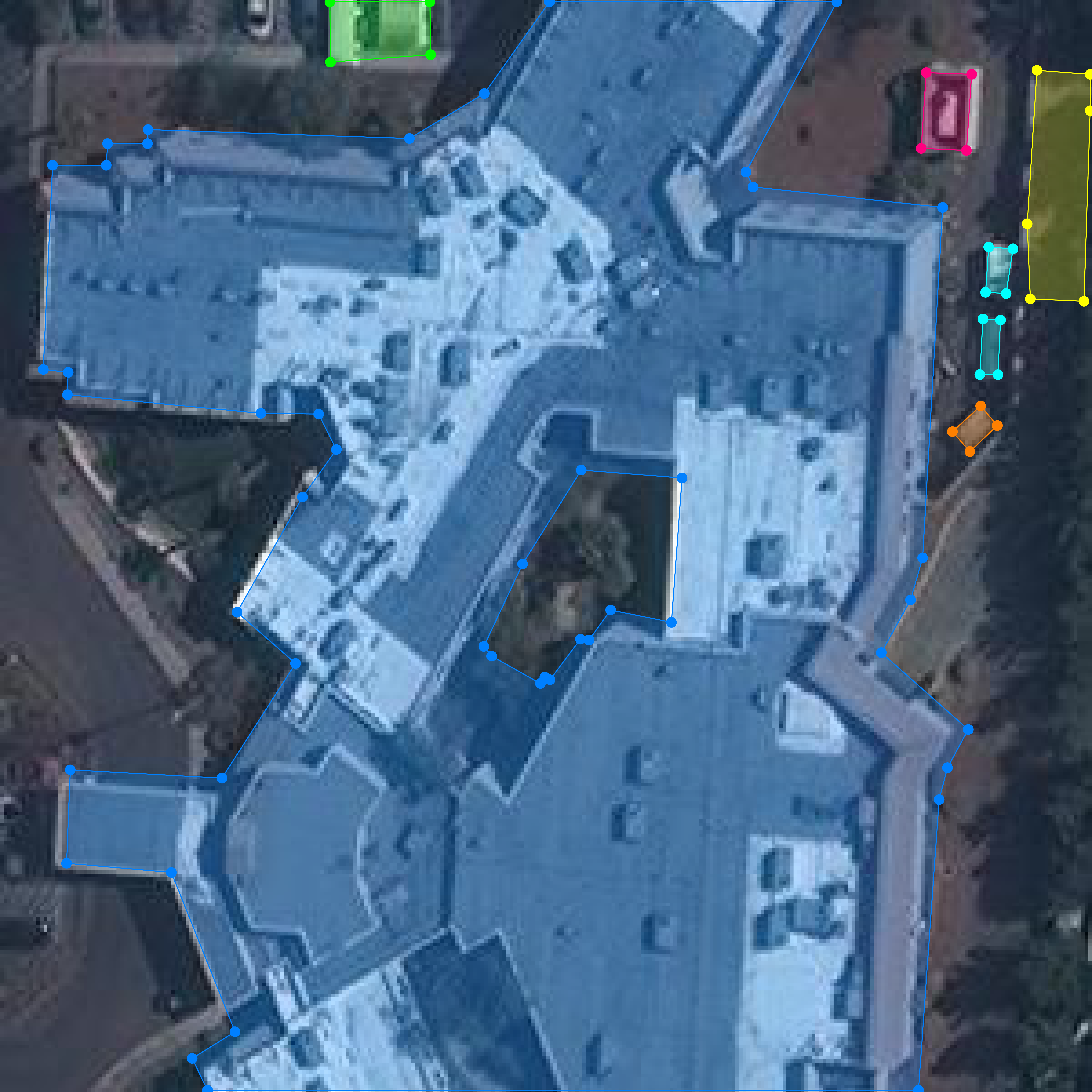}
	\includegraphics[width=0.19\linewidth]{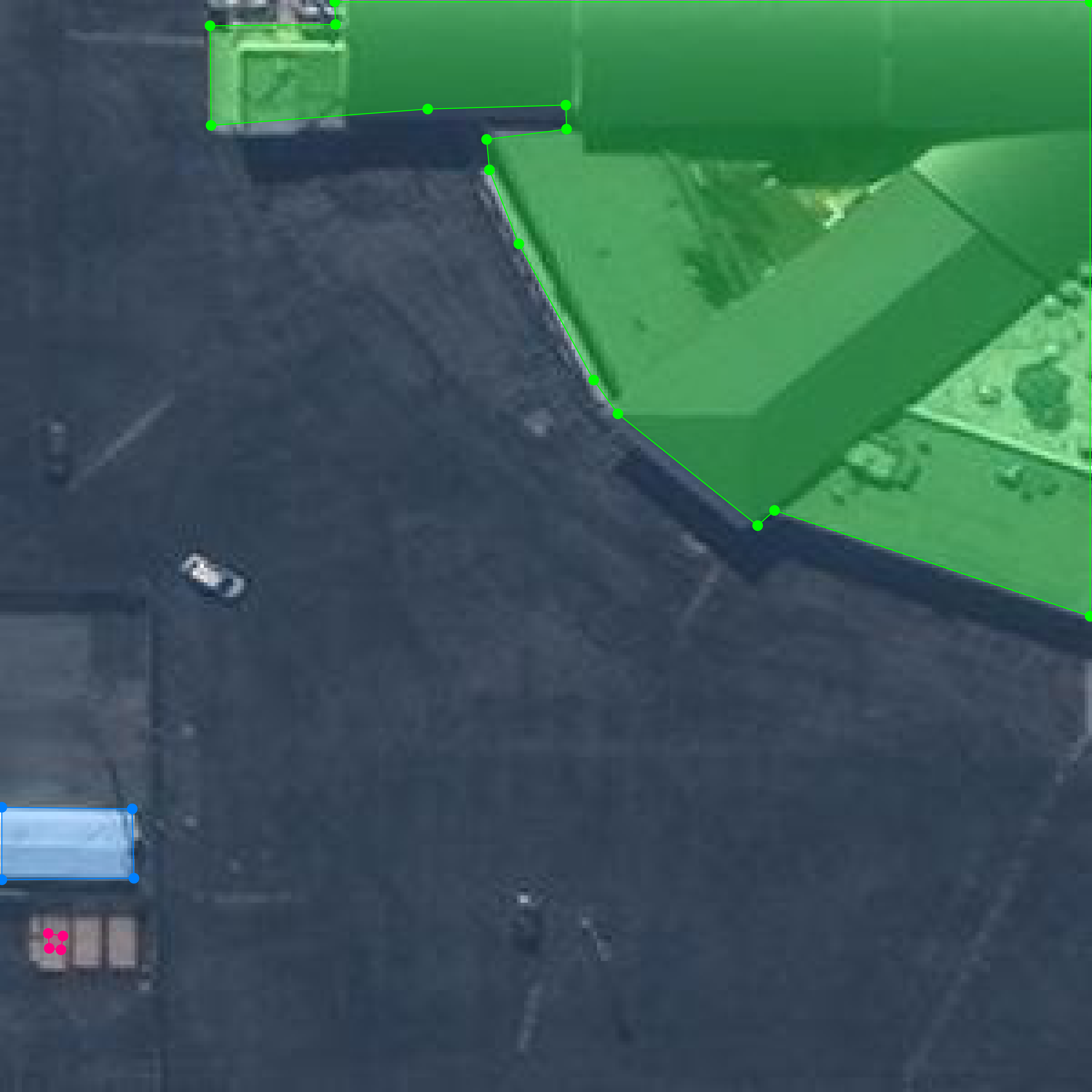}
	\includegraphics[width=0.19\linewidth]{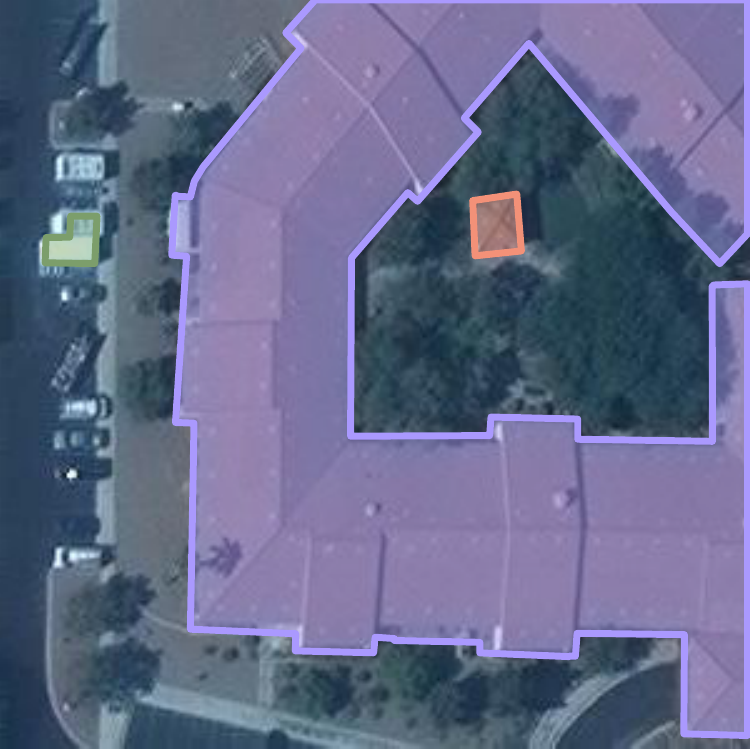}
	
	\caption{Example building extraction results on CrowdAI test images. Buildings become more complex from left to right. (top) U-Net variant~\cite{open_solution_crowdai} + ASIP~\cite{cvpr2020li}, (middle) PolyMapper~\cite{PolyMapper}, and (bottom) \textbf{ours}: UResNet101 (full), frame field polygonization.}
	\label{fig:polygonization_crownai_comparison_viz}
\end{figure*}

We visualize our polygon extraction results for the \emph{CrowdAI dataset} and compare them to other methods in Fig.~\ref{fig:polygonization_crownai_comparison_viz}. The ASIP polygonization method~\cite{cvpr2020li} inputs the probability maps of a U-Net variant~\cite{open_solution_crowdai} that won the CrowdAI challenge. All methods perform well on common building types, e.g., houses and residential buildings, but we can see that results of ASIP are less regular than PolyMapper and ours. For more complex building shapes (e.g., not rectangular or with a hole inside), ASIP outputs reasonable results, albeit still not very regular. However, the PolyMapper approach of object detection followed by polygonal outline regression does not work in the most difficult cases. It does not support nontrivial topology by construction, but also, it struggles with large complex buildings. We hypothesize that PolyMapper suffers from the fact that there are not many complex buildings and does not generalize as well as fully-convolutional networks.

We report results on the original validation set of the \emph{CrowdAI dataset} for the \emph{max tangent angle error} in Table~\ref{tab:max_tangent_angle_error_results} and MS COCO metrics in Table~\ref{tab:coco_metrics_main_results}. ``(with field)'' refers to models trained with our full frame field learning method, ``(no field)'' refers to models trained without any frame field output, ``mask'' refers to the output raster segmentation mask of the network, ``our poly.'' refers to our frame field polygonization method, and ``simple poly.'' refers to the baseline polygonization of marching squares followed by Ramer-Douglas-Peucker simplification. We also applied our polygonization method to the same probability maps used by the ASIP polygonization method (U-Net variant~\cite{open_solution_crowdai}) for fair comparison of polygonization methods. 

In Table~\ref{tab:max_tangent_angle_error_results}, ``simple poly.'' performs better using ``(with field)'' segmentation compared to ``(no field)'' because of a regularization effect from frame field learning. PolyMapper performs significantly better than ``simple poly.'' even though it is not explicitly regularized. Our frame field learning and polygonization method is necessary to decrease the error further and compare favorably to PolyMapper.
\begin{table}
	\centering
	\resizebox{\linewidth}{!}{
		\begin{tabular}{ l | c }
			\hline
			\textbf{Method}                                                           & Mean \emph{max tangent angle errors} $\downarrow$ \\ 
			\hline
			UResNet101 (no field), simple poly.                                       & 51.9\degree \\
			UResNet101 (with field), simple poly.                                     & 45.1\degree \\
			U-Net variant~\cite{open_solution_crowdai}, ASIP poly.~\cite{cvpr2020li}  & 44.0\degree \\
			UResNet101 (with field), ASIP poly.~\cite{cvpr2020li}                     & 38.3\degree \\
			U-Net variant~\cite{open_solution_crowdai}, UResNet101 \textbf{our} poly. & 36.6\degree \\
			PolyMapper~\cite{PolyMapper}                                              & 33.1\degree \\
			UResNet101 (with field), \textbf{our} poly.                               & \textbf{31.9}\degree \\
			\hline
		\end{tabular}
	}
	\caption{Mean \emph{max tangent angle errors} over all the original validation polygons of the \emph{CrowdAI dataset}~\cite{CrowdAI}.}
	\label{tab:max_tangent_angle_error_results}
\end{table}

In Table~\ref{tab:coco_metrics_main_results}, our UResNet101 (with field) outperforms most previous works, except ``U-Net variant~\cite{open_solution_crowdai}, ASIP poly.~\cite{cvpr2020li}'' due to the U-Net variant being the winning entry to the challenge. However our polygonization applied after that same U-Net variant achieves better max tangent angle error and AP than ASIP but worse AR. The same is true when applying ASIP to our UResNet101 (with field): it has slightly worse AP, AR, and max tangent angle error. However, the ASIP method also results in better max tangent angle error when using our UResNet101 (with field) compared to using the U-Net variant.
\begin{table}
	\centering
	\resizebox{\linewidth}{!}{
		\begin{tabular}{ l | c c c c c c | c c c c c c }
			\hline
			\textbf{Method}                                                           & $AP$ $\uparrow$ & $AP_{50}$ $\uparrow$ & $AP_{75}$ $\uparrow$ & $AR$ $\uparrow$ & $AR_{50}$ $\uparrow$ & $AR_{75}$ $\uparrow$ \\
			\hline
			UResNet101 (no field), mask                                               & 62.4 & 86.7 & 72.7 & 67.5 & 90.5 & 77.4 \\
			UResNet101 (no field), simple poly.                                       & 61.1 & 87.4 & 71.2 & 64.7 & 89.4 & 74.1 \\
			\hline
			UResNet101 (with field), mask                                             & 64.5 & 89.3 & 74.6 & 68.1 & 91.0 & 77.7 \\
			UResNet101 (with field), simple poly.                                     & 61.7 & 87.7 & 71.5 & 65.4 & 89.9 & 74.6 \\
			UResNet101 (with field), \textbf{our} poly.                               & 61.3 & 87.5 & 70.6 & 65.0 & 89.4 & 73.9 \\
			UResNet101 (with field), ASIP poly.~\cite{cvpr2020li}                     & 60.0 & 86.3 & 69.9 & 64.0 & 88.8 & 73.4 \\
			\hline
			U-Net variant~\cite{open_solution_crowdai}, UResNet101 \textbf{our} poly. & \textbf{67.0} & \textbf{92.1} & \textbf{75.6} & 73.2 & 93.5 & 81.1 \\
			\hline
			\hline
			Mask R-CNN~\protect\cite{He_2017_ICCV}~\protect\cite{CrowdAIBaseline}     & 41.9 & 67.5 & 48.8 & 47.6 & 70.8 & 55.5 \\
			PANet~\protect\cite{liu2018path}                                          & 50.7 & 73.9 & 62.6 & 54.4 & 74.5 & 65.2 \\
			PolyMapper~\protect\cite{PolyMapper}                                      & 55.7 & 86.0 & 65.1 & 62.1 & 88.6 & 71.4 \\
			U-Net variant~\cite{open_solution_crowdai}, ASIP poly.~\cite{cvpr2020li}  & 65.8 & 87.6 & 73.4 & \textbf{78.7} & \textbf{94.3} & \textbf{86.1} \\
			\hline
	\end{tabular}}
	\caption{AP and AR results on the \emph{CrowdAI dataset}~\protect\cite{CrowdAI} for all polygonization experiments.}
	\label{tab:coco_metrics_main_results}
\end{table}

\begin{table}
	\centering
	\begin{tabular}{ r | c | c }
		\hline
		\textbf{Method} & Time (sec) $\downarrow$ & Hardware\\ 
		\hline
		PolyMapper~\protect\cite{PolyMapper} & 0.38 & GTX 1080Ti  \\
		ASIP~\protect\cite{cvpr2020li} & 0.15 & Laptop CPU \\
		\textbf{Ours} & \textbf{0.04} & GTX 1080Ti \\
		\hline
	\end{tabular}
	\caption{Average times to extract buildings from a $300\!\times\!300$ pixel patch. \textbf{Ours} refers to UResNet101 (with field), our poly. ASIP's time does not include model inference.}
	\label{tab:running_times}
\end{table}

\paragraph{Runtimes.} We compare runtimes in Table~\ref{tab:running_times}. ASIP does not have a GPU implementation. In their paper they give an average runtime of 1-3s on CPU with \texttildelow10\% CPU utilization. Assuming perfect parallelization, they estimate their average runtime to be 0.15s with 100\% CPU utilization. Their method uses a priority queue for optimizing the polygonal partitioning with various geometric operators and is harder to implement on GPU. Our efficient data structure makes our building extraction competitive with prior work.

\subsection{Inria OSM dataset}

\begin{figure}[!thb]
	\begin{subfigure}[t]{0.49\linewidth}
		{\small U-Net16 (no field), simple poly.}
		
		\includegraphics[width=\linewidth]{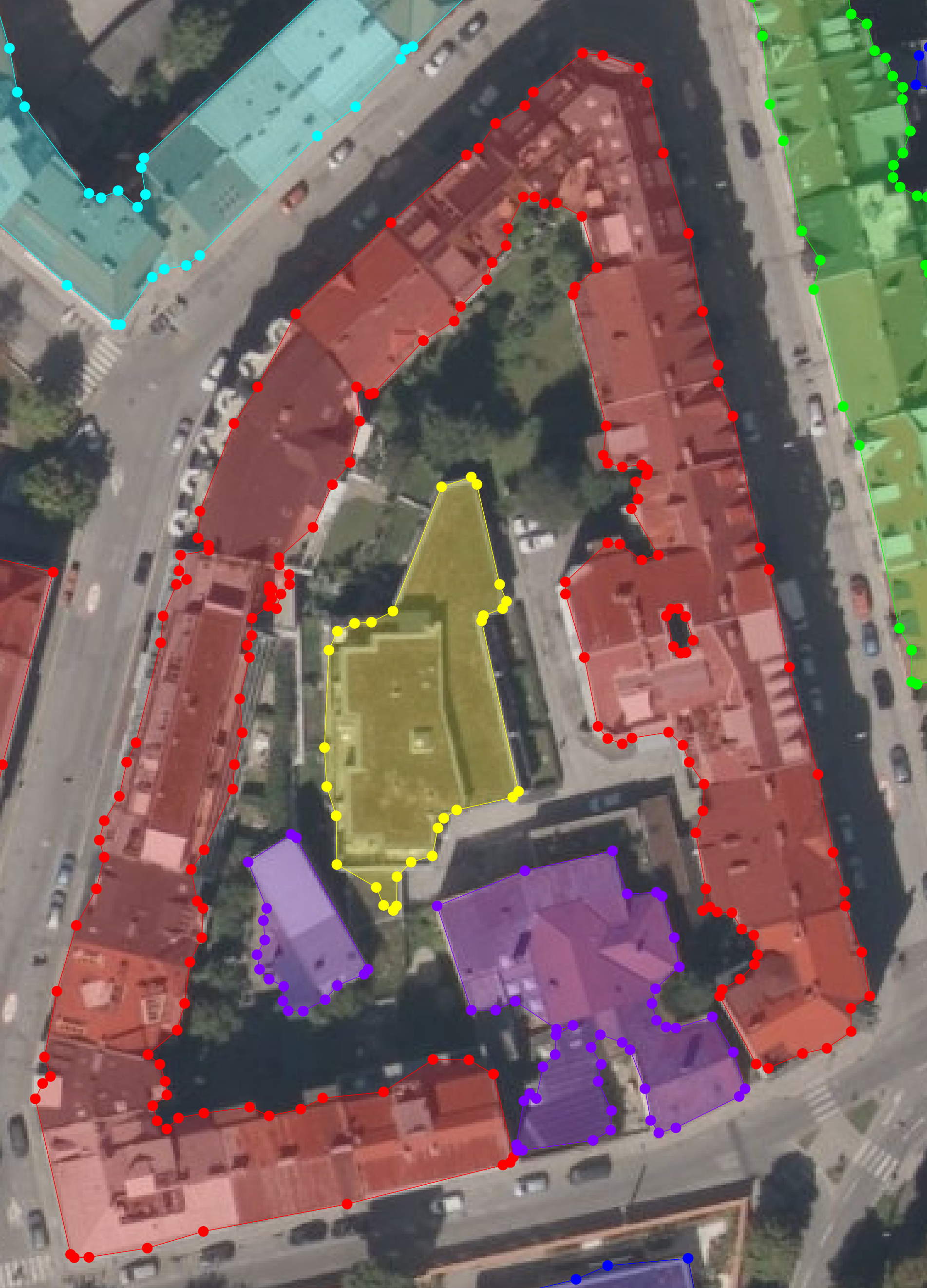}
	\end{subfigure}
	\begin{subfigure}[t]{0.49\linewidth}
		{\small U-Net16 (with field), \textbf{our} poly.}
		
		\includegraphics[width=\linewidth]{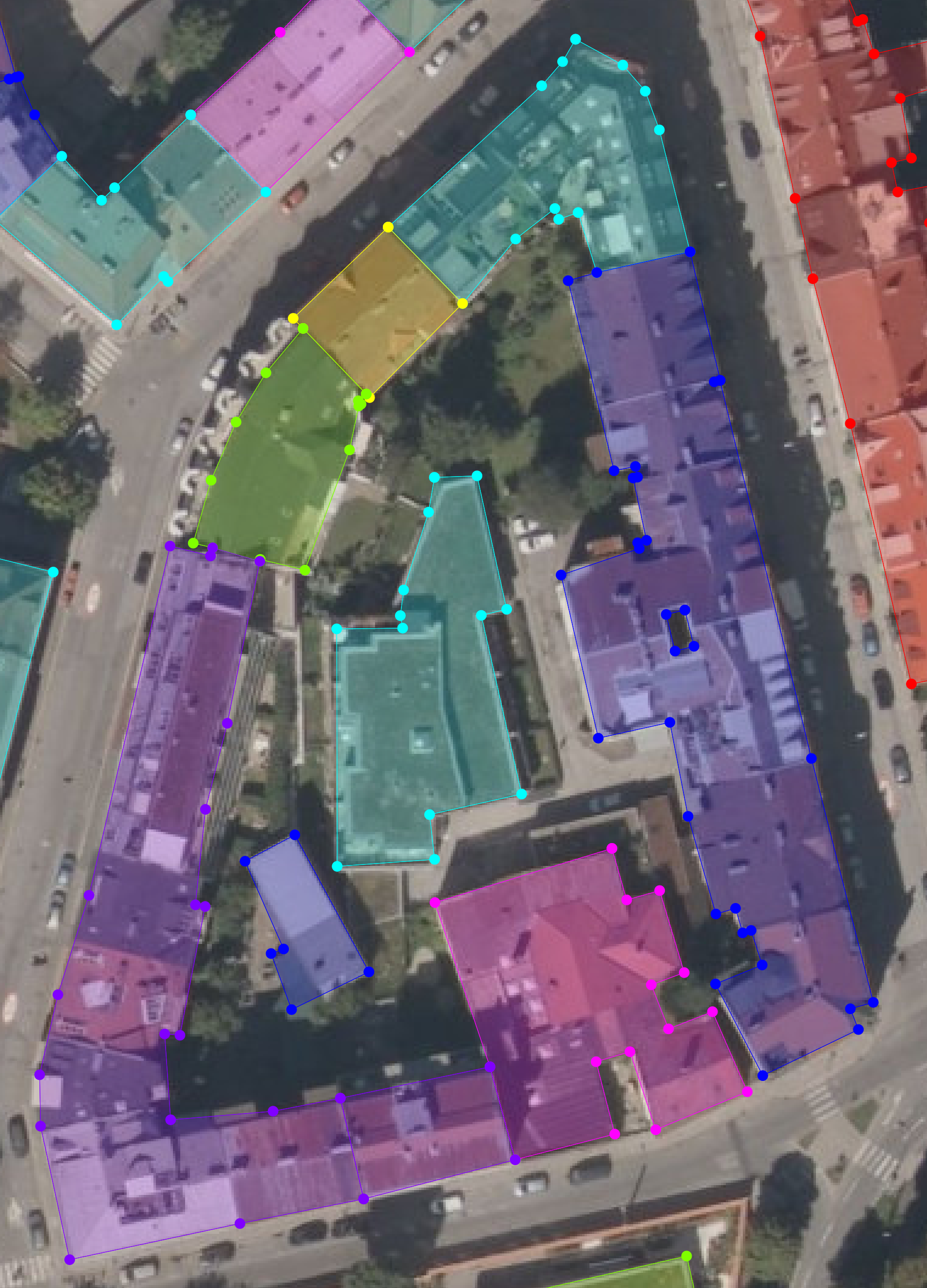}
	\end{subfigure}	
	\caption{Small crop of \emph{Inria dataset} results.}
	\label{fig:inria_results_innsbruck19_small_crop}
\end{figure}

The \emph{Inria OSM dataset} is more challenging than the \emph{CrowdAI dataset} because it contains more varied areas (e.g., countryside, city center, residential, and commercial) with different building types. It also contains adjacent buildings with common walls, which our edge segmentation output can detect. The mean IoU on test images of the output classification maps is 78.0\% for the U-Net16 trained with a frame field compared to 76.9\% for the U-Net16 with no frame field. The IoU does not significantly penalize irregular contours, but, by visually inspecting segmentation outputs as in Fig.~\ref{fig:inria_results_innsbruck19_small_crop}, we can see the effect of the regularization. Our method successfully handles complex building shapes which can be very large, with blocks of buildings featuring common walls and holes. See the supplementary materials for more results.

\subsection{Inria polygonized dataset}

\begin{figure}[!htb]
	\begin{subfigure}[t]{0.49\linewidth}
		\resizebox{\linewidth}{!}{Eugene Khvedchenya\repeatfootnote{inria-challenge}, simple poly.}
		
		\includegraphics[width=\linewidth]{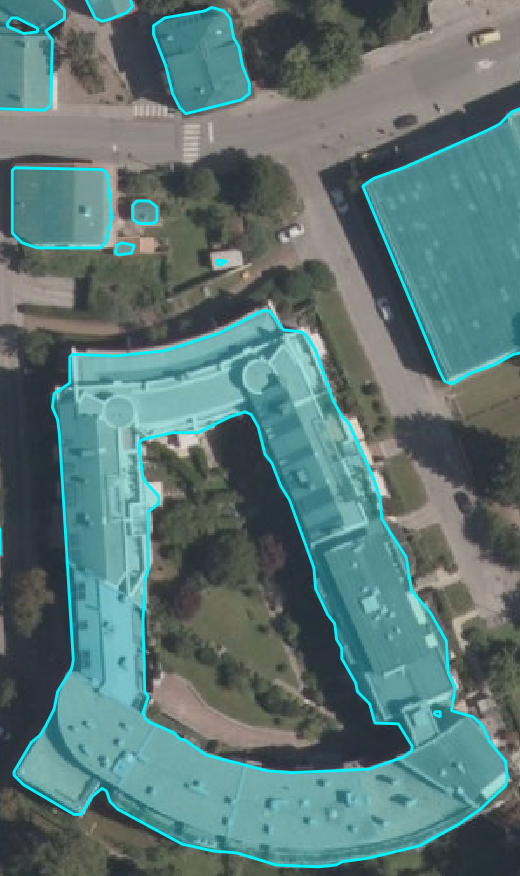}
	\end{subfigure}
	\begin{subfigure}[t]{0.49\linewidth}
		\resizebox{\linewidth}{!}{UResNet101 (with field), \textbf{our} poly.}
		
		\includegraphics[width=\linewidth]{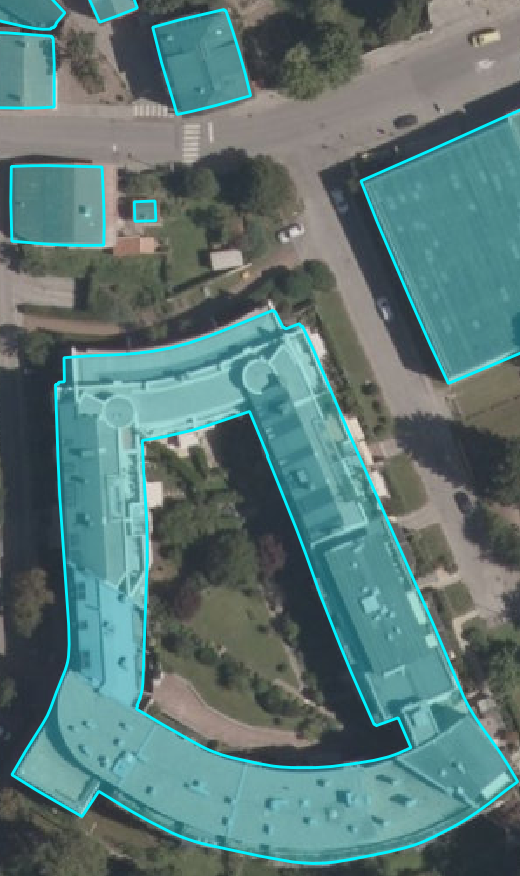}
	\end{subfigure}	
	
	\caption{Crop of an \emph{Inria polygonized dataset} test image.}
	\label{fig:inria_polygonized_results_small_crop}
\end{figure}

\begin{table}
	\centering
	\resizebox{\linewidth}{!}{
		\begin{tabular}{ l | c | c }
			\hline
			\textbf{Method}                                                  & mIoU $\uparrow$ & Mean \emph{max tangent angle errors} $\downarrow$ \\
			\hline
			Eugene Khvedchenya\repeatfootnote{inria-challenge}, simple poly. & \textbf{80.7}\% & 52.2 \degree \\
			ICTNet~\cite{chatterjee2019building}, simple poly.               & \textbf{80.1}\% & 52.1\degree \\
			UResNet101 (no field), simple poly.                              & 73.2\% &          52.0\degree \\
			Zorzi et al.~\cite{zorzi2020machinelearned} poly.                & 74.4\% &          34.5\degree \\
			UResNet101 (with field), \textbf{our} poly.                      & 74.8\% &          \textbf{28.1}\degree \\
			\hline
		\end{tabular}
	}
	\caption{IoU and mean \emph{max tangent angle errors} for polygon extraction methods on the \emph{Inria polygonized dataset}.}
	\label{tab:inria_results}
\end{table}

The \emph{Inria polygonized dataset} with its associated challenge\savefootnote{inria-challenge}{\url{https://project.inria.fr/aerialimagelabeling/leaderboard/}} allows us to directly compare to other methods trained on the same ground truth, even though it does not consider learning of separate buildings. In Table~\ref{tab:inria_results}, our method matches~\cite{zorzi2020machinelearned} in terms of mIoU, with lower \emph{max tangent angle error}. The two top methods on the leaderboard (ICTNet~\cite{chatterjee2019building} and ``Eugene Khvedchenya'') achieve a mIoU over 80\%, but they lack contour regularity with high \emph{max tangent angle error}; they also only output segmentation masks, needing \emph{a posteriori} polygonization to extract polygonal buildings. Fig.~\ref{fig:inria_polygonized_results_small_crop} shows the cleaner geometry of our method. The ground truth of the \emph{Inria polygonized dataset} has misalignment noise, yielding imprecise corners that produce rounded corners in the prediction if no regularization is applied. See the supplementary materials for more results.

\section{Conclusion}

We improve on the task of building extraction by learning an additional output to a standard segmentation model: a frame field. This motivates the use of a regularization loss, leading to more regular contours, e.g., with sharp corners. Our approach is efficient since the model is a single fully-convolutional network. The training is straightforward, unlike adversarial training, direct shape regression, and recurrent networks, which require significant tuning and more computational power. The frame field adds virtually no cost to inference time, and it disambiguates tough polygonization cases, making our polygonization method less complex. Our data structure for the polygonization makes it parallelizable on the GPU. We handle the case of holes in buildings as well as common walls between adjoining buildings. Because of the skeleton graph structure, common wall polylines are naturally guaranteed to be shared by the buildings on either side. 
As future work, we could apply our method to any image segmentation network, including multi-class segmentation, where the frame field could be shared between all classes. 

\section{Acknowledgements}

Thanks to ANR for funding the project EPITOME ANR-17-CE23-0009 and to Inria Sophia Antipolis - Méditerranée ``Nef" computation cluster for providing resources and support. The MIT Geometric Data Processing group acknowledges the generous support of Army Research Office grant W911NF2010168, of Air Force Office of Scientific Research award FA9550-19-1-031, of National Science Foundation grant IIS-1838071, from the CSAIL Systems that Learn program, from the MIT–IBM Watson AI Laboratory, from the Toyota--CSAIL Joint Research Center, from a gift from Adobe Systems, from an MIT.nano Immersion Lab/NCSOFT Gaming Program seed grant, and from the Skoltech--MIT Next Generation Program. This work was also supported by the National Science Foundation Graduate Research Fellowship under Grant No. 1122374.

%% file: supp_mat_content.tex
\begingroup
    \hypersetup{linkcolor=black}
    \tableofcontents
\endgroup


\section{Frame field learning details}

\subsection{Model architecture}
\begin{figure*}[ht]
	\includegraphics[width=\linewidth]{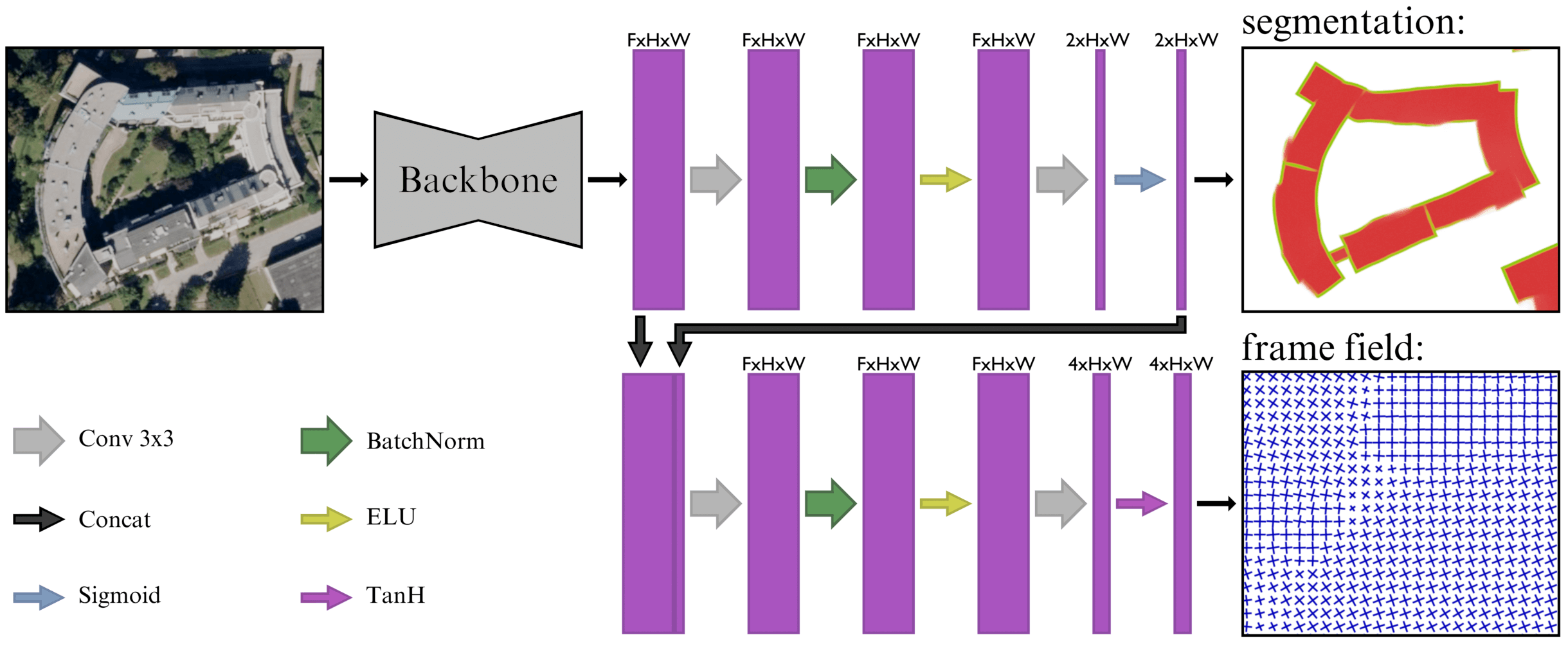}
	\caption{Details of our network's architecture with the addition of the frame field output.}
	\label{fig:backbone_with_extra_layers}
\end{figure*}

We show in Fig.~\ref{fig:backbone_with_extra_layers} how we add a frame field output to an image segmentation backbone. The backbone can be any (possibly pretrained) network as long as it outputs an $F$-dimensional feature map $\yh_\textit{backbone} \in \R^{F \times H \times W}$.

\subsection{Losses}

We define image segmentation loss functions below:
{
	\small
	\begin{multline}
		L_{\textit{BCE}}(y, \yh) = \frac{1}{HW} \sum_{{x} \in I} y({x}) \cdot \log(\yh({x})) \\
		+ (1 - y({x})) \cdot \log(1 - \yh({x})),
	\end{multline}
	\begin{equation}
		L_{\textit{Dice}}(y, \yh) = 1 - 2 \cdot \frac{|y \cdot y| + 1}{|y + y| + 1},
	\end{equation}
	\begin{equation}
		L_{\textit{int}} = \alpha \cdot L_{\textit{BCE}}(y_{\textit{int}}, \yh_{\textit{int}}) + (1-\alpha) \cdot L_{\textit{Dice}}(y_{\textit{int}}, \yh_{\textit{int}}),
	\end{equation}
	\begin{equation}
		L_{\textit{edge}} = \alpha \cdot L_{\textit{BCE}}(y_{\textit{edge}}, \yh_{\textit{edge}}) + (1-\alpha) \cdot L_{\textit{Dice}}(y_{\textit{edge}}, \yh_{\textit{edge}}),
	\end{equation}
}
where $0 < \alpha < 1$ is a hyperparameter. In practice, $\alpha = 0.25$ gives good results.

\begin{figure}[ht]
	\centering
	\includegraphics[width=0.5\textwidth]{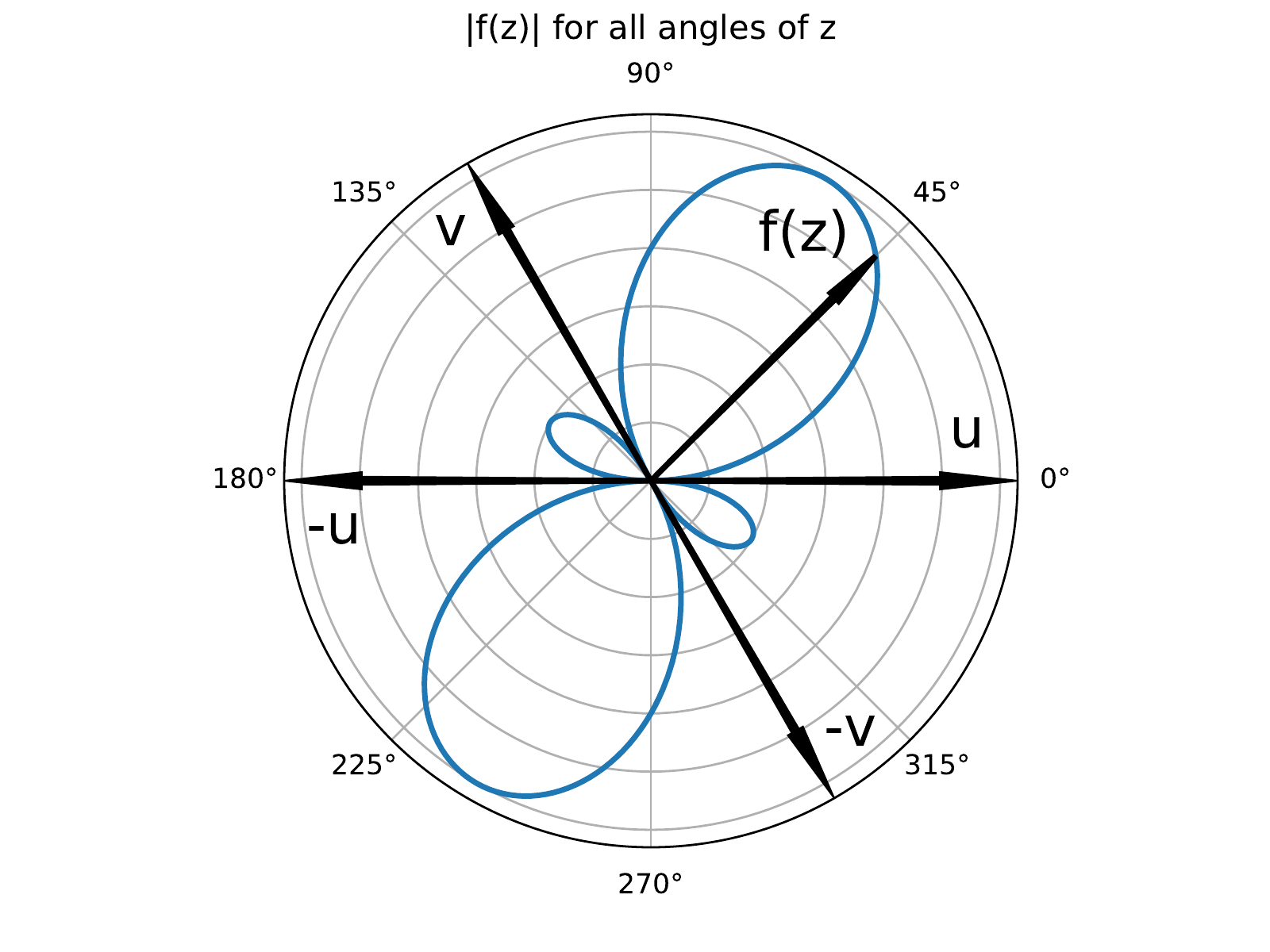}
	\caption{Visualization of the frame field align loss $L_{\textit{align}}$ (in blue) for a certain configuration of $\{ -u, u, -v, v \}$ and all possible ground truth $z = e^{i\theta_{\tau}}$ directions.}
	\label{fig:frame_field_loss}
\end{figure}

For the frame field, we show a visualization of the $L_{\textit{align}}$ loss in Fig.~\ref{fig:frame_field_loss}.

\subsection{Handling numerous heterogeneous losses}

We linearly combine our eight losses using eight coefficients, which can be challenging to balance. Because the losses have different units, we first compute a normalization coefficient $N_{\langle\textit{loss name}\rangle}$ by computing the average of each loss on a random subset of the training dataset using a randomly-initialized network. Then each loss can be normalized by this coefficient. The total loss is a linear combination of all normalized losses:
\begin{multline}
	\lambda_{\textit{int}} \frac{L_{\textit{int}}}{N_{\textit{int}}} + 
	\lambda_{\textit{edge}} \frac{L_{\textit{edge}}}{N_{\textit{edge}}} + 
	\lambda_{\textit{align}} \frac{L_{\textit{align}}}{N_{\textit{align}}} + 
	\lambda_{\textit{align90}} \frac{L_{\textit{align90}}}{N_{\textit{align90}}} \\ + 
	\lambda_{\textit{smooth}} \frac{L_{\textit{smooth}}}{N_{\textit{smooth}}} + 
	\lambda_{\textit{int align}} \frac{L_{\textit{int align}}}{N_{\textit{int align}}} \\ + 
	\lambda_{\textit{edge align}} \frac{L_{\textit{edge align}}}{N_{\textit{edge align}}} + 
	\lambda_{\textit{int edge}} \frac{L_{\textit{int edge}}}{N_{\textit{int edge}}},
\end{multline}
where the $\lambda_{\langle\textit{loss name}\rangle}$ coefficients are to be tuned. It is also possible to separately group the main losses and the regularization losses and have a single $\lambda$ coefficient balancing the two loss groups:
\begin{equation}
	\lambda L_{\textit{main}} + (1 - \lambda) L_{\textit{regularization}},
\end{equation}
with
\begin{equation}
	L_{\textit{main}} = \frac{L_{\textit{int}}}{N_{\textit{int}}} + \frac{L_{\textit{edge}}}{N_{\textit{edge}}} + \frac{L_{\textit{align}}}{N_{\textit{align}}},
\end{equation}
\begin{multline}
	L_{\textit{regularization}} = \frac{L_{\textit{align90}}}{N_{\textit{align90}}} + 
	\frac{L_{\textit{smooth}}}{N_{\textit{smooth}}} \\ + 
	\frac{L_{\textit{int align}}}{N_{\textit{int align}}} + 
	\frac{L_{\textit{edge align}}}{N_{\textit{edge align}}} + 
	\frac{L_{\textit{int edge}}}{N_{\textit{int edge}}}.
\end{multline}
In practice we started experiments with the single-coefficient version with $\lambda = 0.75$ and then used the multi-coefficient version to have more control by setting $\lambda_{\textit{int}} = \lambda_{\textit{edge}} = 10$, $\lambda_{\textit{align}} = 1$, $\lambda_{\textit{align90}} = 0.2$, $\lambda_{\textit{smooth}} = 0.005$, $\lambda_{\textit{int align}} = \lambda_{\textit{edge edge}} = \lambda_{\textit{int edge}} = 0.2$.

\subsection{Training details}

We do not heavily tune our hyperparameters: once we find a value that works based on validation performance we keep it across ablation experiments. We employ early stopping for the U-Net16 and DeepLabV3 models (25 and 15 epochs, respectively) chosen by first training the full method on the training set of the \emph{CrowdAI dataset}, choosing the epoch number of the lowest validation loss, and finally re-training the model on the train and validation sets for that number of total epochs.

Segmentation losses $L_{\textit{int}}$ and $L_{\textit{edge}}$ are both a combination of 25\% cross-entropy loss and 75\% Dice loss. To balance the losses in ablation experiments, we used the single-coefficient version with $\lambda = 0.75$. For our best performing model UResNet101 we used the multi-coefficients version to have more control by setting $\lambda_{\textit{int}} = \lambda_{\textit{edge}} = 10$, $\lambda_{\textit{align}} = 1$, $\lambda_{\textit{align90}} = 0.2$, $\lambda_{\textit{smooth}} = 0.005$, $\lambda_{\textit{int align}} = \lambda_{\textit{edge edge}} = \lambda_{\textit{int edge}} = 0.2$. 
The U-Net16 was trained on 4 GTX 1080Ti GPUs in parallel on $512\!\times\!512$ patches and a batch size of 16 per GPU (effective batch size 64). For all training runs, we compute for each loss its normalization coefficient $N_{\langle loss\_name\rangle}$ on 1000 batches before optimizing the network.

Our method is implemented in PyTorch~\cite{PyTorch}. On the \emph{CrowdAI dataset}, training takes 2 hours per epoch on 4 1080Ti GPUs for the U-Net16 model and 3.5 hours per epoch for the DeepLabV3 backbone on 4 2080Ti GPUs. Inference with the U-Net16 on a $5000\!\times\!5000$ image (requires splitting into $1024\!\times\!1024$ patches) takes 7 seconds on a Quadro M2200 (laptop GPU).

\section{Frame field polygonization details}

We expand here on the algorithm and implementation details of our frame field polygonization method.

\subsection{Data structure}

Our polygonization method needs to be initialized with geometry, which is then optimized to align to the frame field (among other objectives we will present later).

In the case of extracting individual buildings, we use the marching squares~\cite{marching_cubes} contour finding algorithm on the predicted interior probability map $y_{\textit{int}}$ with an isovalue $\ell$ (set to $0.5$ in practice). The result is a collection of contours $\{\mathcal{C}_i\}$ where each contour is a sequence of 2D points: 
$$ \mathcal{C}_i = \big((r_0, c_0), (r_1, c_1), ..., (r_{n_i-1}, c_{n_i-1})\big) \,. $$
where $r_i, c_i \in \R$ correspond to vertex $i$'s position along the row axis and the column axis respectively (they are not restricted to being integers).
A contour is generally closed with $(r_0, c_0) = (r_{n_i-1}, c_{n_i-1})$, but it can be open if the corresponding object touches the border of the image (therefore start and end vertices are not the same).

\begin{figure*}[ht]
	\includegraphics[width=\linewidth]{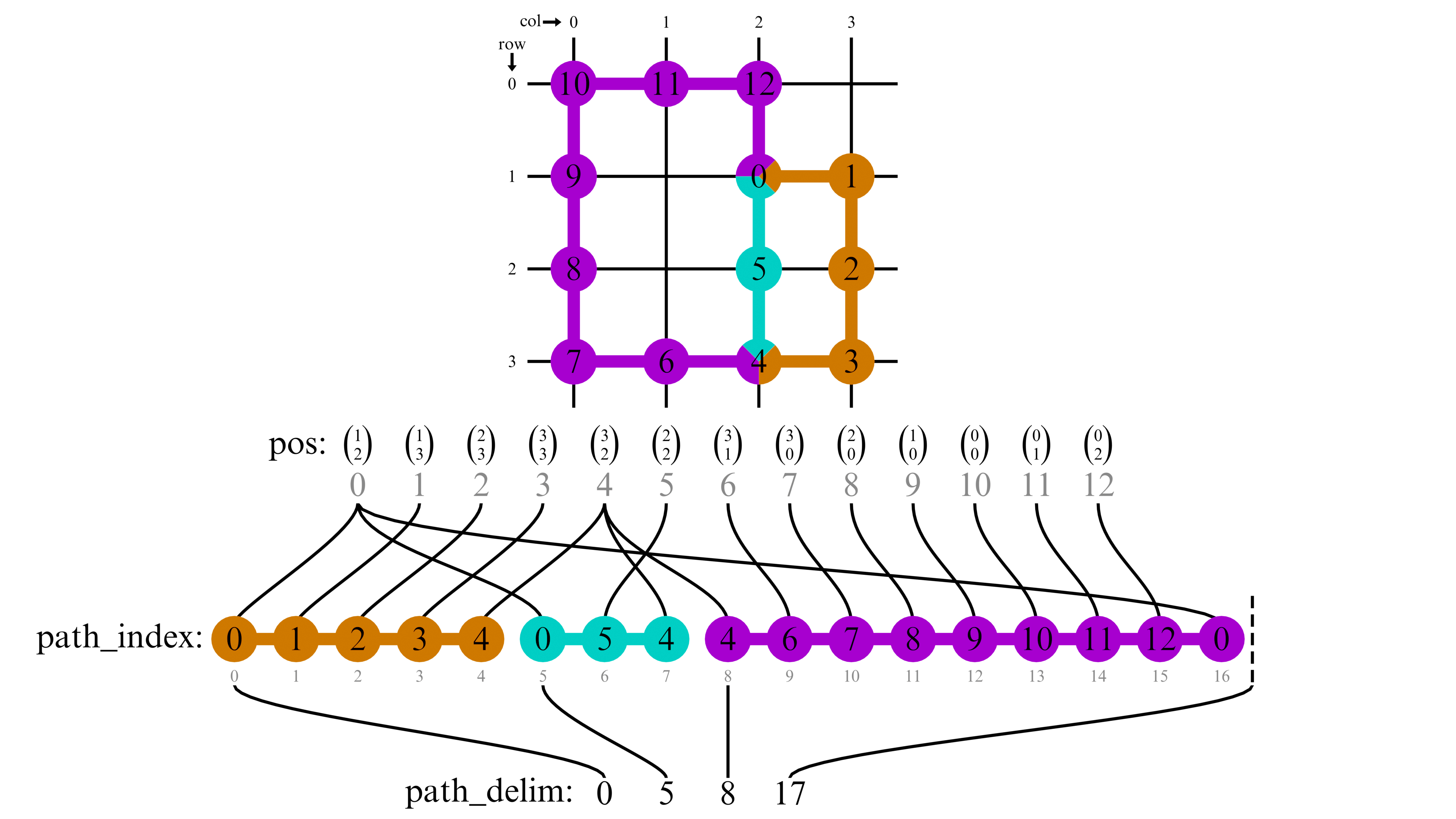}
	\caption{Our data structure of an example skeleton graph. It represents two buildings with a shared wall, necessitating 3 polyline paths. Here nodes $0$ and $4$ are shared among paths and are thus repeated in $path\_index$. We can see $path\_index$ is a concatenation of the node indices in $pos$ of the paths. Finally, $path\_delim$ is used to store the separation indices in $path\_index$ of those concatenated paths. Indices of arrays are in gray.}
	\label{fig:skeleton_data_structure}
\end{figure*}

In the case of extracting buildings with potential adjoining buildings sharing a common wall, we extract the skeleton graph of the predicted edge probability map $y_{\textit{edge}}$. This skeleton graph is a hyper-graph made of nodes connected together by chains of vertices (i.e., polylines) called paths (see Fig.~\ref{fig:polygonization_asm_results} for examples). To obtain this skeleton graph, we first compute the skeleton image using the thinning method~\cite{Zha84} on the binary edge mask (computed by thresholding $y_{\textit{edge}}$ with $\ell = 0.5$). It reduces binary objects to a one-pixel-wide representation. We then use the Skan~\cite{skan} Python library to convert this representation to a graph representation connecting those pixels. The resulting graph is a collection of paths that are polylines connecting junction nodes together. We use an appropriate data structure only involving arrays (named tensors in deep learning frameworks) so that it can be manipulated by the GPU. We show in~Fig.~\ref{fig:skeleton_data_structure} an infographic of the data structure. A sequence of node coordinates ``$\textit{pos}$'' holds the location of all nodes $i \in [0 \twodots n-1]$ belonging to the skeleton:
$$\textit{pos} = \big( (r_0, c_0), (r_1, c_1), ..., (r_{n-1}, c_{n-1}) \big) \,$$
where $n$ is the total number of skeleton pixels and $(r_i, c_i) \in [0 \twodots H-1] \times [0 \twodots W-1]$ correspond to the row number and column number, respectively (of skeleton pixel $i$).
The skeleton graph connects junction nodes through paths, which are polylines made up of connected vertices. These paths are represented by the ``paths'' binary matrix $P_{p,n}$ where element $(i, j)$ is one if node j is in path i. This $P_{p,n}$ is sparse, and, thus, it is more efficient to use the CSR (compressed sparse row) format, which represents a matrix by three (one-dimensional) arrays respectively containing nonzero values, the column indices and the extents of rows. As $P_{p,n}$ is binary we do not need the array containing non-zeros values. The column indices array, which we name ``$\textit{path\_index}$'' holds the column indices of all ``on'' elements:
$$\textit{path\_index} = (j_0, j_1, ..., j_{n - n_{\textit{junctions}} + n_{\textit{degrees sum}} - 1}),$$
where $n_{\textit{junctions}}$ is the total number of junction nodes, $n_{\textit{degrees sum}}$ is the sum of the degrees of all junction nodes and $\forall k \in [0 \twodots n - n_{\textit{junctions}} + n_{\textit{degrees sum}} - 1], j_k \in [0 \twodots n - 1] $. The extents of rows array which we name ``\textit{$path\_delim$}'' holds the starting index of each row (it also contains an extra end element which is the number of non-zeros elements $n$ for easier computation):
$$\textit{path\_delim} = (s_0, s_1, ..., s_p) \,.$$
Thus, in order to get row $i$ of $P_{p,n}$ we need to look up the slice $(s_i, s_{i+1})$ of $path\_index$. In the skeleton graph case, this representation is also easily interpretable. Indices of path nodes are all concatenated in $\textit{path\_index}$ and $\textit{path\_delim}$ is used to separate those concatenated paths.
And finally a sequence of integers ``degrees'' stores for each node the number of nodes connected to it:
$$\textit{degrees} = ( d_0, d_1, ..., d_{n-1}) \,.$$

As a collection of contours is a type of graph, in order to use a common data structure in our algorithm, we also use the skeleton graph representation for the contours $\{\mathcal{C}_i\}$ given by the marching squares algorithm (note we could use other contour detection algorithms for initialization). Each contour is thus an isolated path in the skeleton graph.

In order to fully leverage the parallelization capabilities of GPUs, the largest amount of data should be processed concurrently to increase throughput, i.e., we should aim to use the GPU memory at its maximum capacity. When processing a small image (such as $300\times300$~pixels from the \emph{CrowdAI dataset}), only a small fraction of memory is used. We thus build a batch of such small images to process them at the same time. As an example, on a GTX 1080Ti, we use a polygonization batch size $B=1024$ for processing the \emph{CrowdAI dataset}, which induces a significant speedup. Building a batch of images is very simple: they can be concatenated together along an additional batch dimensions, i.e., $B$    images $I_i \in \R^{3 \times H \times W}$ are grouped in a tensor $\mathbf{I} \in \R^{B \times 3 \times H \times W}$. This is the case for the output segmentation probability maps as well as the frame field. However, it is slightly more complex to build a batch of skeleton graphs because of their varying sizes. Given a collection of skeleton graphs $\{(\textit{pos}_i, \textit{degrees}_i, \textit{path\_index}_i, \textit{path\_delim}_i)\}_{i \in [ \twodots B-1]}$, all $\textit{pos}_i$ and $\textit{degrees}_i$ are concatenated in their first dimension to give batch arrays:
$$\textit{pos}_\textit{batch} = [\textit{pos}_0, \textit{pos}_1, \ldots, \textit{pos}_{B-1}] \,,$$ 
and:
$$\textit{degrees}_\textit{batch} = [\textit{degrees}_0, \textit{degrees}_1, \ldots, \textit{degrees}_{B-1}].$$
All $\textit{path\_index}_i$ need their indices to be shifted by a certain offset:
$$\textit{offset}_i = \sum_{k=0}^{i - 1} |\textit{pos}_k|,$$
with $|\textit{pos}_k|$ the number of points in $\textit{pos}_k$, so that they point to the new locations in $\textit{pos}_\textit{batch}$ and $\textit{degree}_\textit{batch}$. They are then concatenated in their first dimension:
\resizebox{\linewidth}{!}
{%
	$\textit{path\_index}_\textit{batch} = [\textit{path\_index}_0 + \textit{offset}_0, \ldots, \textit{path\_index}_{B-1} + \textit{offset}_{B-1}].$%
}
In a similar manner, we concatenate all $\textit{path\_delim}_i$ into $\textit{path\_delim}_\textit{batch}$ while taking care of adding the appropriate offset.
We then obtain a big batch skeleton graph which is represented in the same way as a single skeleton graph. In order to later recover individual skeleton graphs in the batch, similar to \emph{path\_delim}, we need a \emph{batch\_delim} array that stores the starting index of each individual skeleton graph in the \emph{path\_delim} array (it also contains an extra end element which is the total number of paths in the batch for easier computation). While we apply the optimization on the batched arrays $pos_{batch}$, $path\_index_{batch}$, and so on, for readability we will now refer to them as $pos$, $path\_index$ and so on. Note that in the case of big images (such as $5000\times5000$~pixels from the \emph{Inria dataset}), we set the batch size to 1, as the probability maps, the frame field, and the skeleton graph data structure fills the GPU's memory well.

At this point the data structure is fixed, i.e., it will not change during optimization. Only the values in $pos$ will be modified. This data structure is efficiently manipulated in parallel on the GPU. All the operations needed for the various computations performed in the next sections are run in parallel on the GPU.

We compute other tensors from this minimal data structure which will be useful for computations:
\begin{itemize}
	\item \lstinline{path_pos = pos[path_index]} which expands the positions tensor for each path (junction nodes are thus repeated in $path\_pos$).
	\item A $batch$ tensor which for each node in $pos\_batch$ stores the index $i \in [0 \twodots B-1]$ of the individual skeleton this node belongs to. This is used to easily sample the batched segmentation maps and the batched frame fields at the position of a node.
\end{itemize}

\subsection{Active Skeleton Model}

\begin{figure*}[ht]
	\begin{subfigure}{0.33\textwidth}
		\includegraphics[trim={75 75 75 75},clip,width=\textwidth]{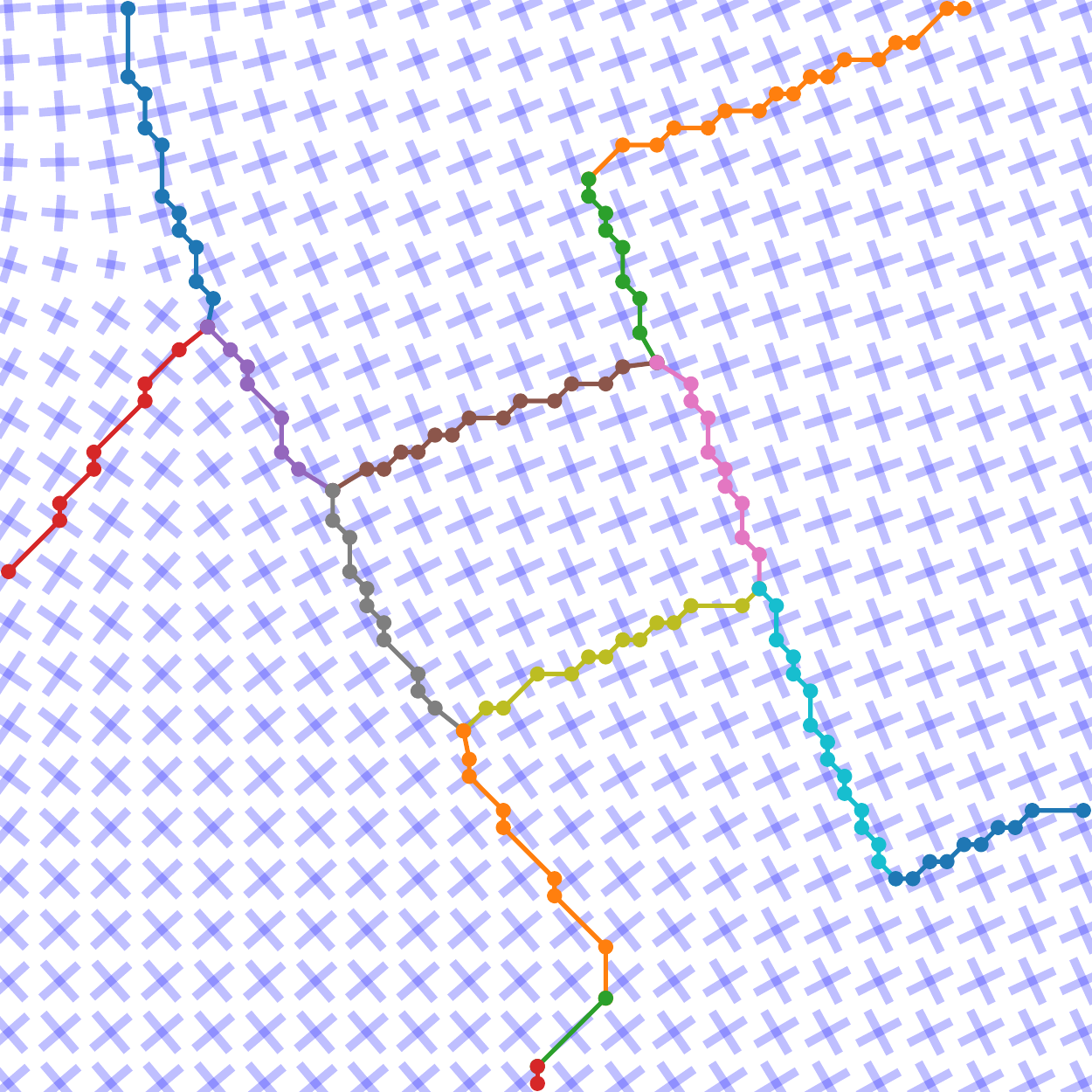}
		\caption{Step 0}
	\end{subfigure}
	\begin{subfigure}{0.33\textwidth}
		\includegraphics[trim={75 75 75 75},clip,width=\textwidth]{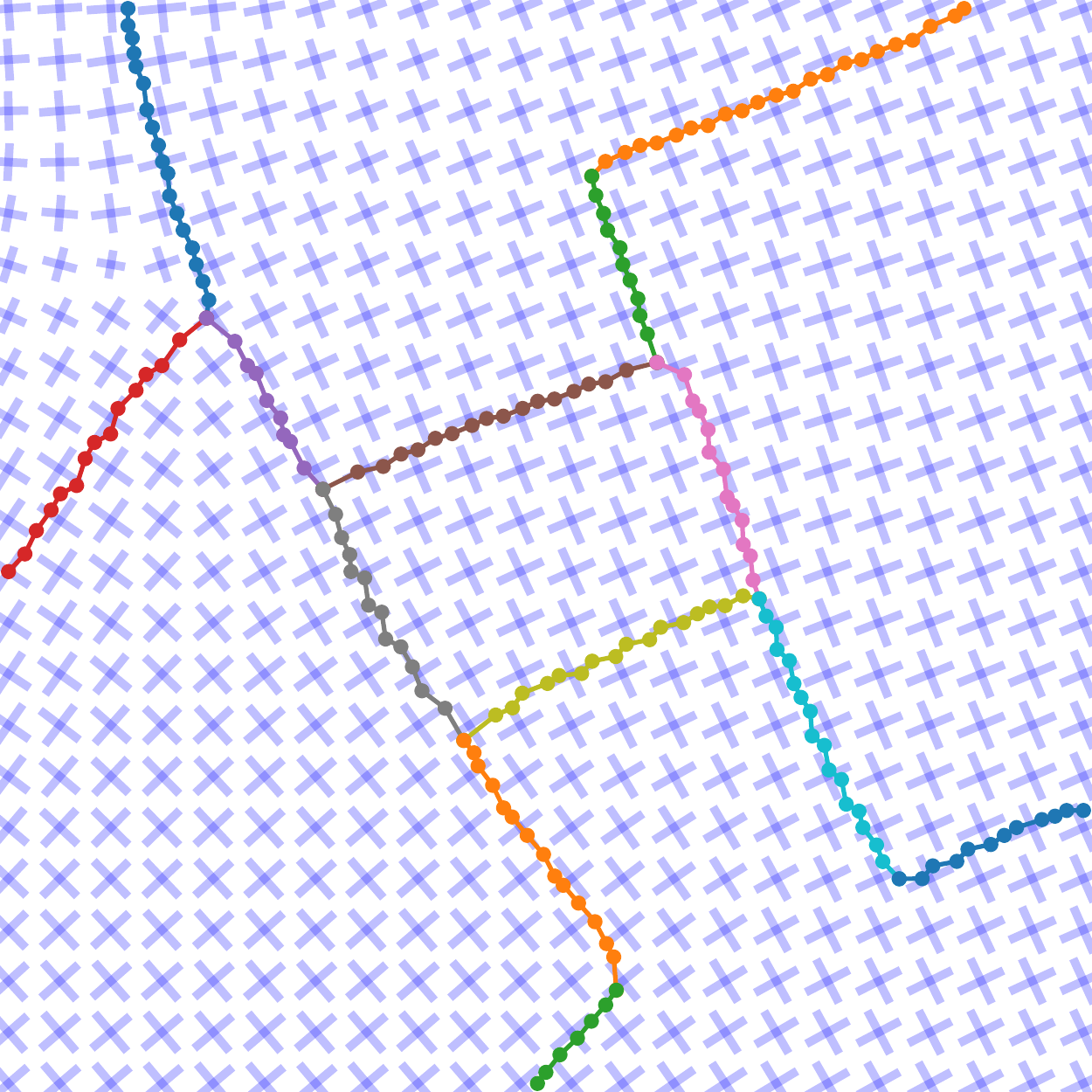}
		\caption{Step 2}
	\end{subfigure}
	\begin{subfigure}{0.33\textwidth}
		\includegraphics[trim={75 75 75 75},clip,width=\textwidth]{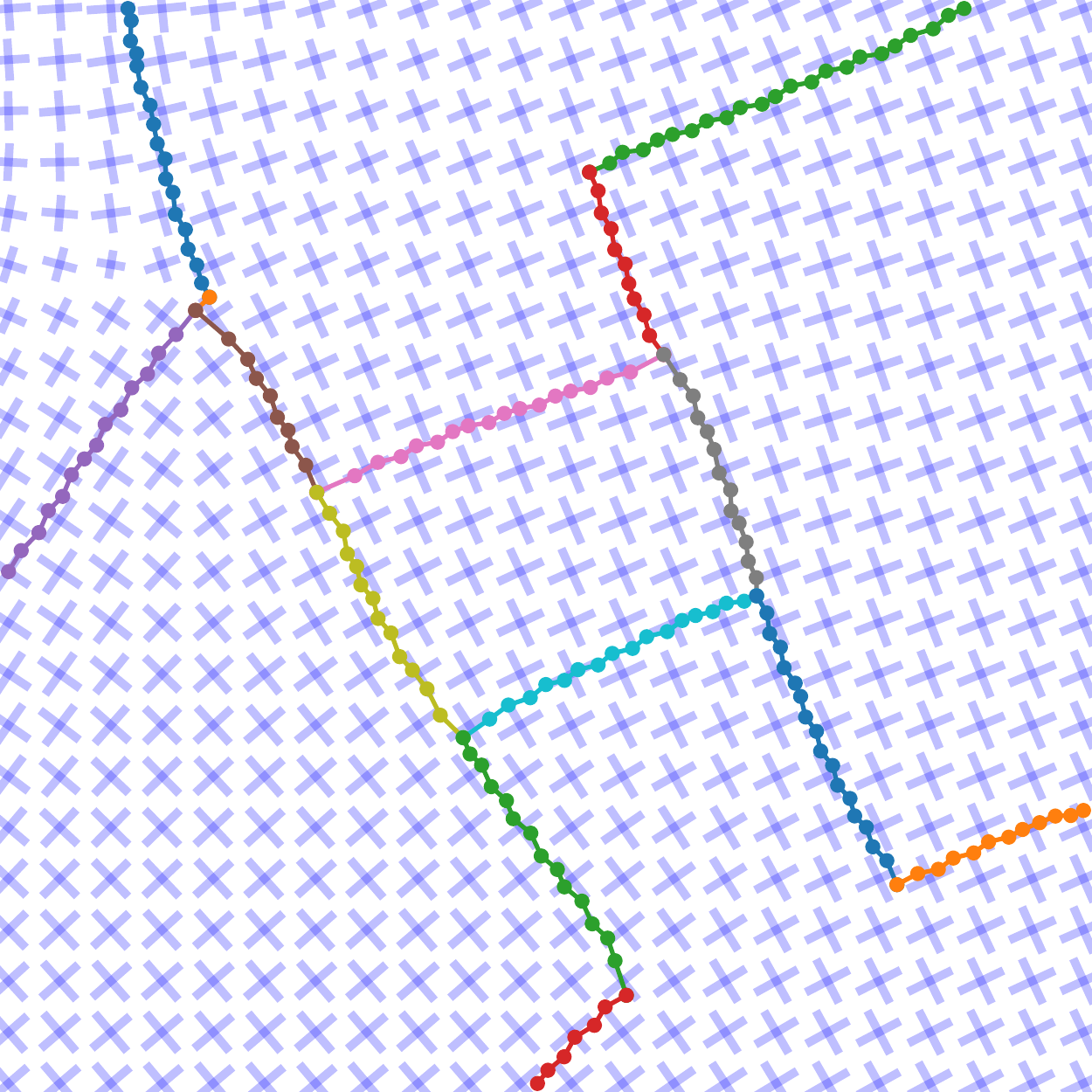}
		\caption{Step 5}
	\end{subfigure}
	
	\begin{subfigure}{0.33\textwidth}
		\includegraphics[trim={75 75 75 75},clip,width=\textwidth]{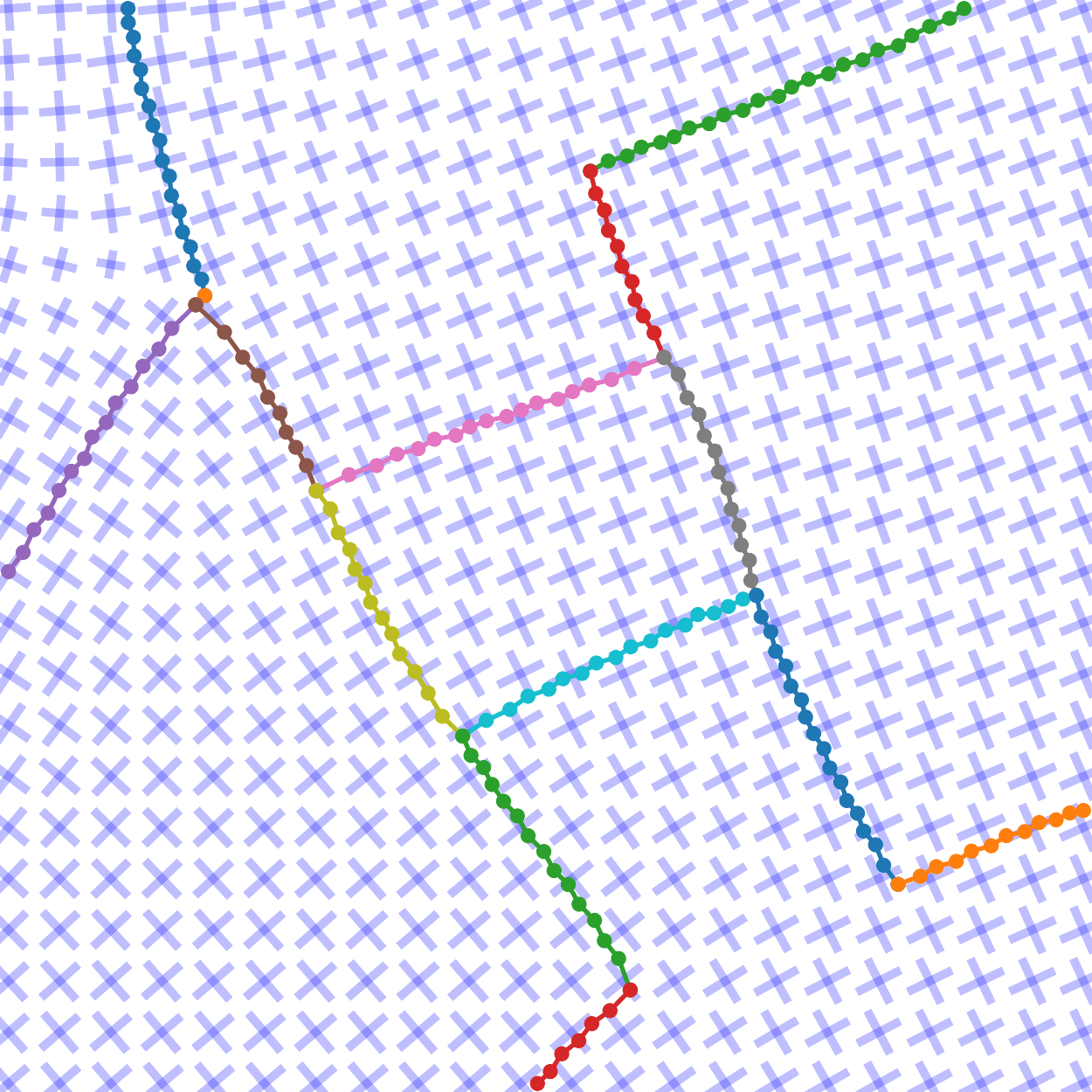}
		\caption{Step 10}
	\end{subfigure}
	\begin{subfigure}{0.33\textwidth}
		\includegraphics[trim={75 75 75 75},clip,width=\textwidth]{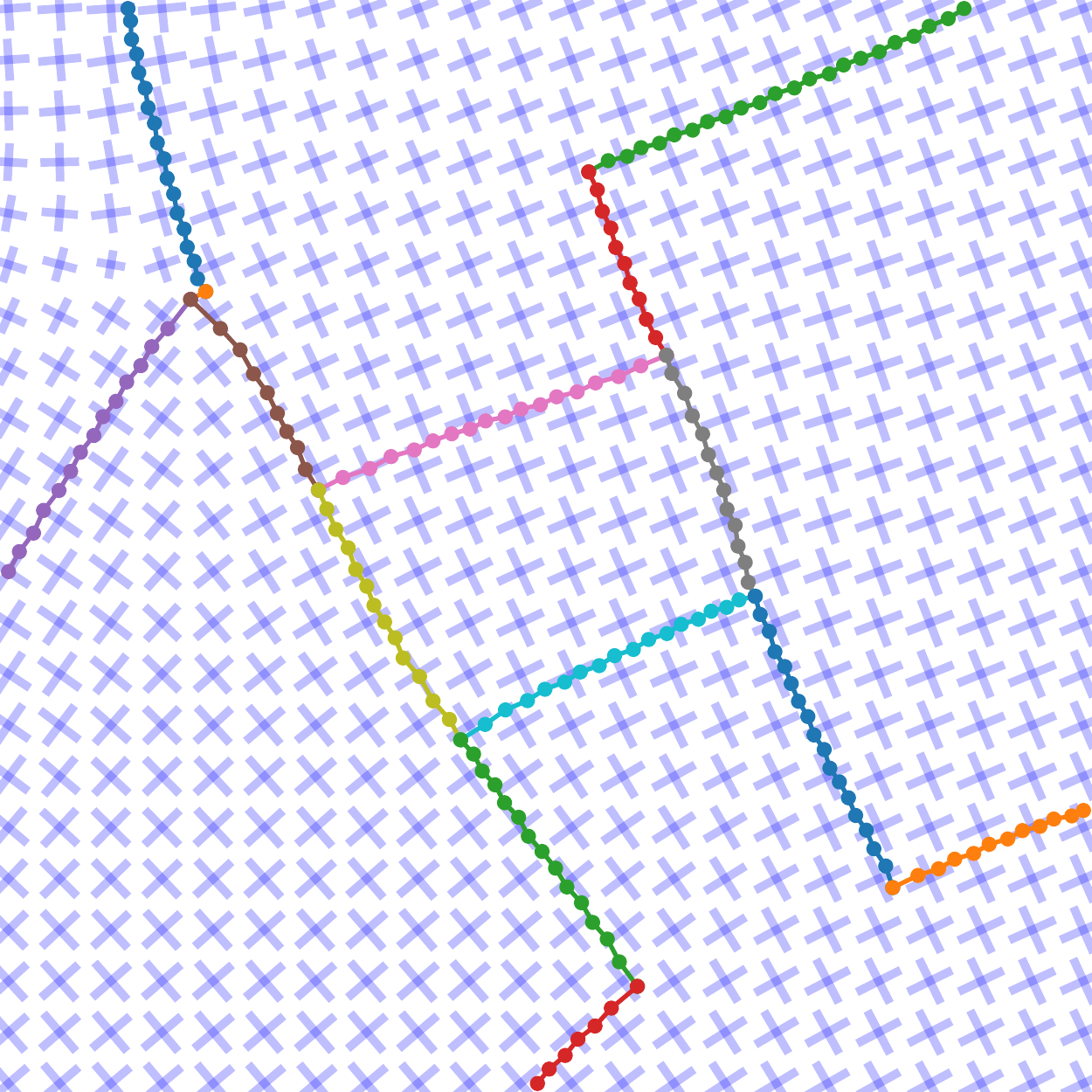}
		\caption{Step 25}
	\end{subfigure}
	\begin{subfigure}{0.33\textwidth}
		\includegraphics[trim={75 75 75 75},clip,width=\textwidth]{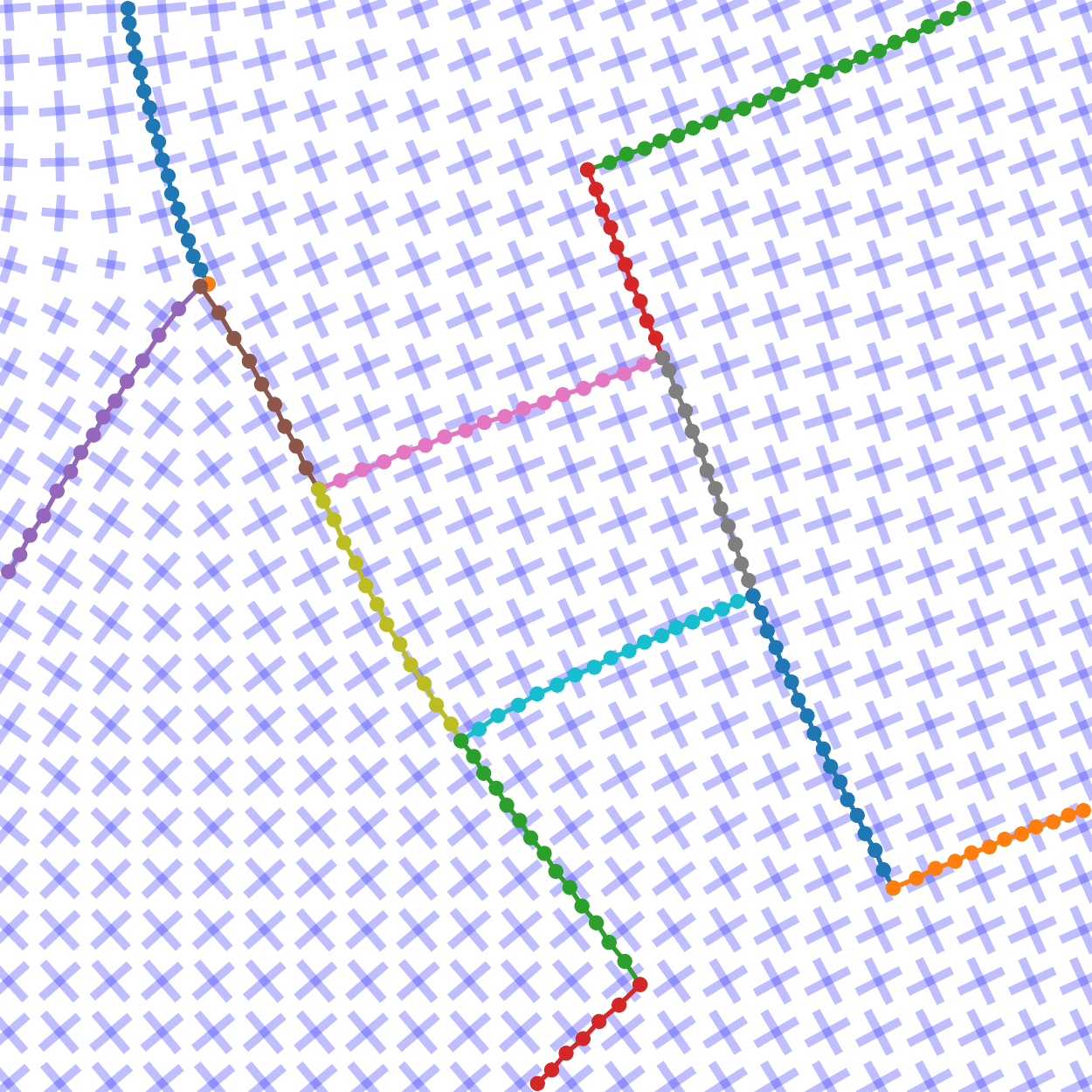}
		\caption{Step 100}
	\end{subfigure}
	\caption{ASM optimization steps (zoomed example). Frame field in blue crosses.}
	\label{fig:polygonization_asm_results}
\end{figure*}

We adapt the formulation of the Active Contours Model (ACM) to an Active Skeleton Model (ASM) in order to optimize our batch skeleton graph. The advantage of using the energy minimization formulation of ACM is to be able to add extra terms if needed (we can imagine adding regularization terms to, e.g., reward 90\degree corners, uniform curvature, and straight walls).

Energy terms will be parameterized by the node positions $\p \in pos$, which are the variables being optimized. The first important energy term is $E_{\textit{probability}}$ which aims to fit the skeleton paths to the contour of the building interior probability map $y_{\textit{int}}({v})$ at a certain probability level $\ell$ (which we set to 0.5 in practice, just like the isovalue used to initialize the contours by marching squares):
$$ E_{\textit{probability}} = \sum_{\p \in pos} (y_{\textit{int}}(\p) - l)^2 \,.$$
The value $y_{\textit{int}}({\p})$ is computed by bilinear interpolation so that gradients can be back-propagated to $\p$. Additionally, $y_{\textit{int}}({\p})$ implicitly entails using the $batch$ array to know which slice in the batch dimension of $y_{\textit{int}} \in \R^{B \times 1 \times H \times W}$ to sample $\p$ from. This will be the case anytime batched image-like tensors are sampled at a point $\p$. In the case of the marching squares initialization, this $E_{\textit{probability}}$ energy is actually zero at the start of optimization, since the initialized contour already is at isovalue $\ell$. For the skeleton graph initialization, paths that trace inner walls between adjoining buildings will not be affected since the gradient is zero in a neighborhood of homogeneous values (i.e., $y_{\textit{int}} = 1$ inside buildings).

The second important energy term is $E_{\textit{frame field align}}$ which aligns each edge of the skeleton paths to the frame field. Edge vectors are computed in parallel as:
$$\e = \lstinline{path_pos[1:] - path_pos[:-1]} \,,$$
while taking care of removing from the energy computation ``hallucinated'' edges between paths (using the $path\_delim$ array). For readability we call $E$ the set of valid edge vectors. For each edge vector $\e \in E$, we refer to its direction as $\e_{dir} = \frac{\e}{\Vert \e \Vert}$. We also refer to its center point as $\e_{\textit{center}} = \frac{1}{2} \lstinline{(path_pos[1:] + path_pos[:-1])} $. The frame field align term is defined as:
$$ E_{\textit{frame field align}} = \sum_{\e \in E} |f({\e_{\textit{dir}}} ; c_0({\e_{\textit{center}}}), c_2({\e_{\textit{center}}}))|^2 \,. $$
This is the term that disambiguates between slanted walls and corners and results in regular-looking contours.

The last important term is the internal energy term $E_{\textit{length}}$ which ensures node distribution along paths remains homogeneous as well as tight:
$$ E_{\textit{length}} = \sum_{\e \in E} |\e|^2 \,.$$

All energy terms are then linearly combined:
\resizebox{\linewidth}{!}
{
	$ E_{\textit{total}} = \lambda_{\textit{probability}} E_{\textit{probability}} + \lambda_{\textit{frame field align}} E_{\textit{frame field align}} + \lambda_{\textit{length}} E_{\textit{length}} \,.$
}
In practice, the final result is robust to different values of coefficients for each of these three energy terms, and we determine them using a small cross-validation set.
The total energy is minimized with the RMSprop~\cite{RMSprop} gradient descent method with a smoothing constant $\gamma = 0.9$ with an initial learning rate of $\eta = 0.1$ which is exponentially decayed. The optimization is run for 300 iterations to ensure convergence. Indeed since the geometry is initialized to lie on building boundaries, it is not expected to move more than a few pixels and the optimization converges quickly. See Fig.~\ref{fig:polygonization_asm_results} for a zoomed example of different stages of the ASM optimization.

\subsection{Corner-aware polygon simplification}

\begin{figure}[ht]
	\centering
	\includegraphics[width=\linewidth]{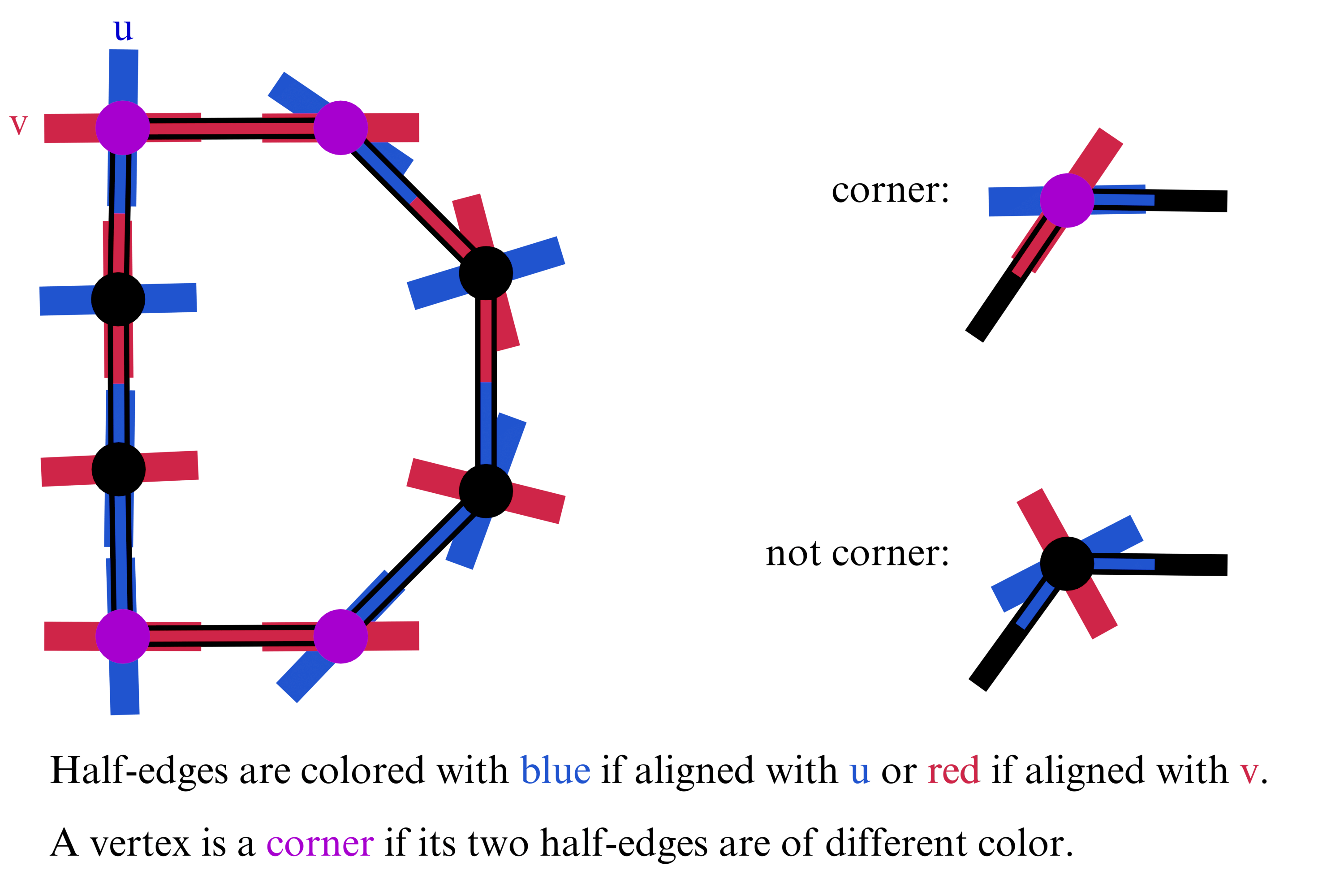}
	\caption{Corner detection using the frame field. For each vertex, the frame field is sampled at that location (with nearest neighbor) and represented by the $\{\pm \textcolor{blue}{u}, \pm \textcolor{red}{v}\}$ vectors.}
	\label{fig:corner_detection}
\end{figure}

We now have a collection of connected polylines that forms a planar skeleton graph. As building corners should not be removed during simplification, only polylines between corners are simplified. For the moment our data structure encodes a collection of polyline paths connecting junction nodes in the skeleton graph. However, a single path can represent multiple walls. It is the case for example of an individual rectangular building: one path describes its contour while it has 4 walls. In order to split paths into sub-paths each representing a single wall we need to detect building corners along a path and add this information to our data structure. This is another reason to use a frame field input, as it implicitly models corners: at a given building corner, there are two tangents of the contour. The frame field learned to align one of $u$ or $-u$ to the first tangent and one of $v$ or $-v$ to the other tangent. Thus when walking along a contour path if the local direction of walking switches from $\pm u$ to $\pm u$ or vice versa, it means we have come across a corner, see Fig.~\ref{fig:corner_detection} for an infographic for corner detection. Specifically for each node $i$ with position $\p = \lstinline{path_pos[i]}$ its preceding and following edge vectors are computed as: $\e_{\textit{prev}} = \lstinline{path_pos[i] - path_pos[i-1]}$ and $\e_{\textit{next}} = \lstinline{path_pos[i+1] - path_pos[i]}$. As the frame field is represented by the coefficients $\{c_0, c_2\}$ at each pixel, we first need to convert it to its $\{u, v\}$ representation with the simple formulas of eq.~\ref{eq:frame_field_convert}.
{
	\footnotesize
	\begin{equation}
		\label{eq:frame_field_convert}
		\left\{\begin{aligned} 
			c_0 &= u^2v^2 \\
			c_2 &= -(u^2 + v^2)
		\end{aligned}\right. \iff 
		\left\{\begin{aligned}
			u^2 &= -\tfrac{1}{2}\left(c_2 + \sqrt{c_2^2 - 4c_0}\right)\\ 
			v^2 &= -\tfrac{1}{2}\left(c_2 - \sqrt{c_2^2 - 4c_0}\right) \,.
		\end{aligned}\right.
	\end{equation}
}
The frame field is sampled at that position $\p$: $u_{\p} = u(\p)$ and $v_{\p} = v(\p)$. Alignment between $\e_{\textit{prev}}, \e_{\textit{next}}$ and $\pm u_{\p}, \pm v_{\p}$ is measured with the absolute scalar product so that it is agnostic to the sign of $u$ and $v$. For example alignment between $\e_{\textit{prev}}$ and $\pm u_{\p}$ is measured by $|\langle \e_{\textit{prev}}, u_{\p} \rangle|$ and if $|\langle \e_{\textit{prev}}, u_{\p} \rangle| < |\langle \e_{\textit{prev}}, v_{\p} \rangle|$ then $\e_{\textit{prev}}$ is aligned to $\pm v$ and not $\pm u$. The same is done for $\e_{\textit{next}}$. Finally if $\e_{\textit{prev}}$ and $\e_{\textit{prev}}$ do not align to the same frame field direction, then node $i$ is a corner. As a summary for corner cases we refer to Table.~\ref{tab:frame_field_corner}.

\begin{table}
	\centering
	\resizebox{\linewidth}{!}{
		\begin{tabular}{ c | c | c }
			\hline
			& $|\langle \e_{\textit{prev}}, u_{\p} \rangle| < |\langle \e_{\textit{prev}}, v_{\p} \rangle|$ & $|\langle \e_{\textit{prev}}, v_{\p} \rangle| < |\langle \e_{\textit{prev}}, u_{\p} \rangle|$ \\
			\hline
			$|\langle \e_{\textit{next}}, u_{\p} \rangle| < |\langle \e_{\textit{next}}, v_{\p} \rangle|$ & False & True \\
			$|\langle \e_{\textit{next}}, v_{\p} \rangle| < |\langle \e_{\textit{next}}, u_{\p} \rangle|$ & True & False \\
			\hline
		\end{tabular}
	}
	\caption{Summary table for deciding if node $i$ with position $\p = \lstinline{path_pos[i]}$ is a corner (True) or not (False).}
	\label{tab:frame_field_corner}
\end{table}

Because the path positions are concatenated together in the $path\_pos$ tensor, some care must be taken for nodes at the extremities of paths (i.e., junction nodes) as they do not have both preceding and following edges. The $path\_delim$ tensor is used to mark those nodes as not corners.
Once corners are detected we obtain a tensor $is\_corner\_index = \{ i \;|\; \text{node } i \text{ is a corner} \}$ which can be used to separate paths into sub-paths each representing a single wall by merging $is\_corner\_index$ with the $path\_delim$ tensor through concatenation and sorting.


Now that each sub-path polyline represents a single wall between two corners, we apply the Ramer-Douglas-Peucker~\cite{ramer-d-p,r-douglas-peucker} simplification algorithm separately on all sub-path polylines. As explained in the related works, the simplification tolerance $\epsi$ represents the maximum Hausdorff distance between the original polyline and the simplified one.

\subsection{Detecting building polygons in planar graph}

To obtain our final output of building polygons, the collection of polylines is polygonized by detecting connecting regions separated by the polylines. A list of polygonal cells that partition the entire image is thus obtained. The last step computes a building probability value for each polygon using the predicted interior probability map and removes low-probability polygons (in practice those that have an average probability less than 50\%).

\section{Experimental setup details}

\subsection{Datasets}

\paragraph{CrowdAI dataset.} The \emph{CrowdAI dataset}~\cite{CrowdAI} originally has 280741 training images, 60317 validation images, and 60697 test images. All images are $300\!\times\!300$~pixels with unknown ground sampling distance, although they are aerial images. As the ground truth annotations of the test set are unreleased because of the challenge, we use the original validation set as our test set and discard the original test images as is commonly done by other methods comparing themselves with that dataset~\cite{PolyMapper,cvpr2020li}. We then use 75\% of the original training images as our initial training set and 25\% for validation. Out final models are then trained on the entire original training set with hyperparameters selected using our validation test.

\paragraph*{Inria dataset.} The \emph{Inria dataset}~\cite{maggiori2017dataset} has 360 aerial images of $5000\times5000$~pixels each with a Ground Sampling Distance of $30$~cm. In total, 10 cities from Europe and the USA are represented, each city having 36 images. Each image is accompanied by its building ground truth mask with an average of a few thousand buildings per image. This dataset provides building ground truth in the form of binary mask images for each image. However, our method requires the ground truth annotations to be in vector format (polygons) so that the ground truth for the frame field can be computed: the tangent angle $\theta_\tau$ used in $L_{\textit{align}}$. We thus build two dataset variants with vector annotations.

The first variant is the \emph{Inria OSM dataset} for which we discard completely the original ground truth masks and instead download annotations from Open Street Map (OSM)~\cite{osm}. Because the OSM annotations are not always aligned, we align them using~\cite{Girard_2019_IGARSS}. We randomly split the images into train (50\%), validation (25\%), and test (25\%) sets. Because the OSM annotations have a lot of missing buildings in certain images, our test results on this dataset are somewhat skewed. Thus, for the test images, we manually select those with few missing buildings in the annotations, giving us 54 test images in total.

The second variant is the \emph{Inria Polygonized dataset} for which we take the original ground truth masks and convert them to polygon format with our polygonization method. In this setting, the input to our network (we used the small U-Net16) is just the binary mask and the output a frame field. In order to train this model, we need a dataset of (binary masks, $\theta_\tau$) pairs. We used the OSM annotations of the \emph{Inria OSM dataset}, which we rasterized to obtain the input binary masks and which we used to compute $\theta_\tau$. After our model finished training, we applied our frame field polygonization method on the original binary masks of the \emph{Inria dataset} and their predicted frame fields. The new \emph{Inria polygonized dataset} is thus made of (RGB image, polygonized annotations) pairs. We thus obtain the same ground truth as the original dataset but in vector format. This allows us to only use the same ground truth data as the other competitors of the Inria Aerial Image Labeling challenge and thus we can directly compare our method to them. Thus we keep the original train and test splits which do not have any cities overlap and tests cross-city generalization (the principal aim of the associated challenge). We then split the original train split into our train (75\%) and validation (25\%) splits.

\paragraph{Private dataset.} The \emph{private dataset} is a large-scale dataset of satellite images built by a company we collaborate with. The images in this dataset were acquired using three types of satellites (Pleiades, WorldView, and GeoEye) over different types of cities (dense, industrial, residential areas, and city centers). We uniformized the image sampling at $50$~cm/pixel spatial resolution, with 3-band RGB images. 57 images of 30 cities across 5 continents are present in the training dataset. The size of images varies from around $2000\!\times\!2000$ pixels to $20000\!\times\!20000$ pixels. The total dataset covers an area spanning around 700~km$^2$. The building outline polygons were manually labeled precisely by an expert. Satellite images are more challenging than aerial images (such as the CrowdAI and Inria images) because they are less clear due to atmospheric effects. This dataset also contains much more varied images compared to CrowdAI and Inria, making up for its smaller size. We pre-process the training images by splitting them into smaller $512\!\times\!512$ pixel patches. We then keep 90\% of patches for training and 10\% for validation.

\subsection{Metrics}

\paragraph{IoU, AP and AR.} The usual metric for the image segmentation task is Intersection over Union (IoU) which computes the overlap between a predicted segmentation and the ground truth annotation. The IoU is then used to compute other metrics such as the MS COCO~\cite{MS_COCO} Average Precision (AP and its variants AP$_{50}$, AP$_{75}$, AP$_S$, AP$_M$, AP$_L$) and Average Recall (AR and its variants AR$_{50}$, AR$_{75}$, AR$_S$, AR$_M$, AR$_L$) evaluation metrics.
Precision and recall are computed for a certain IoU threshold: detections with an IoU above the threshold are counted as true positives whiles others are false positives and ground truth annotations with an IoU below the threshold are false negatives. Each object is also given a score value representing the model's confidence in the detection. In our case, it is the mean value of the interior probability map inside the detection. The Precision-Recall curve can be obtained by varying the score threshold that determines what is counted as a model-predicted positive detection. Average Precision (AP) is the average value of the precision across all recall values and Average Recall (AR) is the maximum recall given a fixed number of detections per image (100 in our case). Finally, the mean Average Precision (mAP) is calculated by taking the mean AP over multiple IoU thresholds (from 0.50 to 0.95 with a step of 0.05). Likewise for the mean Average Recall (mAR). Following MS COCO's convention, we make no distinction between AP and mAP (and likewise AR and mAR) and assume the difference is clear from context. The AP$_{50}$ variant is AP computed with a single IoU threshold of $50$\% (similarly for AP$_{75}$, AR$_{50}$, and AR$_{75}$). The AP$_S$, AP$_M$ and AP$_L$ variants are AP computed for small ($\textit{area} < 32^2$), medium ($32^2 < \textit{area} < 96^2$) and large ($\textit{area} > 96^2$) objects respectively (like-wise for the AR equivalents).

\paragraph{Max tangent angle error.} We introduce a max tangent angle error metric between predicted polygons and the ground truth to capture the regularity of the predicted contours. A max tangent angle scalar error is computed for each predicted contour. Only predicted contours with at least 50\% overlap with the ground truth are selected, so that their measure makes sense. Each predicted contour is first sampled homogeneously with points $\{P_i\}_{i \in [1 \twodots n]}$ (specifically a point is sampled every 0.1~pixel). Then the $P_i$ points are projected to the ground truth, meaning for each $P_i$ we find the closest point $Q_i$ belonging to the ground truth annotation. For both sequences of points $P_i$ and $Q_i$, corresponding normed tangent directions are computed as:
$$T(P_i) = \frac{P_{i+1} - P_i}{\Vert P_{i+1} - P_i \Vert} \;\;\;\text{and}\;\;\; T(Q_i) = \frac{Q_{i+1} - Q_i}{\Vert Q_{i+1} - Q_i \Vert} \,.$$
The angle differences between the two are computed from the scalar product:
$$\Delta\theta_i = cos^{-1}( \langle T(P_i), T(Q_i) \rangle ) \,.$$
Before computing the maximum angle error $\max_{i} \Delta\theta_i$ along the whole contour, some angle errors $\Delta\theta_i$ need to be filtered out as they are invalid. Angle error invalidity is due to the projection step. Indeed around ground truth corners, part of the predicted contour will we be squashed to be zero-length for example. Another issue is when $P_i$ and $P_{i+1}$ are projected to two different ground truth polygon sides: the projected edge $P_{i+1} - P_{i}$ does not represent a ground truth tangent anymore. We thus filter out tangents whose projection is stretched more than a factor of 2, i.e., we keep all $\Delta\theta_j, \forall j \in V$ where $V = \{j \;|\; j \in [1 \twodots n] , \frac{1}{2} < \frac{\Vert Q_{i+1} - Q_i \Vert}{\Vert P_{i+1} - P_i \Vert} < 2 \} $. The final max tangent angle error for that contour is then:
$$E_{\textit{max tangent angle}} = \max_{j \in V} \Delta\theta_j \,.$$
As each contour gives a scalar error, we aggregate all the errors for a certain dataset by averaging this max tangent angle error metric.

%

\section{Additional results}

\subsection{CrowdAI dataset}

\subsubsection{Complexity vs. fidelity}

For the polygon complexity/fidelity trade-off ablation study we plot the AP and AR scores for difference simplification tolerance values $\epsi$ on the \emph{CrowdAI dataset}.

\begin{figure*}[ht]
	\centering
	\begin{subfigure}{\textwidth}
		\centering
		\rotatebox[origin=l]{90}{Simple poly.}
		\includegraphics[width=0.23\textwidth]{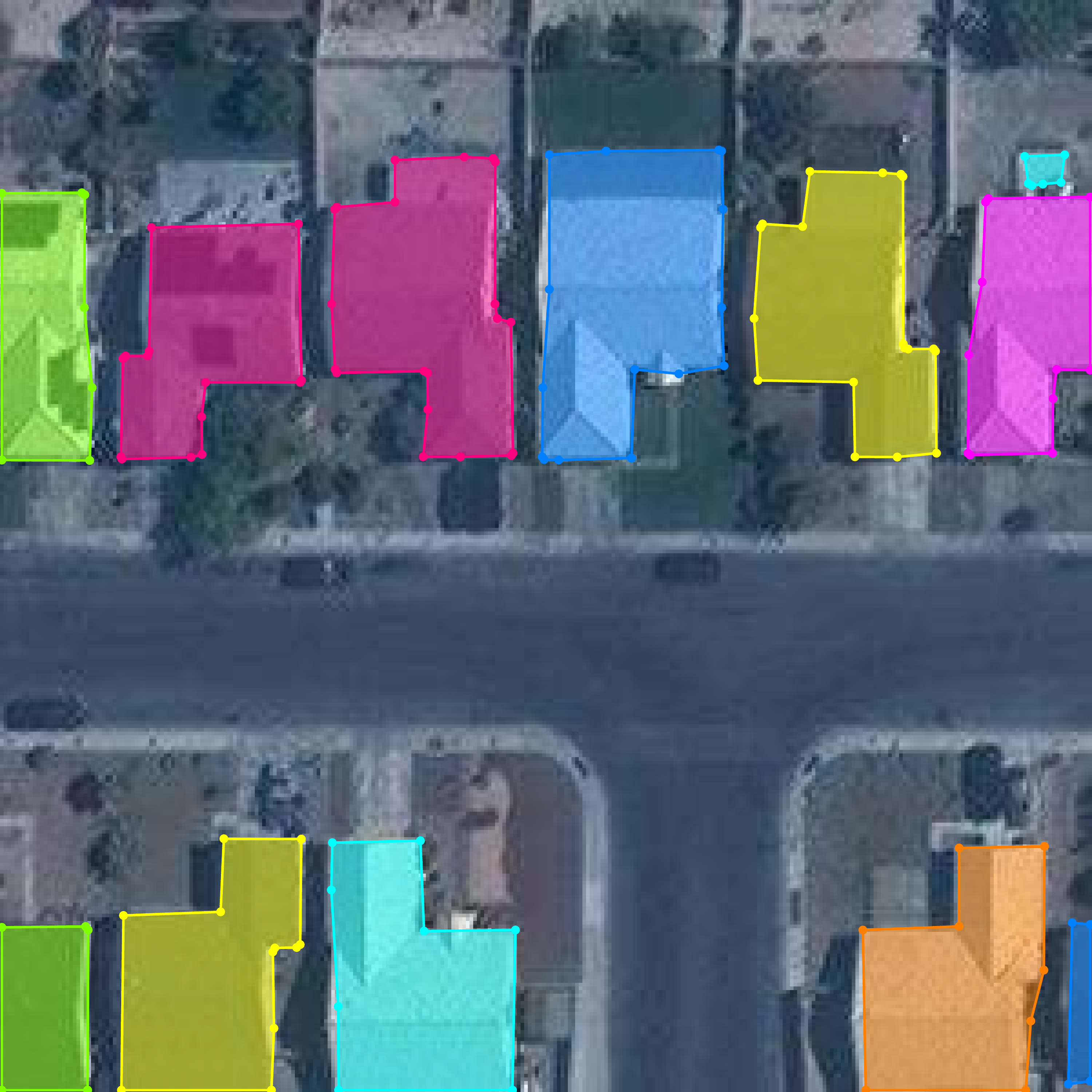}
		\includegraphics[width=0.23\textwidth]{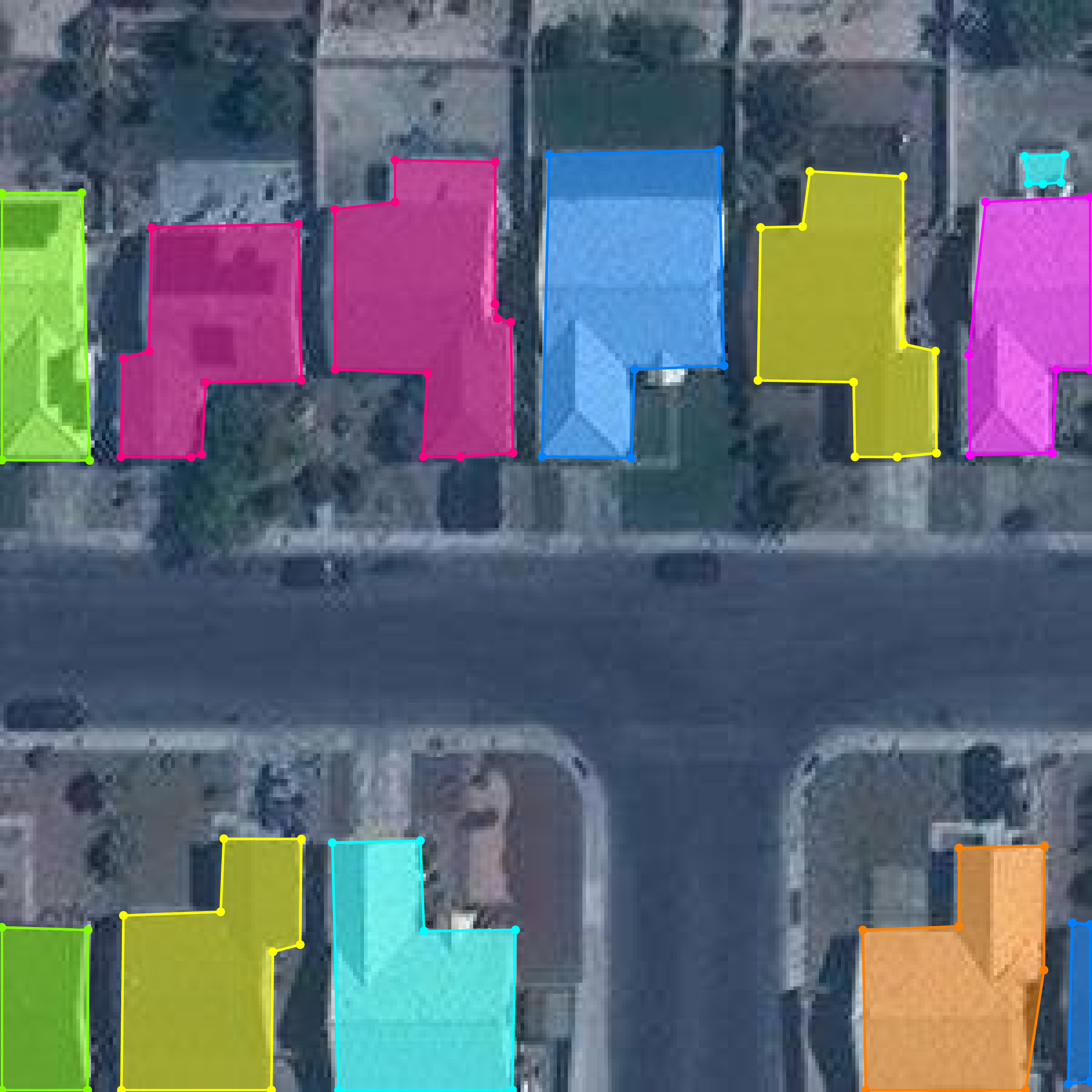}
		\includegraphics[width=0.23\textwidth]{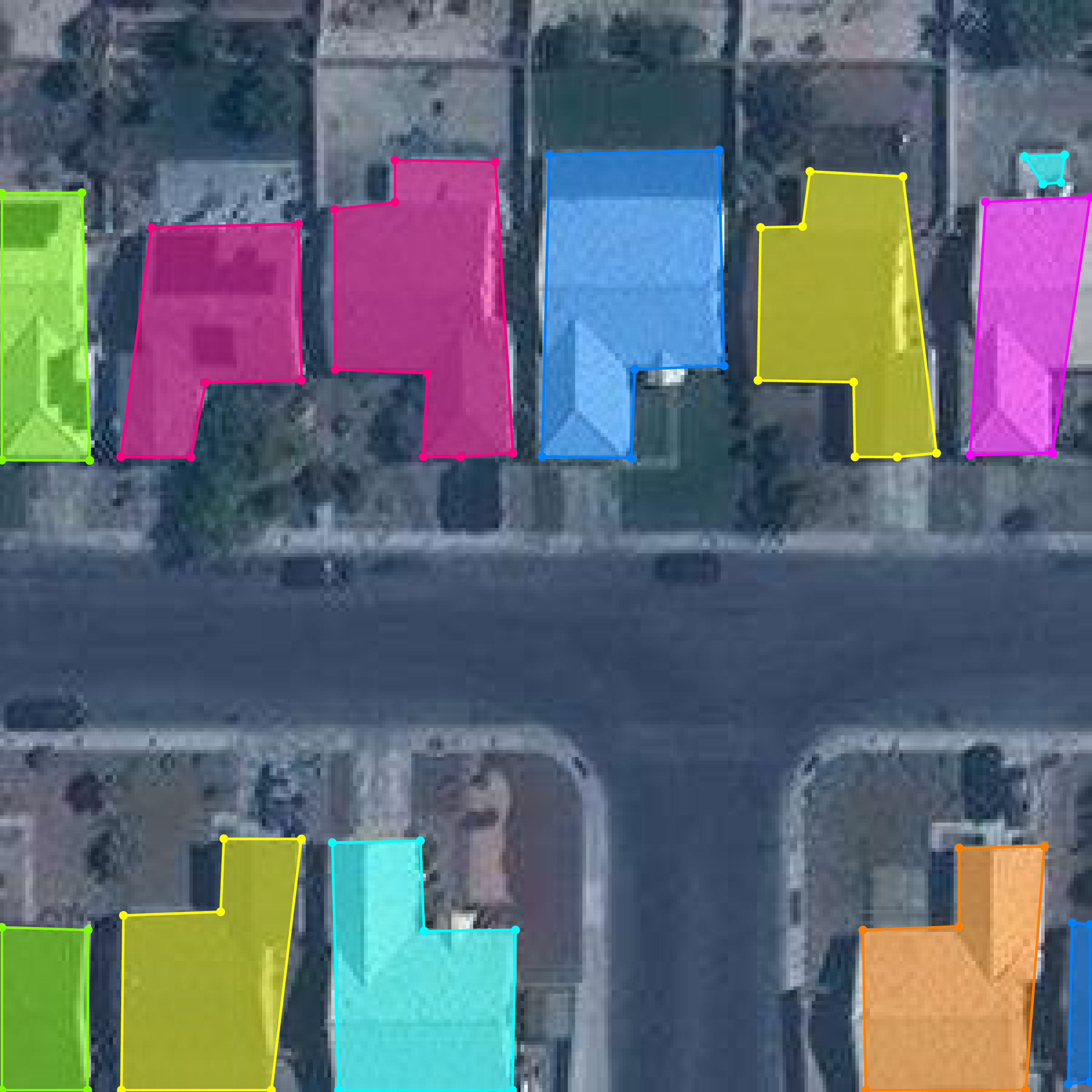}
		\includegraphics[width=0.23\textwidth]{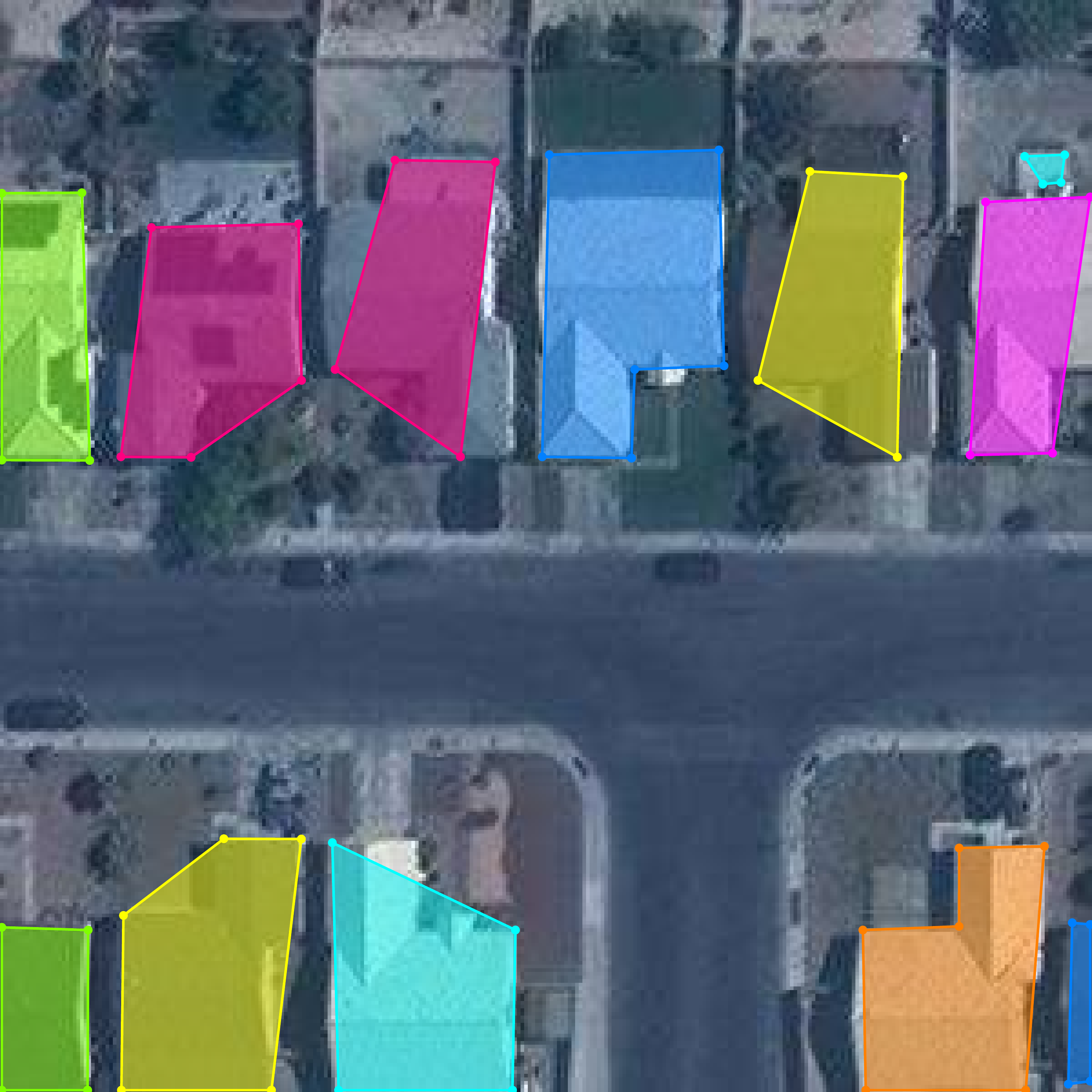}
		
		\rotatebox[origin=l]{90}{\textbf{Ours} \textcolor{white}{p}}
		\includegraphics[width=0.23\textwidth]{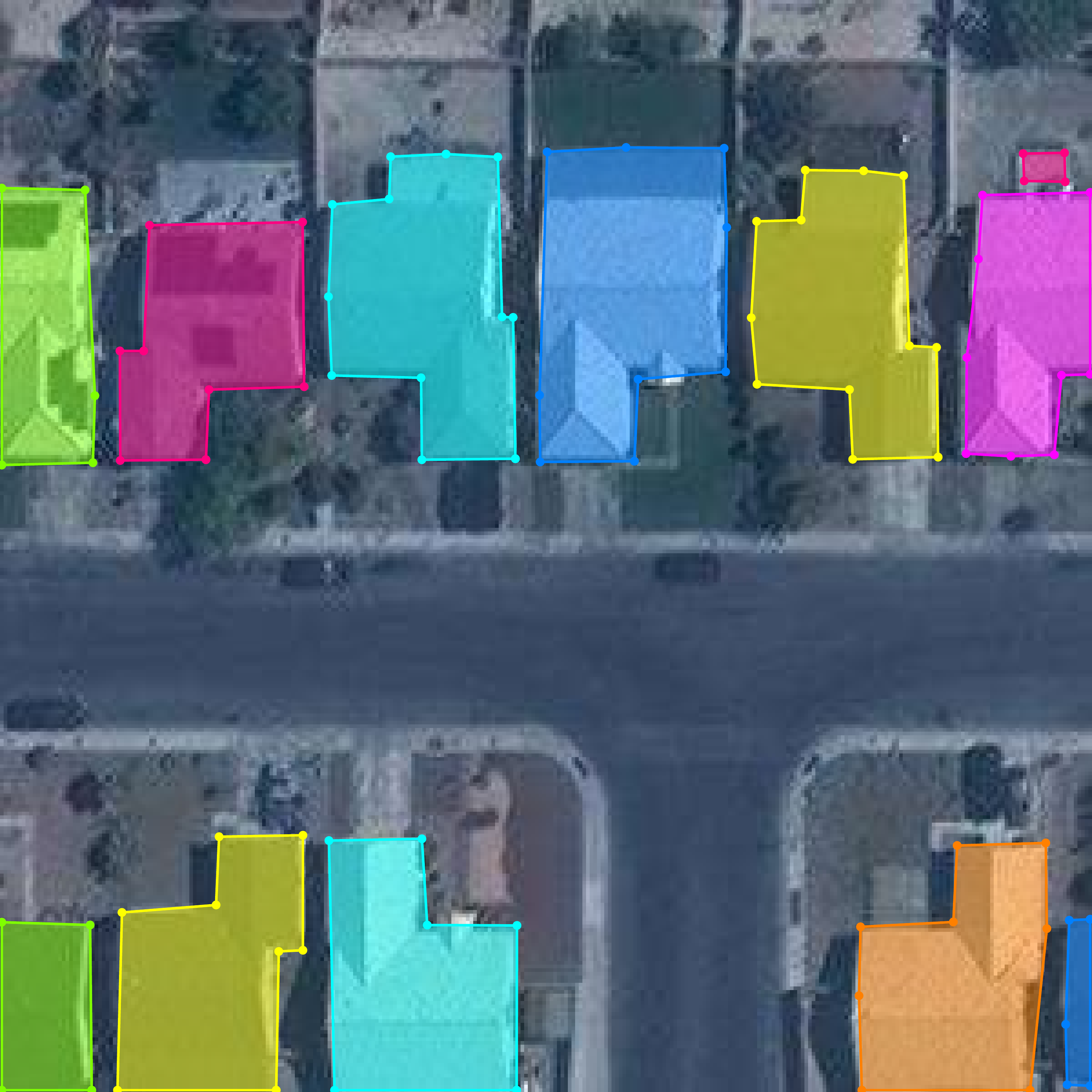}
		\includegraphics[width=0.23\textwidth]{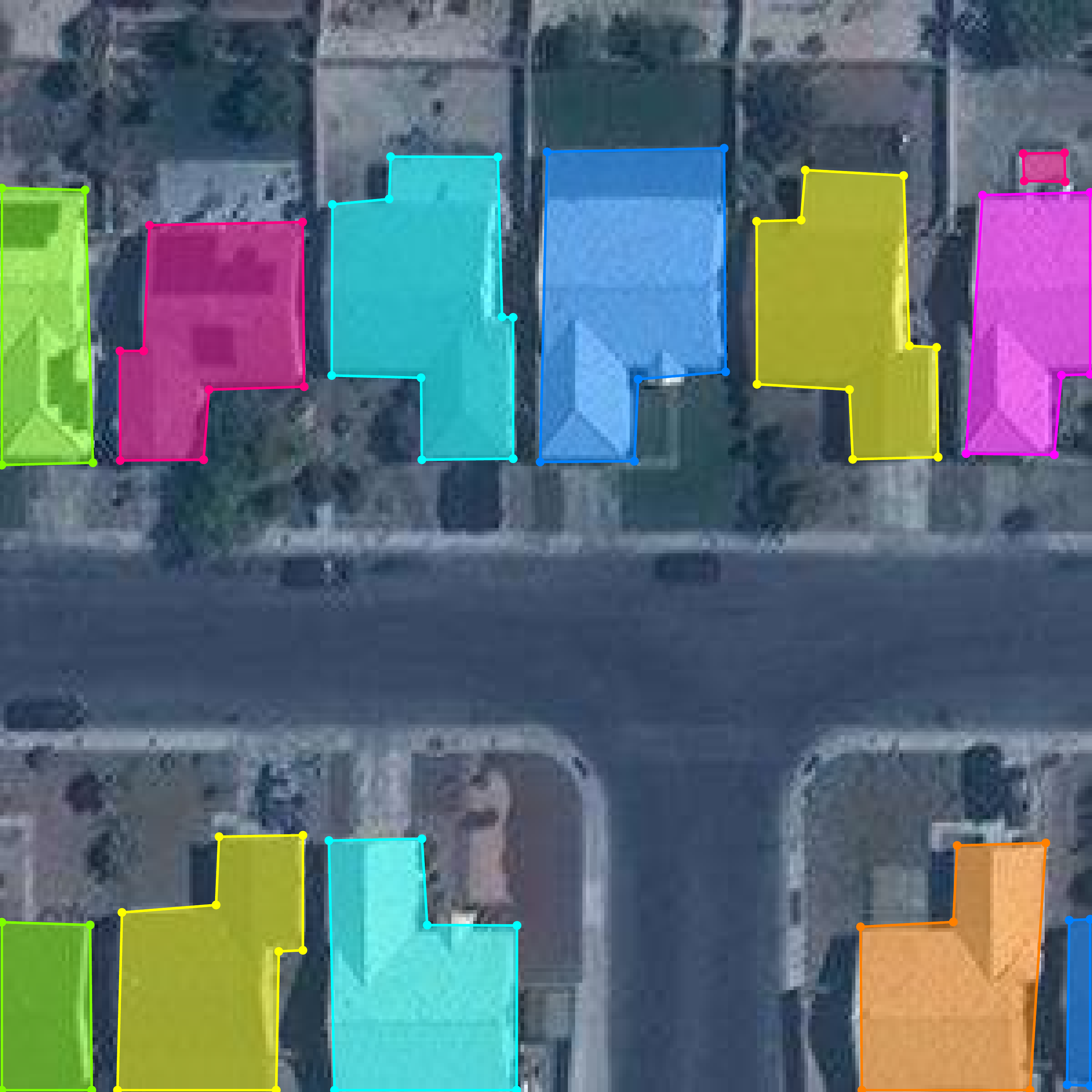}
		\includegraphics[width=0.23\textwidth]{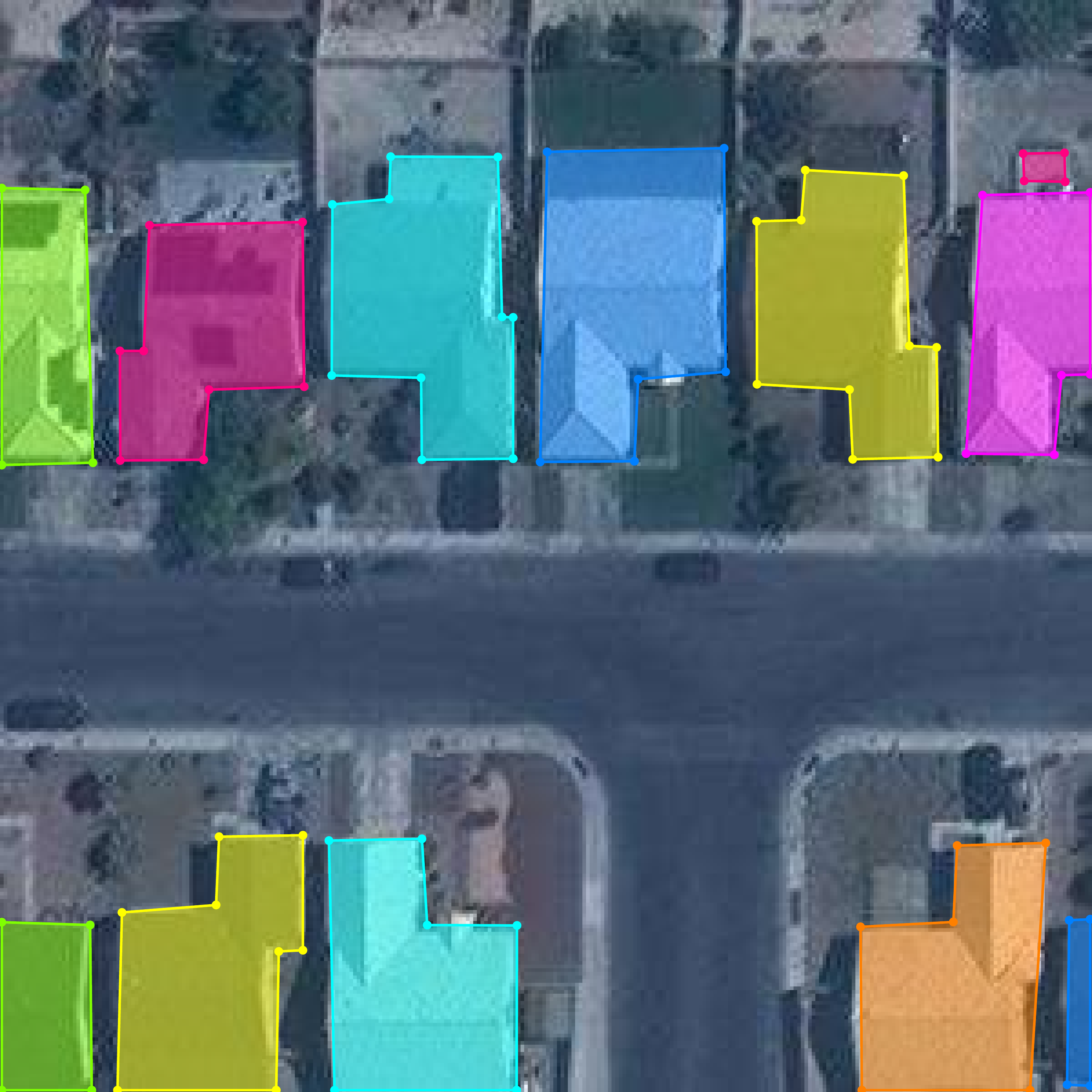}
		\includegraphics[width=0.23\textwidth]{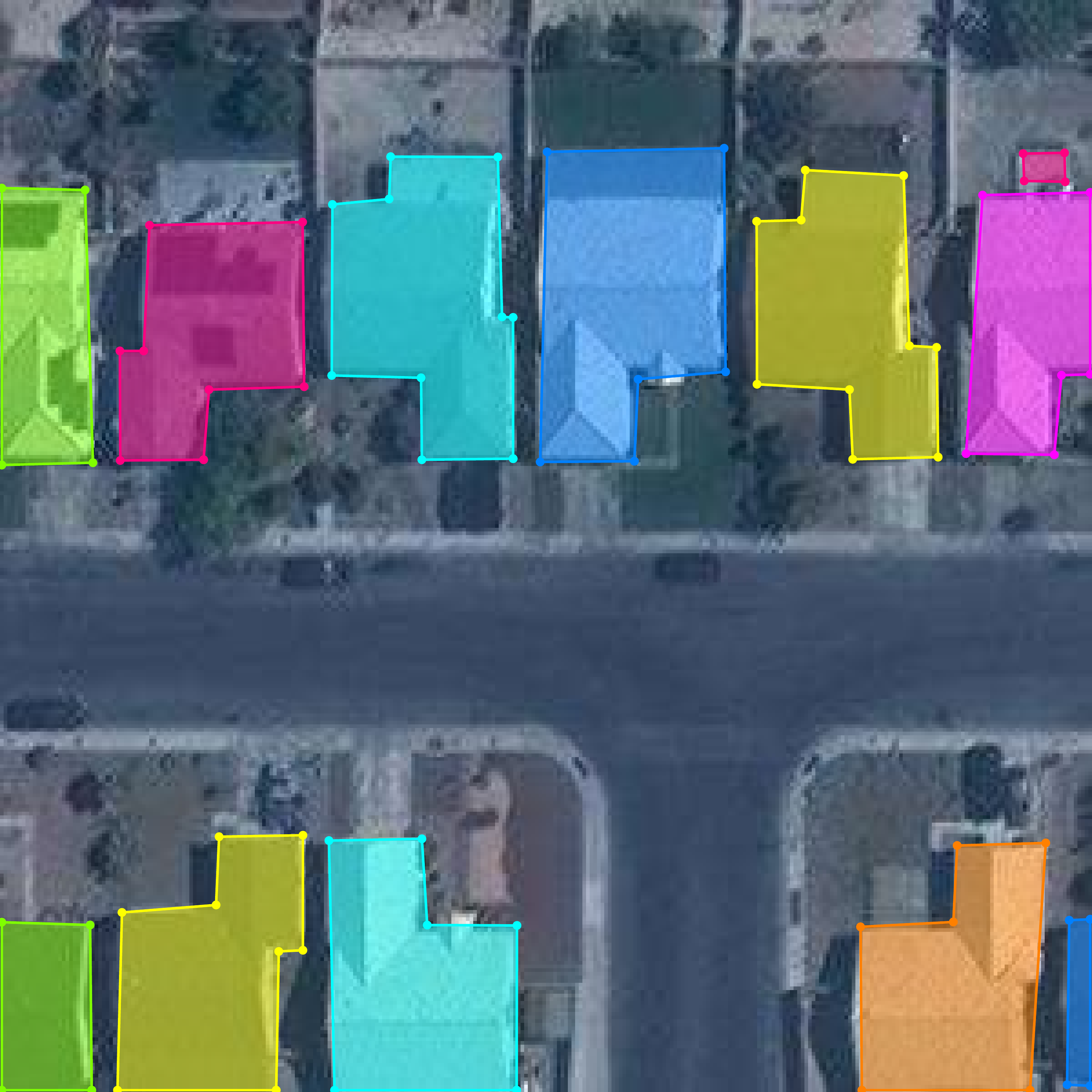}
		\caption{Effect of increasing the simplification tolerance value $\epsi$ from $0.5$~px (left), then $2$~px, then $8$~px and $16$~px (right).}
		\label{fig:results_poly_simple_vs_ours_tolerance:viz}
	\end{subfigure}
	\begin{subfigure}{.45\textwidth}
		\centering
		\includegraphics[width=\linewidth]{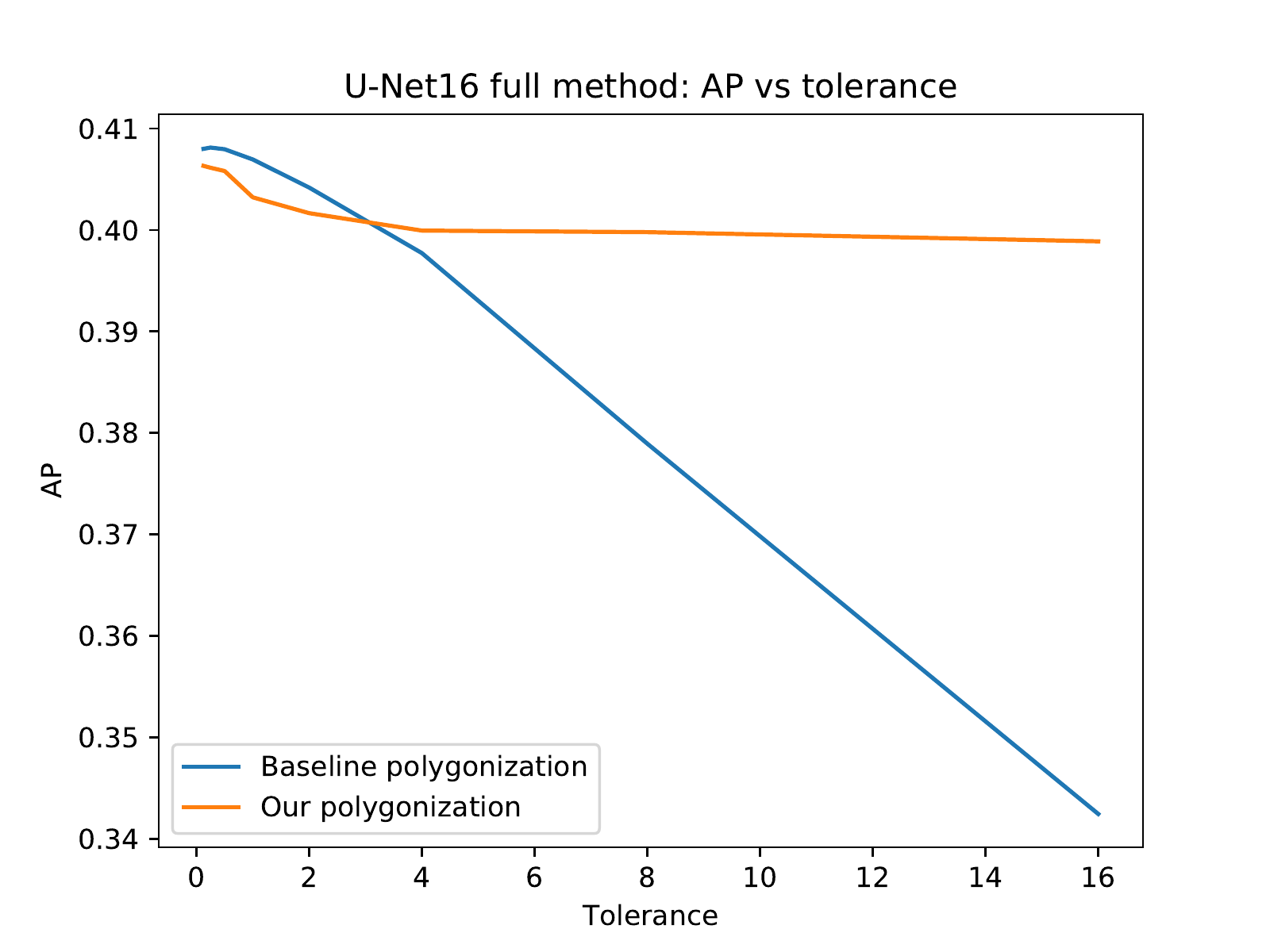}
		\caption{AP for both our corner-aware method and the simple polygonization for various tolerance value $\epsi$.}
	\end{subfigure}
	\hfill
	\begin{subfigure}{.45\textwidth}
		\centering
		\includegraphics[width=\linewidth]{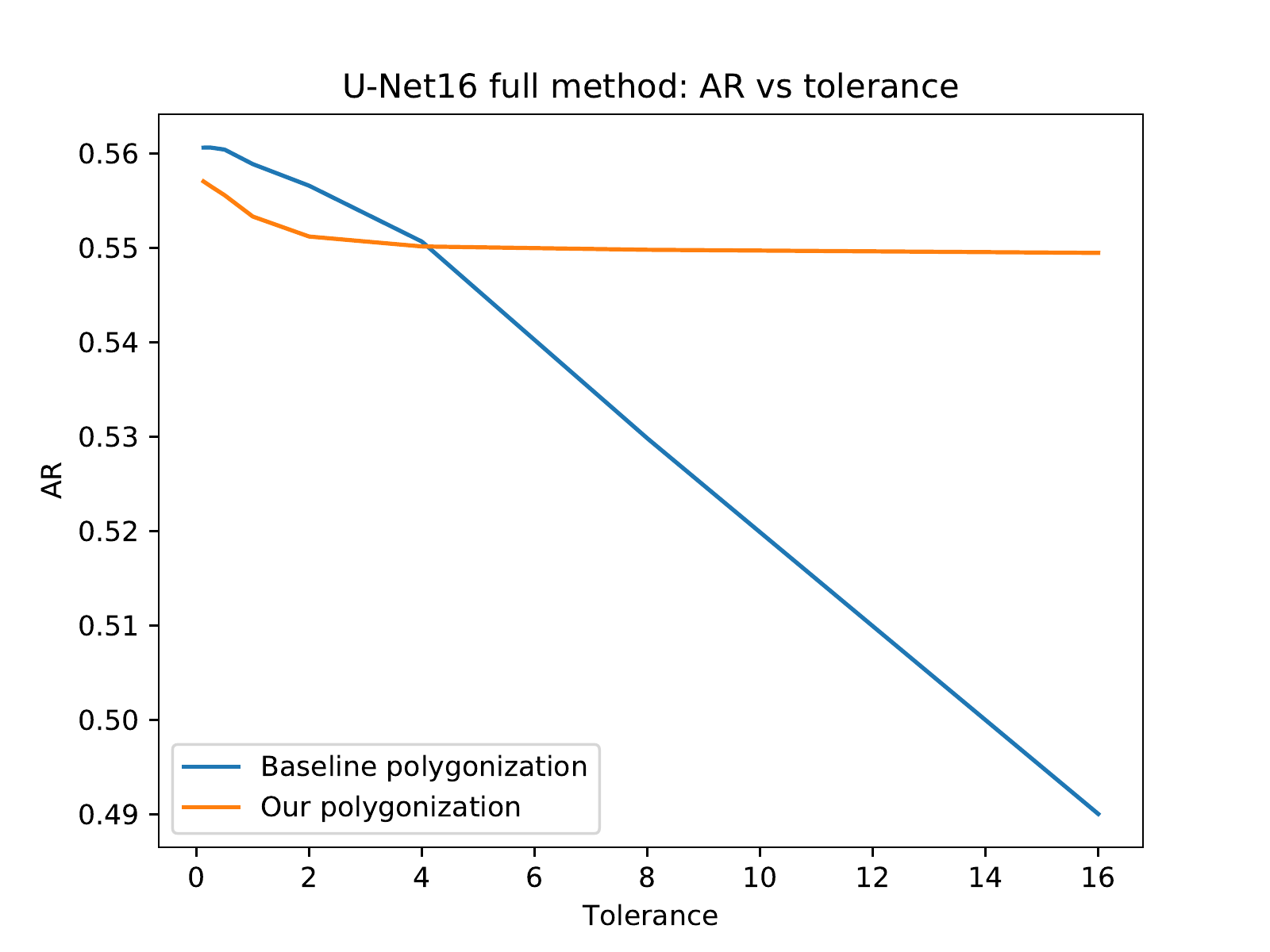}
		\caption{AR for both our corner-aware method and the simple polygonization for various tolerance value $\epsi$.}
	\end{subfigure}
	\caption{Comparison between the baseline simplification algorithm with our corner-aware one. Both take the same classification map as input, but the baseline does not use the frame field. The corner-aware simplification guarantees that no corners will be simplified, regardless of the tolerance value $\epsi$.}
	\label{fig:results_poly_simple_vs_ours_tolerance}
\end{figure*}

We perform an analysis of the polygonization complexity/fidelity trade-off by changing the tolerance value $\epsi$  of the baseline simplification method and our corner-aware method. Fig.~\ref{fig:results_poly_simple_vs_ours_tolerance:viz} shows that preventing the removal of building corners ensures key points of the contours and the global shape of the building remain intact even with extreme simplification tolerance values. We also plot the AP and AR values of both methods while increasing the tolerance value $\epsi$  in Fig.~\ref{fig:results_poly_simple_vs_ours_tolerance}. As expected the score of our method does not drop, unlike the simple polygonization method. 

Our polygonization method allows the complexity-to-fidelity ratio to be tuned with the easy-to-interpret tolerance value $\epsi$ of the Ramer-Douglas-Peucker algorithm, unlike ASIP~\cite{cvpr2020li}, which uses a non-intuitive parameter $\lambda$ to balance complexity and fidelity energies during polygonal partition optimization. Finally, PolyMapper~\cite{PolyMapper} does not have the ability to tune the complexity-to-fidelity ratio.


\subsubsection{Ablation study}

\begin{table}
	\centering
	\resizebox{\linewidth}{!}{
		\begin{tabular}{ l | c }
			\hline
			\textbf{Method}                                                                            & \textbf{Mean max angle error} $\downarrow$ \\ 
			\hline
			UResNet101 (no field), simple poly.                                                        & 51.9\degree \\
			UResNet101 (with field), simple poly.                                                      & 45.1\degree \\
			U-Net variant~\cite{open_solution_crowdai}, ASIP poly.~\cite{cvpr2020li}                   & 44.0\degree \\
			U-Net variant~\cite{open_solution_crowdai}, UResNet101 \textbf{our} poly.                  & 36.6\degree \\
			Zorzi et al.~\cite{zorzi2020machinelearned} poly.                                          & 36.8\degree \\
			UResNet101 (no $L_{\textit{smooth}}$), \textbf{our} poly.                                  & 33.6\degree \\
			UResNet101 (no $L_{\textit{int align}}$ and $L_{\textit{edge align}}$), \textbf{our} poly. & 33.5\degree \\
			UResNet101 (no $L_{\textit{int edge}}$), \textbf{our} poly.                                & 33.4\degree \\
			UResNet101 (no $L_{\textit{align90}}$), \textbf{our} poly.                                 & 33.2\degree \\
			PolyMapper~\cite{PolyMapper}                                                               & 33.1\degree \\
			UResNet101 (with field), \textbf{our} poly.                                                & 31.9\degree \\
			\hline
		\end{tabular}
	}
	\caption{Mean \emph{max tangent angle errors} over all the original validation polygons of the \emph{CrowdAI dataset}~\cite{CrowdAI}.}
	\label{tab:max_tangent_angle_error_all_results}
\end{table}

\begin{table*}
	\centering
	\resizebox{\textwidth}{!}{
		\begin{tabular}{ l | c c c c c c | c c c c c c }
			\hline
			\textbf{Method} & $AP$ $\uparrow$ & $AP_{50}$ $\uparrow$ & $AP_{75}$ $\uparrow$ & $AP_S$ $\uparrow$ & $AP_M$ $\uparrow$ & $AP_L$ $\uparrow$ & $AR$ $\uparrow$ & $AR_{50}$ $\uparrow$ & $AR_{75}$ $\uparrow$ & $AR_S$ $\uparrow$ & $AR_M$ $\uparrow$ & $AR_L$ $\uparrow$\\
			\hline
			U-Net16 (no field), mask                 & 50.9 & 74.3 & 59.5 & 24.5 & 65.6 & 66.3 & 55.9 & 77.9 & 64.7 & 29.8 & 71.2 & 74.6 \\
			U-Net16 (no field), simple poly.         & 50.5 & 76.6 & 59.1 & 22.6 & 66.2 & 69.3 & 54.8 & 78.5 & 63.5 & 26.8 & 71.2 & 75.2 \\
			U-Net16 (no coupling losses), mask       & 53.7 & 77.7 & 62.8 & 25.7 & 69.0 & 68.9 & 57.7 & 79.2 & 66.4 & 31.0 & 73.4 & 74.4 \\
			U-Net16 (with field), mask               & 53.6 & 77.8 & 62.8 & 25.1 & 69.4 & 69.5 & 57.6 & 79.0 & 66.4 & 29.7 & 74.1 & 75.2 \\
			U-Net16 (with field), simple poly.       & 49.6 & 73.8 & 58.1 & 21.2 & 65.5 & 67.0 & 53.8 & 75.6 & 62.2 & 25.5 & 70.5 & 72.5 \\
			U-Net16 (with field), \textbf{our} poly. & 50.5 & 76.6 & 59.3 & 20.4 & 67.4 & 69.0 & 55.3 & 78.1 & 64.0 & 25.7 & 72.8 & 75.0 \\
			\hline
			DeepLab101 (with field)     & 54.9 & 78.1 & 64.9 & 25.6 & 71.2 & 76.8 & 58.7 & 79.8 & 68.1 & 29.5 & 75.8 & 81.6 \\
			DeepLab101 (no field)       & 52.8 & 75.2 & 61.8 & 26.1 & 67.7 & 75.0 & 57.8 & 78.4 & 66.7 & 30.3 & 73.7 & 81.8 \\ 
			\hline
			UResNet101 (no field), mask           & 62.4 & 86.7 & 72.7 & 36.2 & 76.3 & 81.1 & 67.5 & 90.5 & 77.4 & 46.8 & 79.5 & 86.5 \\
			UResNet101 (no field), simple poly.   & 61.1 & 87.4 & 71.2 & 35.1 & 74.5 & 82.3 & 64.7 & 89.4 & 74.1 & 41.7 & 77.9 & 85.7 \\
			
			UResNet101 (with field), mask               & 64.5 & 89.3 & 74.6 & 40.3 & 76.6 & 84.0 & 68.1 & 91.0 & 77.7 & 47.5 & 80.0 & 86.7 \\
			UResNet101 (with field), simple poly.       & 61.7 & 87.7 & 71.5 & 35.8 & 74.9 & 83.0 & 65.4 & 89.9 & 74.6 & 42.6 & 78.6 & 85.9 \\
			UResNet101 (with field), \textbf{our} poly. & 61.3 & 87.5 & 70.6 & 34.0 & 75.1 & 83.1 & 65.0 & 89.4 & 73.9 & 41.2 & 78.7 & 86.0 \\
			
			UResNet101 (no $L_{\textit{align90}}$), mask               & 64.2 & 88.6 & 74.6 & 40.0 & 76.4 & 83.7 & 67.8 & 90.9 & 77.5 & 47.1 & 79.7 & 86.4 \\
			UResNet101 (no $L_{\textit{align90}}$), simple poly.       & 61.4 & 87.7 & 71.4 & 35.4 & 74.5 & 82.7 & 65.0 & 89.7 & 74.4 & 42.1 & 78.2 & 85.6 \\
			UResNet101 (no $L_{\textit{align90}}$), \textbf{our} poly. & 61.1 & 87.5 & 70.6 & 34.1 & 74.9 & 82.8 & 64.7 & 89.3 & 73.8 & 41.2 & 78.4 & 85.6 \\
			
			UResNet101 (no $L_{\textit{int edge}}$), mask               & 63.8 & 88.5 & 74.4 & 39.6 & 75.9 & 83.3 & 67.3 & 90.7 & 77.0 & 46.6 & 79.3 & 86.2 \\
			UResNet101 (no $L_{\textit{int edge}}$), simple poly.       & 61.0 & 87.6 & 70.6 & 35.2 & 74.1 & 82.4 & 64.6 & 89.5 & 74.0 & 41.7 & 77.8 & 85.3 \\
			UResNet101 (no $L_{\textit{int edge}}$), \textbf{our} poly. & 60.9 & 87.4 & 70.5 & 33.7 & 74.4 & 82.5 & 64.4 & 89.1 & 73.4 & 40.7 & 78.1 & 85.4 \\
			
			UResNet101 (no $L_{\textit{int align}}$ and $L_{\textit{edge align}}$), mask               & 64.7 & 89.3 & 74.7 & 40.5 & 76.7 & 84.2 & 68.2 & 91.0 & 77.9 & 47.6 & 80.1 & 86.8 \\
			UResNet101 (no $L_{\textit{int align}}$ and $L_{\textit{edge align}}$), simple poly.       & 61.8 & 87.7 & 71.5 & 35.8 & 74.9 & 83.3 & 65.4 & 89.9 & 74.7 & 42.5 & 78.6 & 86.0 \\
			UResNet101 (no $L_{\textit{int align}}$ and $L_{\textit{edge align}}$), \textbf{our} poly. & 61.5 & 87.5 & 71.3 & 34.2 & 75.2 & 83.4 & 65.0 & 89.5 & 74.0 & 41.3 & 78.8 & 86.1 \\
			
			UResNet101 (no $L_{\textit{smooth}}$), mask               & 64.2 & 88.6 & 74.6 & 40.1 & 76.5 & 83.5 & 67.8 & 90.8 & 77.5 & 47.2 & 79.8 & 86.1 \\
			UResNet101 (no $L_{\textit{smooth}}$), simple poly.       & 61.6 & 87.7 & 71.5 & 35.7 & 74.8 & 82.6 & 65.2 & 89.7 & 74.5 & 42.3 & 78.4 & 85.4 \\
			UResNet101 (no $L_{\textit{smooth}}$), \textbf{our} poly. & 61.3 & 87.5 & 70.7 & 34.1 & 75.0 & 82.7 & 64.8 & 89.3 & 73.9 & 41.1 & 78.6 & 85.5 \\
			\hline
			U-Net variant~\cite{open_solution_crowdai}, UResNet101 \textbf{our} poly. & 67.0 & 92.1 & 75.6 & 42.1 & 84.2 & 92.7 & 73.2 & 93.5 & 81.1 & 48.8 & 87.3 & 95.4 \\
			\hline
			\hline
			Mask R-CNN~\protect\cite{He_2017_ICCV}~\protect\cite{CrowdAIBaseline} & 41.9 & 67.5 & 48.8 & 12.4 & 58.1 & 51.9 & 47.6 & 70.8 & 55.5 & 18.1 & 65.2 & 63.3 \\
			PANet~\protect\cite{liu2018path} & 50.7 & 73.9 & 62.6 & 19.8 & 68.5 & 65.8 & 54.4 & 74.5 & 65.2 & 21.8 & 73.5 & 75.0 \\
			PolyMapper~\protect\cite{PolyMapper} & 55.7 & 86.0 & 65.1 & 30.7 & 68.5 & 58.4 & 62.1 & 88.6 & 71.4 & 39.4 & 75.6 & 75.4 \\
			U-Net variant~\cite{open_solution_crowdai}, ASIP poly.~\cite{cvpr2020li} & 65.8 & 87.6 & 73.4 & 39.3 & 87.0 & 91.9 & 78.7 & 94.3 & 86.1 & 57.2 & 91.2 & 97.6 \\
			\hline
		\end{tabular}
	}
	\caption{AP and AR results on the \emph{CrowdAI dataset}~\protect\cite{CrowdAI} for all polygonization experiments. (with field) refers to models trained with our full frame field learning method. (no field) refers to models trained without any frame field output. ``mask'' refers to the output raster segmentation mask of the network, ``our poly.'' refers to our frame field polygonization method, and ``simple poly.'' refers to the baseline polygonization of marching squares followed by Ramer-Douglas-Peucker simplification.}
	\label{tab:coco_metrics_all_results}
\end{table*}

\begin{figure*}[ht]
	\begin{subfigure}[t]{0.32\textwidth}
		\centering
		\includegraphics[width=\textwidth]{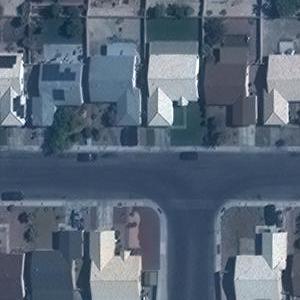}
		\caption{Input}
	\end{subfigure}
	\begin{subfigure}[t]{0.32\textwidth}
		\centering
		\includegraphics[width=\textwidth]{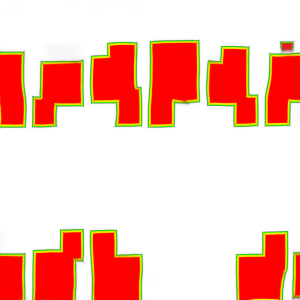}
		\caption{U-Net16 (full): trained with frame field}
	\end{subfigure}
	\begin{subfigure}[t]{0.32\textwidth}
		\centering
		\includegraphics[width=\textwidth]{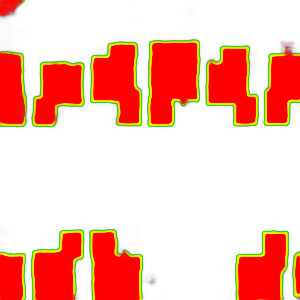}
		\caption{U-Net16 (no field): trained without frame field}
	\end{subfigure}
	
	\begin{subfigure}[t]{0.32\textwidth}
		\centering
		\includegraphics[width=\textwidth]{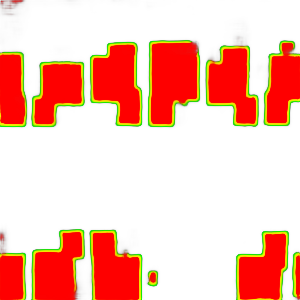}
		\caption{U-Net16 (no coupling losses): trained with frame field but without coupling losses}
		\label{fig:results_ablations_masks:no_coupling_losses}
	\end{subfigure}
	\begin{subfigure}[t]{0.32\textwidth}
		\centering
		\includegraphics[width=\textwidth]{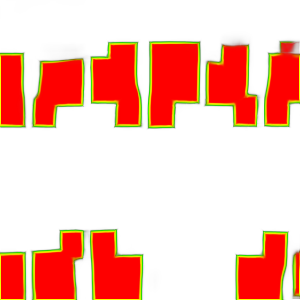}
		\caption{DeepLab101 (full): trained with frame field}
	\end{subfigure}
	\begin{subfigure}[t]{0.32\textwidth}
		\centering
		\includegraphics[width=\textwidth]{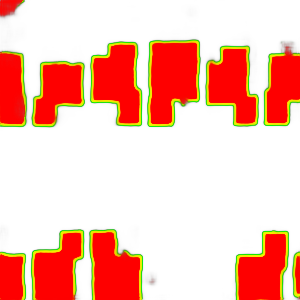}
		\caption{DeepLab101 (no field): trained without frame field}
	\end{subfigure}
	
	\caption{Classification predictions on a test sample for all training ablation studies.}
	\label{fig:results_ablations_masks}
\end{figure*}

We visualize the predicted classification maps from each ablation study for an example test sample in Fig.~\ref{fig:results_ablations_masks}. Both for the U-Net16 and DeepLab101 backbones, the (full) method yields more regular classification maps with sharper corners compared to (no field). Additionally, only learning the frame field with (no coupling losses) is insufficient, as can be seen in Fig.~\ref{fig:results_ablations_masks:no_coupling_losses}.

We observe the effect of only optimizing for IoU when removing coupling losses: we see that it does not impact AP and AR metrics in Table~\ref{tab:coco_metrics_all_results}, while in Fig.~\ref{fig:results_ablations_masks} the (full) segmentations are clearly sharper compared to the (no coupling losses) ones.

In terms of AP and AR metrics, adding a frame field improves the final score (full) compared to (no field) for all backbones: U-Net16, DeepLab101 and UResNet101 (see Table~\ref{tab:coco_metrics_all_results}).

\begin{figure*}[ht]
	\begin{subfigure}[t]{\textwidth}
		\centering
		\includegraphics[trim={0 25cm 25cm 0 0},clip,width=0.5\textwidth]{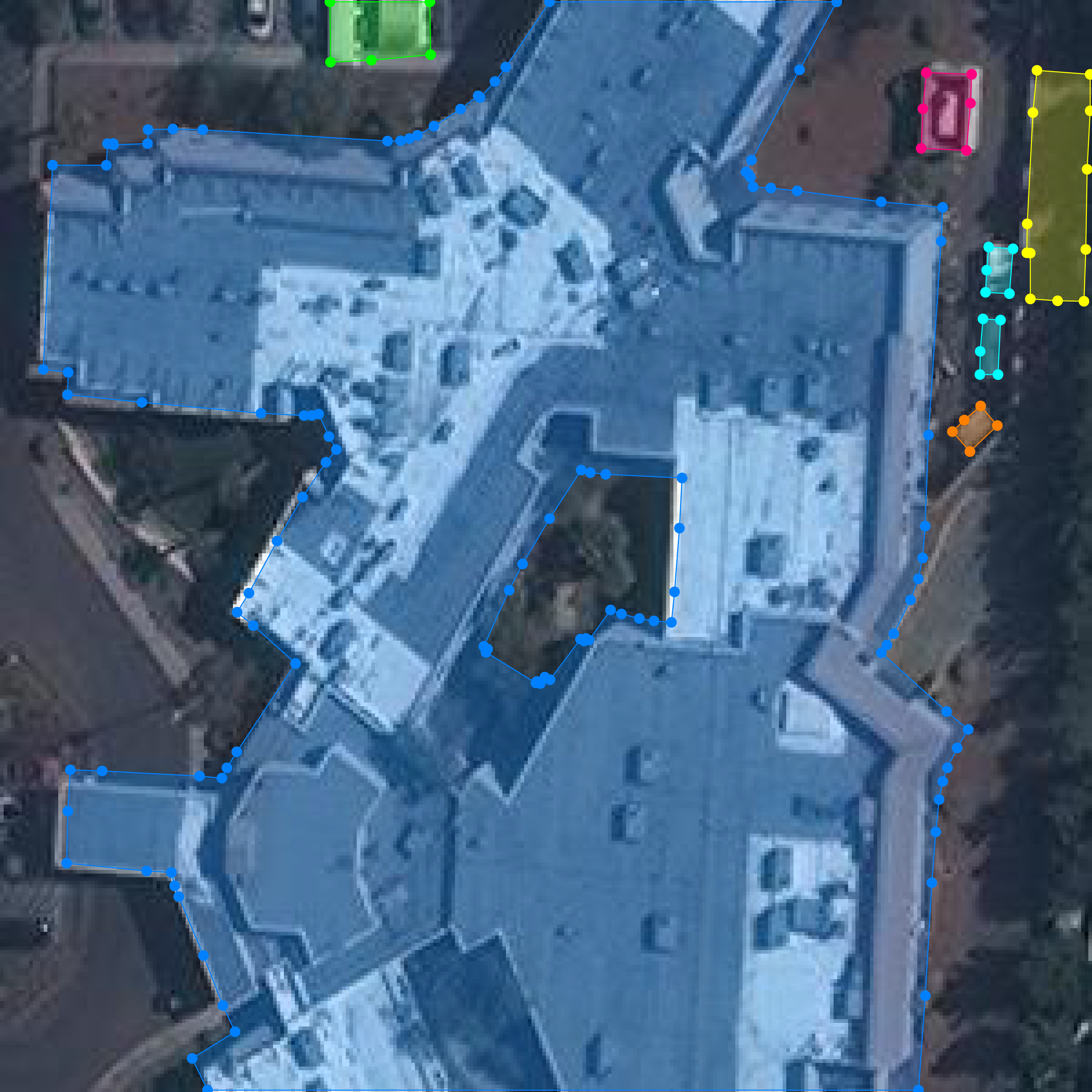}
		\caption{\textbf{Ours}: UResNet101 with frame field learning (full) and frame field polygonization}
		\label{fig:results_ablations_polys:frame_field_learning_polygonization}
	\end{subfigure}
	\begin{subfigure}[t]{0.49\textwidth}
		\centering
		\includegraphics[trim={0 25cm 25cm 0 0},clip,width=\textwidth]{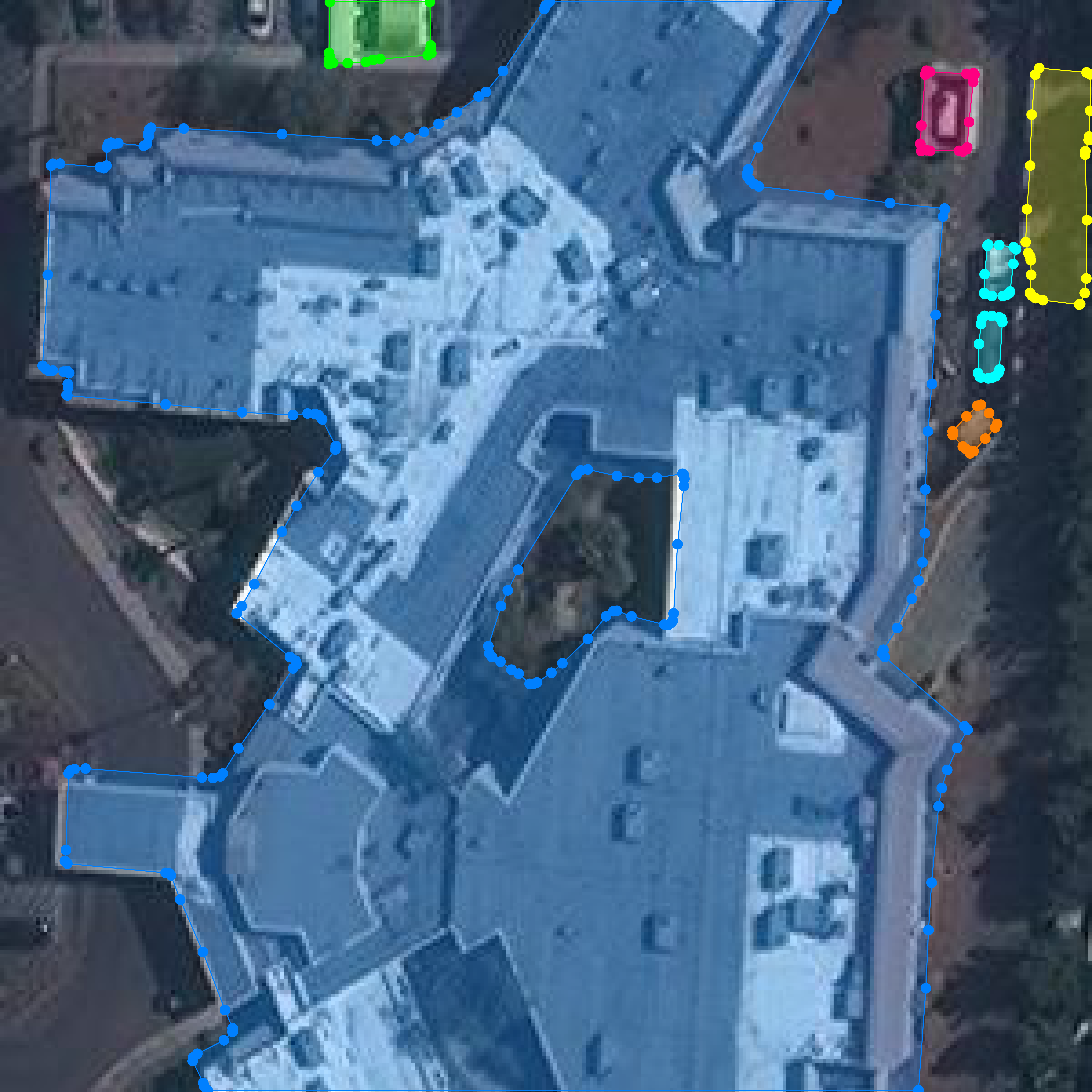}
		\caption{UResNet101 with frame field learning and simple polygonization}
		\label{fig:results_ablations_polys:frame_field_learning_simple_polygonization}
	\end{subfigure}
	\begin{subfigure}[t]{0.49\textwidth}
		\centering
		\includegraphics[trim={0 25cm 25cm 0 0},clip,width=\textwidth]{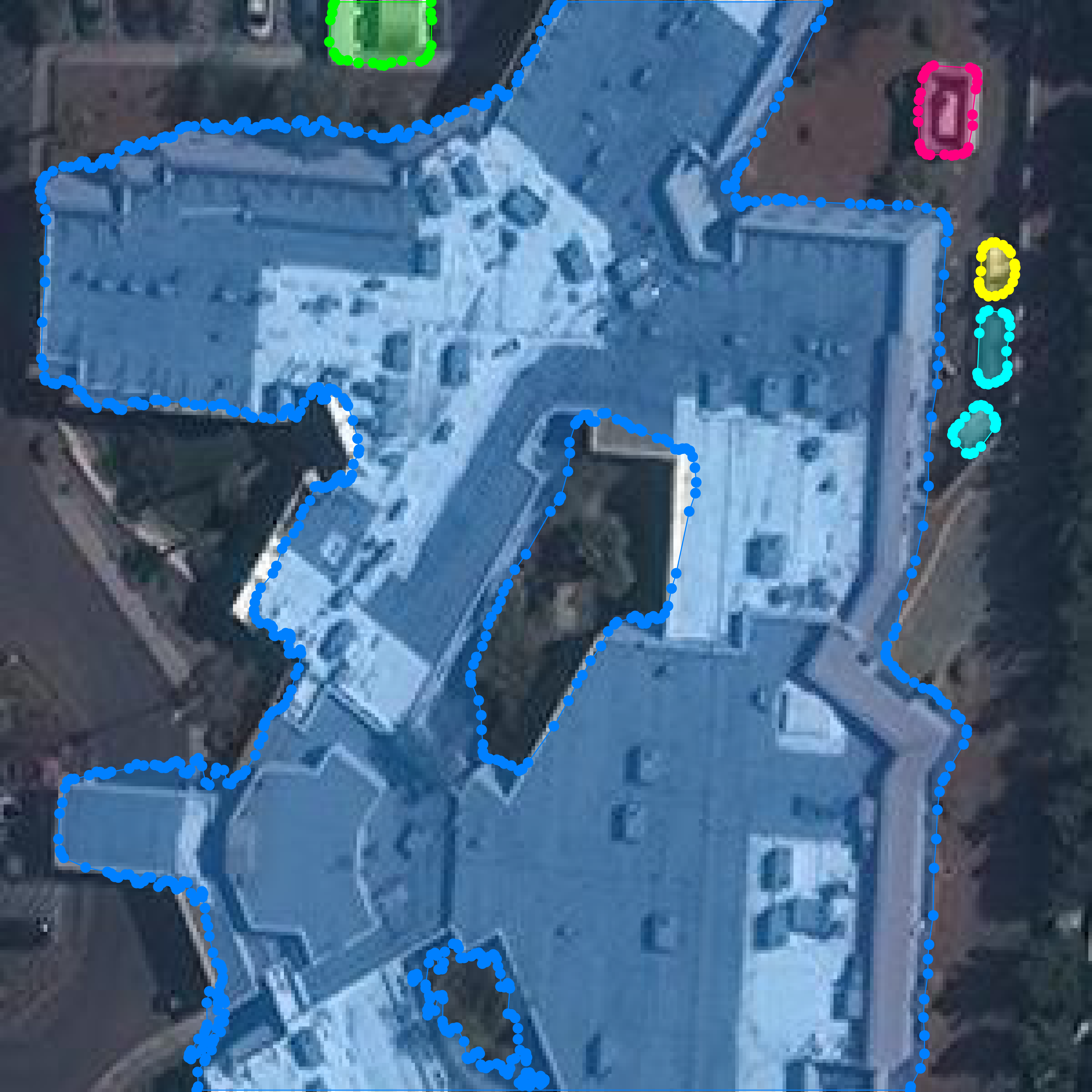}
		\caption{UResNet101 (no field) learning and simple polygonization}
		\label{fig:results_ablations_polys:simple_learning_simple_polygonization}
	\end{subfigure}
	
	\caption{Extracted polygons with: our (full) frame field learning and polygonization method; our frame field learning and simple polygonization method; the (no field) learning and simple polygonization baseline. A low tolerance of $\epsi= 0.125$~pixel was chosen to compare precise contours.}
	\label{fig:results_ablations_polys}
\end{figure*}

We also visually compare our frame field polygonization method with the simple baseline polygonization algorithm (both when the frame field is computed and when it is not) in Fig.~\ref{fig:results_ablations_polys}. The UResNet101 without frame field learning and whose results are polygonized with the simple method performs the worst (see Fig.~\ref{fig:results_ablations_polys:simple_learning_simple_polygonization}), with the UResNet101 with frame field learning and whose results are polygonized with the simple method performs already much better (see Fig.~\ref{fig:results_ablations_polys:frame_field_learning_simple_polygonization}). Our UResNet101 with frame field learning and whose results are polygonized with our frame field polygonization method performs the best, with better corners using fewer vertices (see Fig.~\ref{fig:results_ablations_polys:frame_field_learning_polygonization}). We can see our method provides the missing information needed to resolve ambiguous cases for polygonization and outputs more regular polygons.

Finally Table~\ref{tab:max_tangent_angle_error_all_results}~and~\ref{tab:coco_metrics_all_results} also hold results for additional experiments of the ablation study which each remove a loss during training. We observe that removing one of those losses does not impact the AP or AR result of the final polygonization. However, if one of those loss is removed we observe a performance drop in terms of \emph{max tangent angle errors}, with result polygons for all such experiments having a mean error slightly higher than PolyMapper (at 33.1\degree) while our full method achieves a mean error of 31.6\degree.

\subsubsection{Additional runtimes}

    \begin{table}
	\centering
	\begin{tabular}{ r | c | c }
		\hline
		\textbf{Method} & Time (sec) $\downarrow$ & Hardware\\ 
		\hline
		PolyMapper~\protect\cite{PolyMapper} & 0.38 & GTX 1080Ti  \\
		ASIP~\protect\cite{cvpr2020li} & 0.15 & Laptop CPU \\
		Zorzi et al.~\cite{zorzi2020machinelearned} & 0.11 & GTX 1080Ti \\
		\textbf{Ours} & \textbf{0.04} & GTX 1080Ti \\
		\hline
	\end{tabular}
	\caption{Average time to extract buildings from a $300\!\times\!300$ pixel patch. \textbf{Ours} refers to UResNet101 (with field), our poly. ASIP's time does not include model inference.}
	\label{tab:running_times}
\end{table}

    	We report here the average runtimes for a $300\!\times\!300$ pixel patch of the different steps of the building polygonization pipeline of Zorzi et al.~\cite{zorzi2020machinelearned} along with corresponding GPU memory allocation (GTX 1080Ti):
\begin{enumerate}
    \item segmentation: 0.152s with 20\% GPU memory,
    \item regularization: 0.269s with 12\% GPU memory,
    \item mask2poly: 0.257s with 19\% GPU memory.
\end{enumerate}
As we optimized our own method for maximum throughput, we want to compare to previous methods assuming perfect parallelization (as is done for the ASIP method in the main paper). Zorzi et al. would then get these runtimes:
\begin{enumerate}
    \item segmentation: 0.0304s,
    \item regularization: 0.03228s,
    \item mask2poly: 0,04883s,
\end{enumerate}
for a total of 0,11151s. For comparison we include the runtimes of all methods in Table~\ref{tab:running_times}, where we observe our method being competitive compared to previous works.

\subsection{Inria OSM dataset}

\begin{figure*}[!htb]
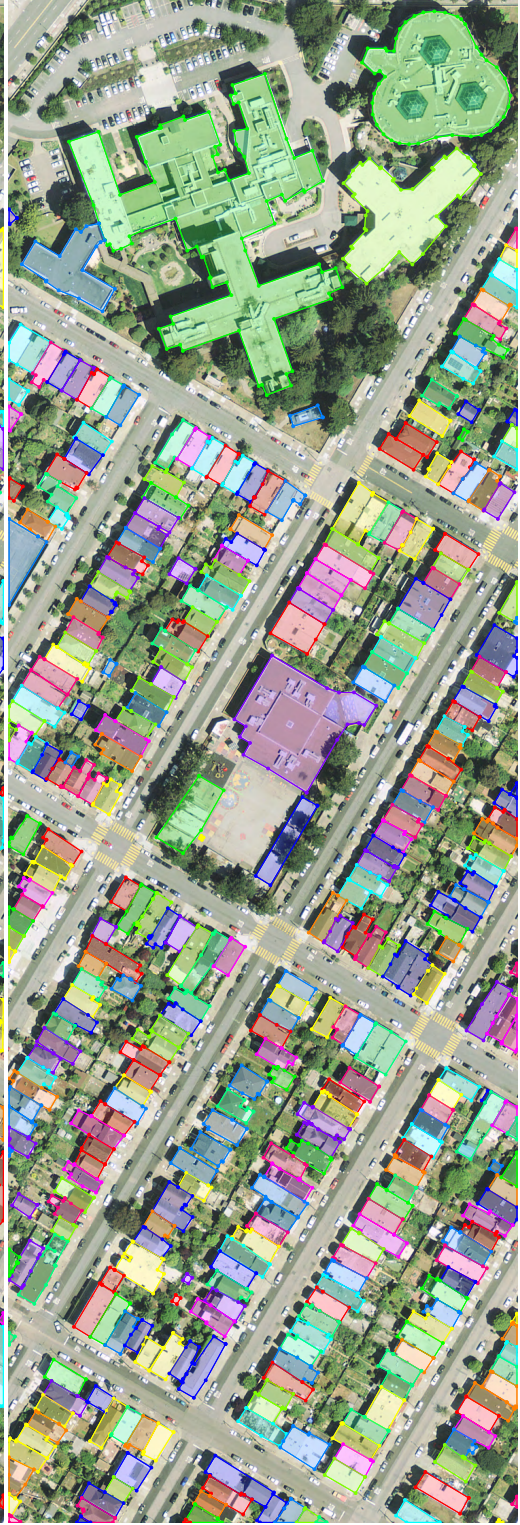

	\begin{subfigure}[t]{0.5\textwidth}
		\centering
		{U-Net16 (no field), simple poly.}
		
		\hfill
		\includegraphics[height=0.9\textheight]{{inria_sfo19.bbox_520_3575_2770_4325.poly_simple}.pdf}
	\end{subfigure}
	\begin{subfigure}[t]{0.5\textwidth}
		{\textbf{Ours}: U-Net16 (with field), our poly.}
		
		\includegraphics[height=0.9\textheight]{{inria_sfo19.bbox_520_3575_2770_4325.poly_asm}.pdf}
	\end{subfigure}	
	\caption{Crop of results on the ``sfo19'' image from the \emph{Inria OSM dataset}.}
	\label{fig:inria_results_sfo19}
\end{figure*}

\begin{figure*}[!htb]
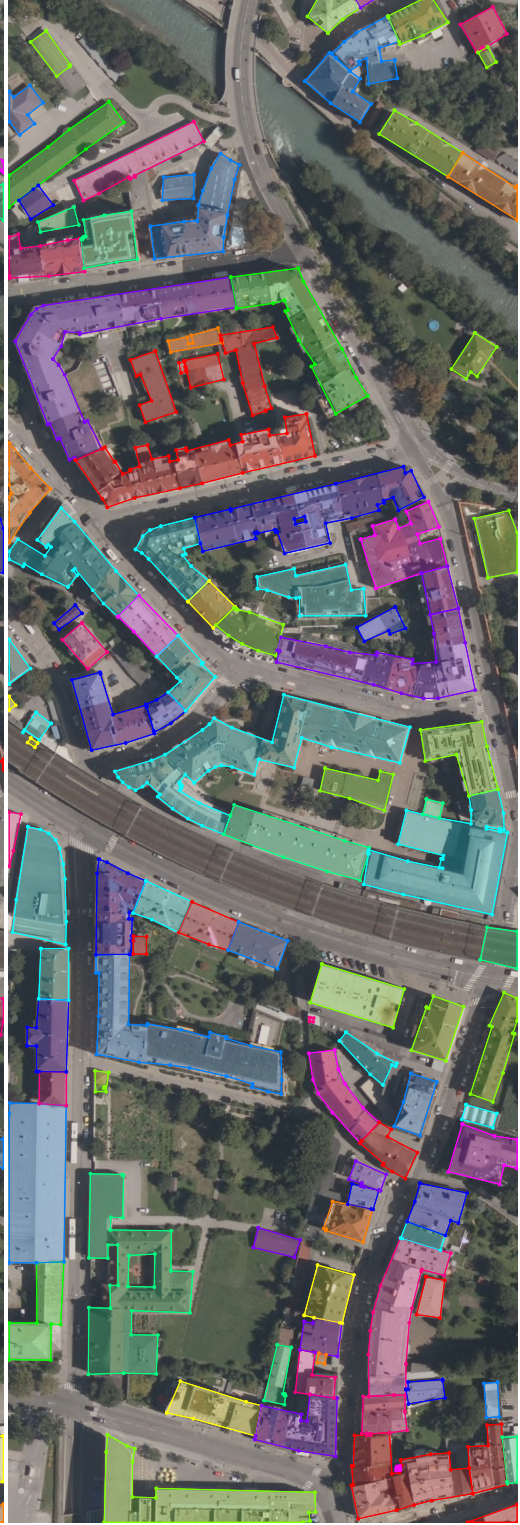

	\begin{subfigure}[t]{0.5\textwidth}
		\centering
		{U-Net16 (no field), simple poly.}
		
		\hfill
		\includegraphics[height=0.3\textheight,angle=90]{{inria_innsbruck19.bbox_2018_590_2768_2840.poly_simple}.pdf}
	\end{subfigure}
	\begin{subfigure}[t]{0.5\textwidth}
		{\textbf{Ours}: U-Net16 (with field), our poly.}
		
		\includegraphics[height=0.3\textheight,angle=90]{{inria_innsbruck19.bbox_2018_590_2768_2840.poly_asm}.pdf}
	\end{subfigure}	
	\caption{Crop of results on the ``innsbruck19'' image from the \emph{Inria OSM dataset}.}
	\label{fig:inria_results_innsbruck19}
\end{figure*}

\begin{figure*}[!htb]
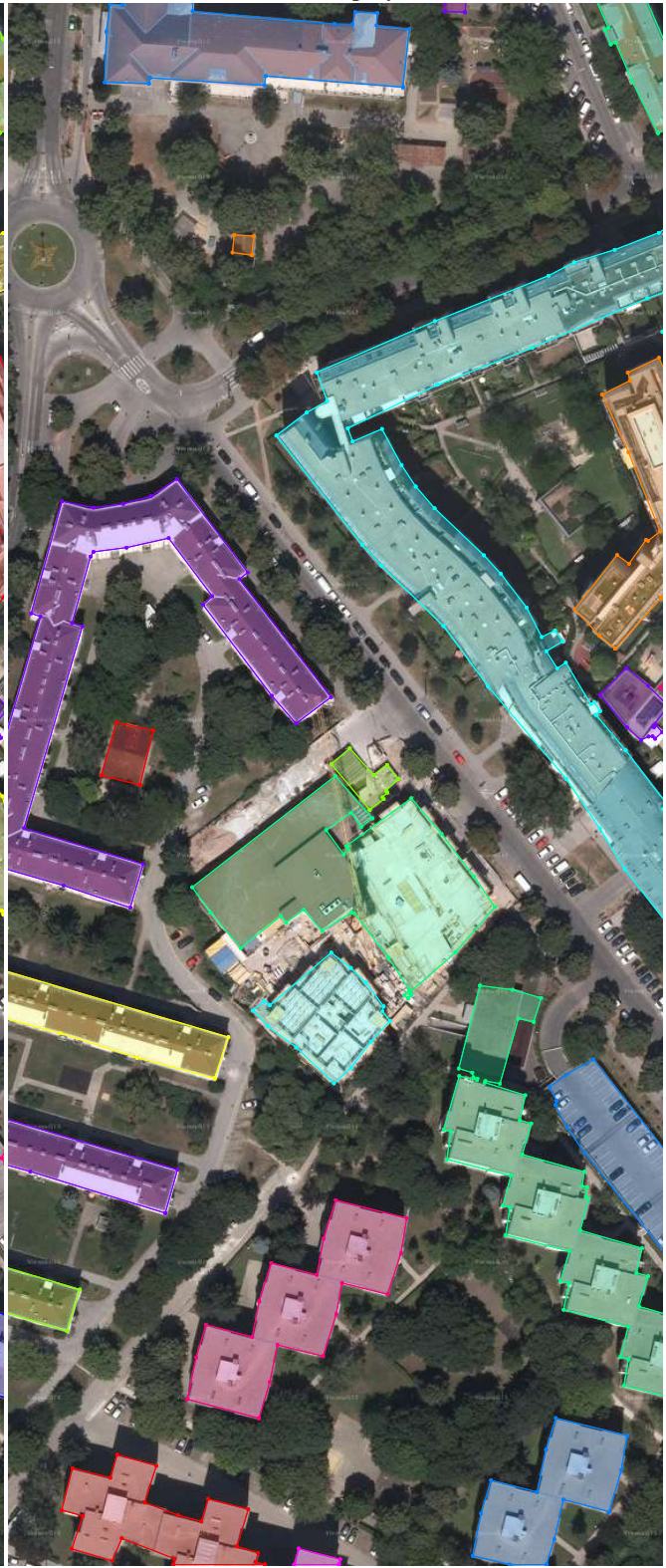

	\begin{subfigure}[t]{0.5\textwidth}
		{U-Net16 (no field), simple poly.}
		
		\includegraphics[width=\textwidth]{{inria_vienna36.bbox_400_3400_2600_5000_crop_vertical.poly_simple}.pdf}
	\end{subfigure}
	\begin{subfigure}[t]{0.5\textwidth}
		{\textbf{Ours}: U-Net16 (with field), our poly.}
		
		\includegraphics[width=\textwidth]{{inria_vienna36.bbox_400_3400_2600_5000_crop_vertical.poly_acm}.pdf}
	\end{subfigure}	
	\caption{Crop of results on the ``vienna36'' image from the \emph{Inria OSM dataset}.}
	\label{fig:inria_results_vienna36}
\end{figure*}

We show bigger crops of the result of our frame field polygonization in Fig.~\ref{fig:inria_results_sfo19},~\ref{fig:inria_results_innsbruck19},~and~\ref{fig:inria_results_vienna36}. We observe the ability of our method to separate adjoining buildings, handle complex shapes with big buildings having non-rectangular shapes and possibly holes.

\subsection{Inria polygonized dataset}

\begin{figure*}[!htb]
	\begin{subfigure}[t]{0.5\textwidth}
	    \centering
		{Eugene Khvedchenya\repeatfootnote{inria-challenge}, simple poly.}
		
		\includegraphics[height=0.45\textheight]{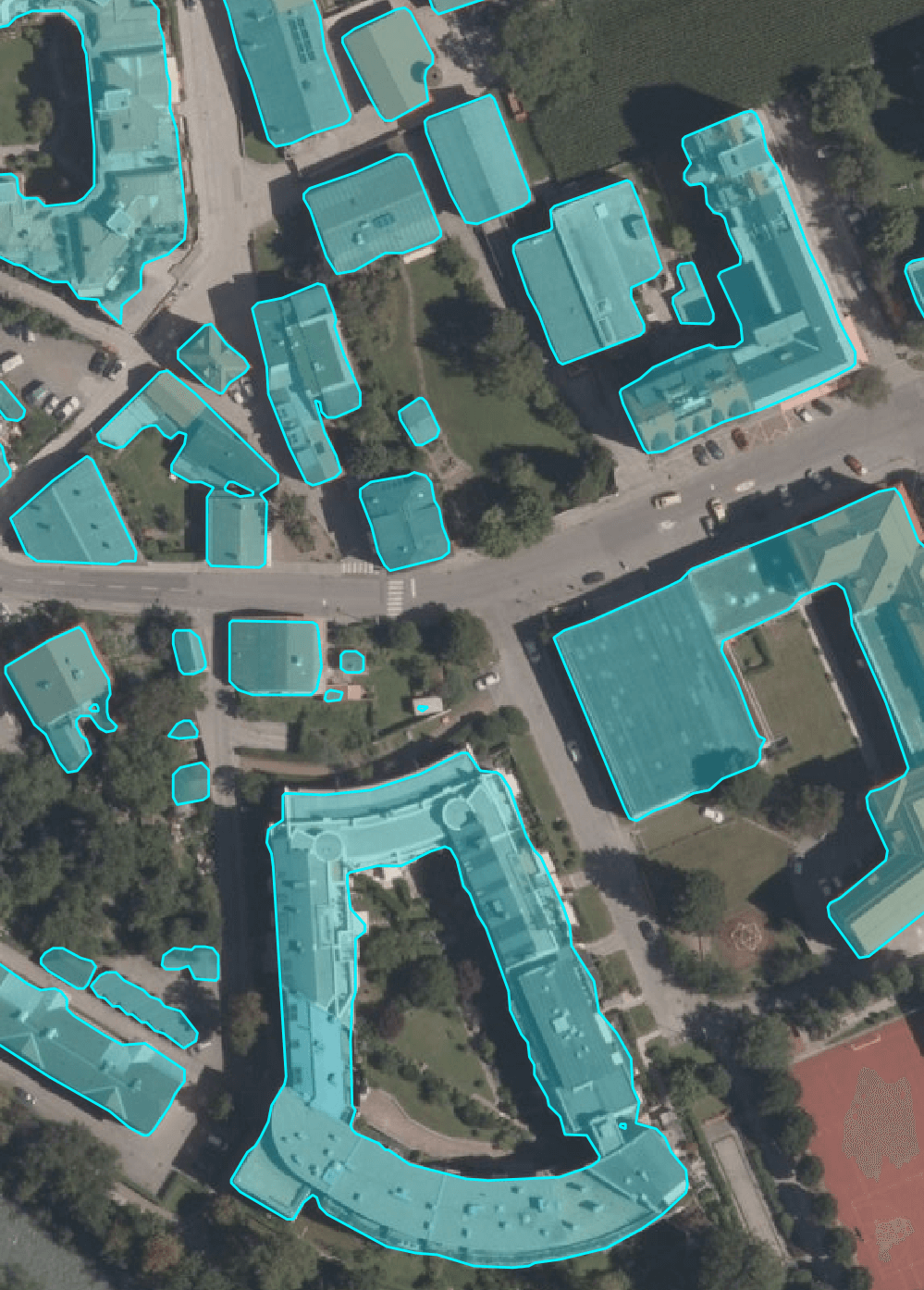}
	\end{subfigure}
	\begin{subfigure}[t]{0.5\textwidth}
	    \centering
		{ICTNet~\cite{chatterjee2019building}, simple poly.}
		
		\includegraphics[height=0.45\textheight]{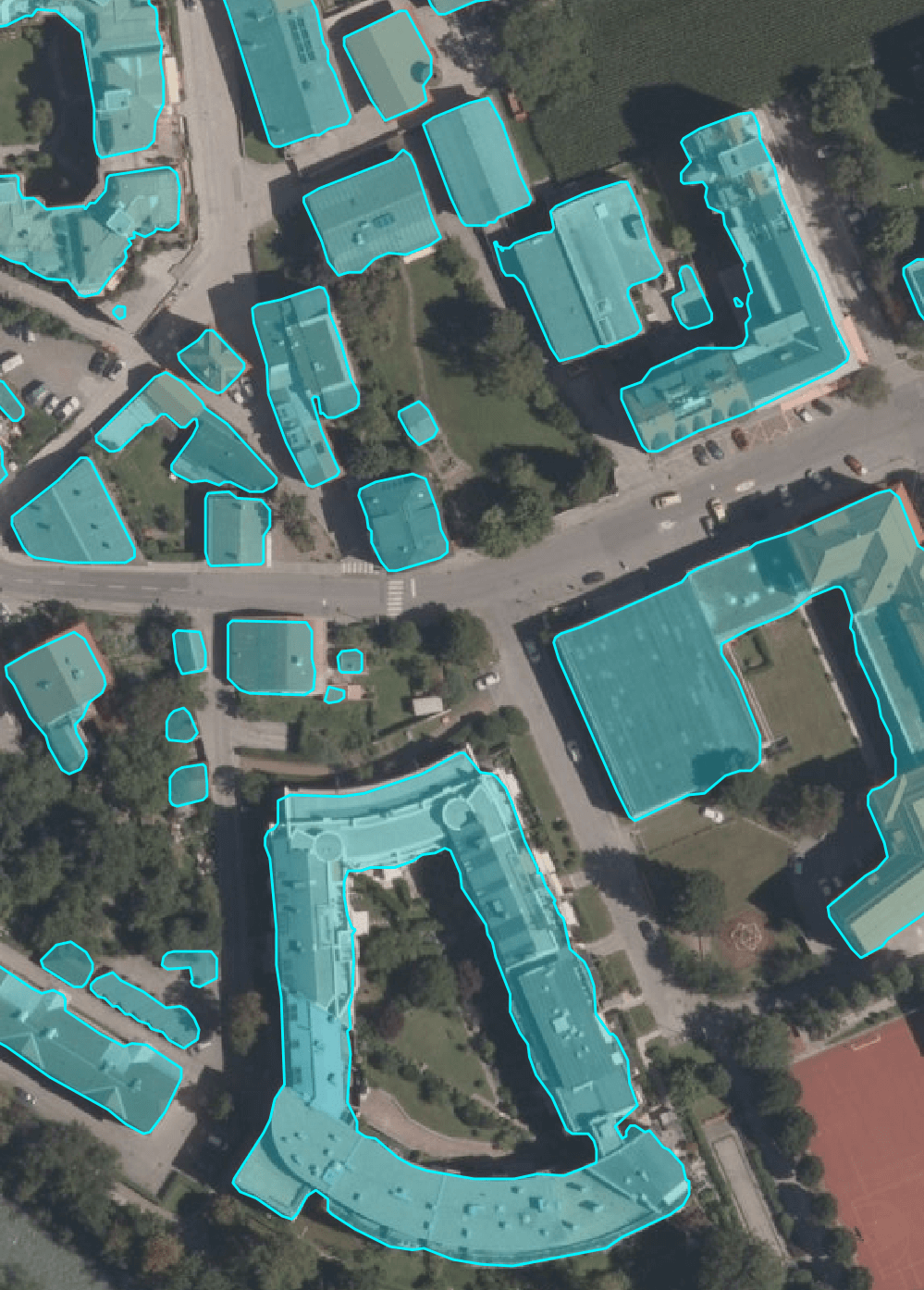}
	\end{subfigure}
	
	\begin{subfigure}[t]{0.5\textwidth}
	    \centering
		{Zorzi et al.~\cite{zorzi2020machinelearned} poly.}
		
		\includegraphics[height=0.45\textheight]{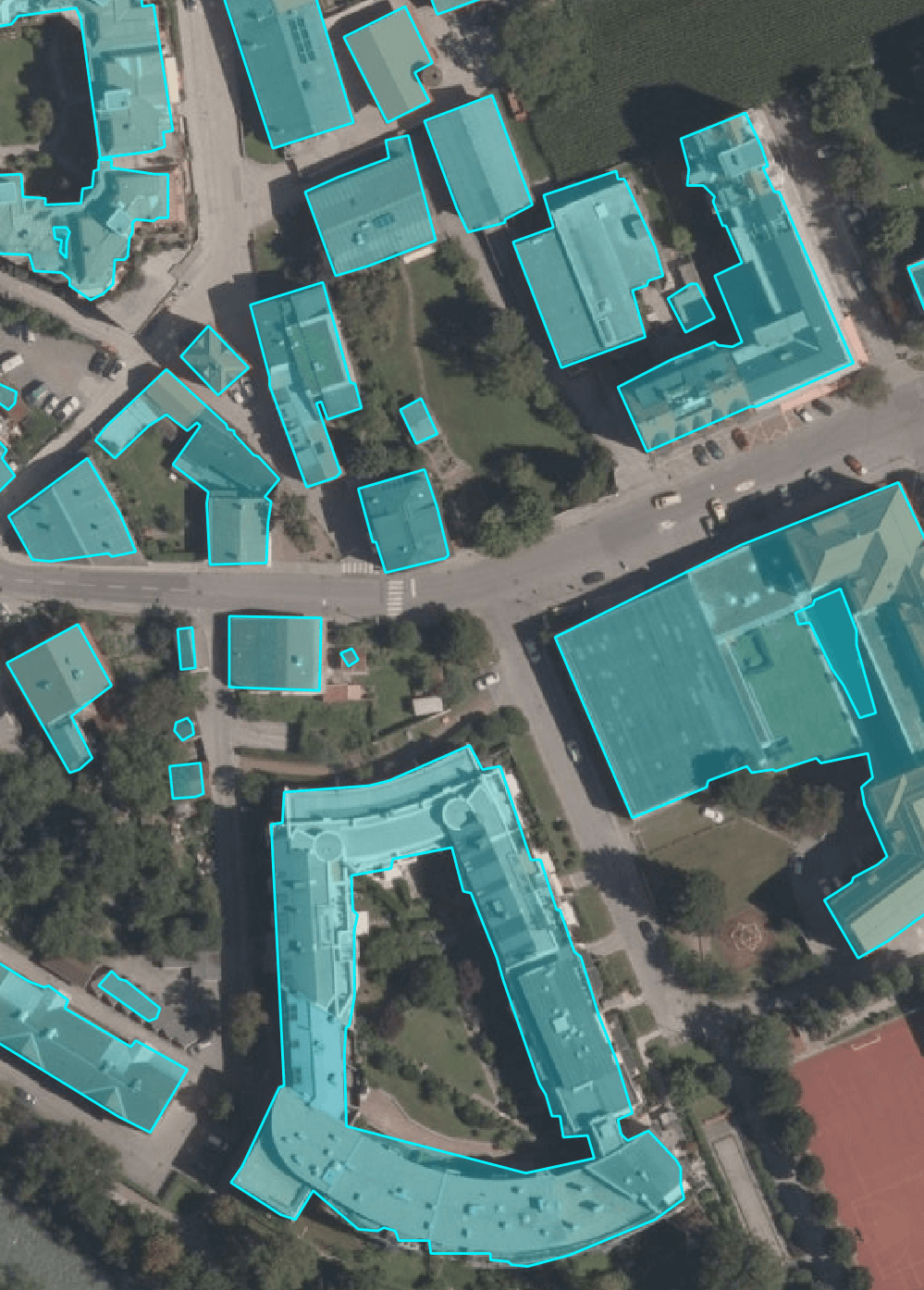}
	\end{subfigure}
	\begin{subfigure}[t]{0.5\textwidth}
	    \centering
		{\textbf{Ours}: UResNet101 (with field), our poly.}
		
		\includegraphics[height=0.45\textheight]{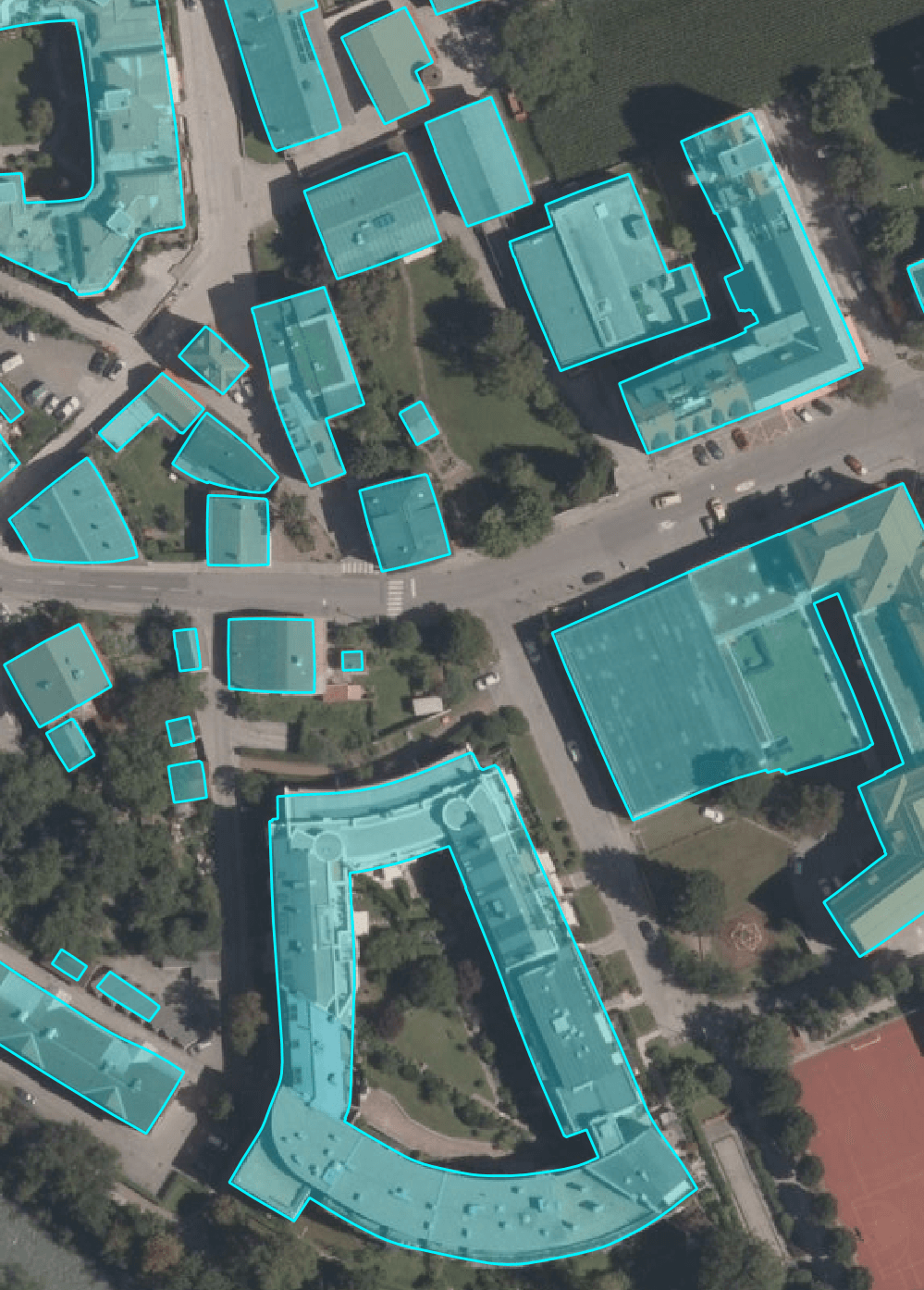}
	\end{subfigure}	
	\caption{Crop of results on an \emph{Inria Polygonized dataset} test image.}
	\label{fig:inria_polygonized_results}
\end{figure*}

We show a larger result comparison to other methods on the \emph{Inria Polygonized dataset} in Fig~\ref{fig:inria_polygonized_results}, including the two best methods on the public leaderboard\savefootnote{inria-challenge}{https://project.inria.fr/aerialimagelabeling/leaderboard/}. While the result from ``Eugene Khvedchenya'' and ICTNet acheive an mIoU over 80\%, they detect buildings with segmentation masks that need polygonization. We thus used the simple polygonization method which follows the boundaries in the segmentation raster image. Their results have blob-like features, with rounded corners and non-regular contours.

In order to compare to the ASIP polygonization method, we started to run the ASIP algorithm on the 180 output probability maps of our network, corresponding to the 180 test images. However the ASIP method is not well-suited for such big images ($5000\!\times\!5000$ pixels) with thousands of buildings, requiring a very high number of iterations (that we set to 10000). The runtime of ASIP varies greatly depending on the building density of images. For the most dense ones, it did not finish within a day of computation, making it impractical to run on the whole test dataset. As such we compare to the ASIP method only on the \emph{CrowdAI dataset}.

\subsection{Private dataset}

\begin{figure*}[!htb]
	\begin{subfigure}[t]{0.5\textwidth}
		{U-Net16 (no field)}
		
		\includegraphics[width=\textwidth]{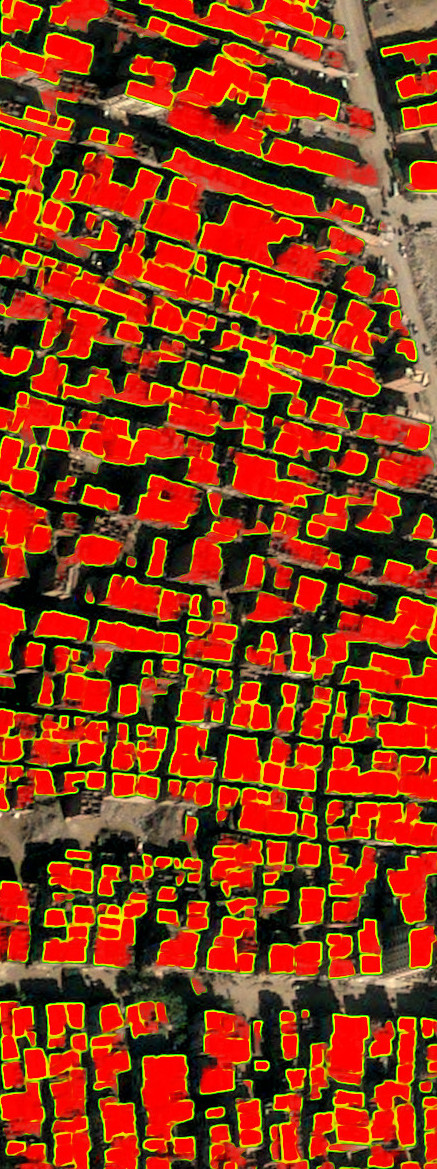}
	\end{subfigure}
	\begin{subfigure}[t]{0.5\textwidth}
		{\textbf{Ours}: U-Net16 (with field)}
		
		\includegraphics[width=\textwidth]{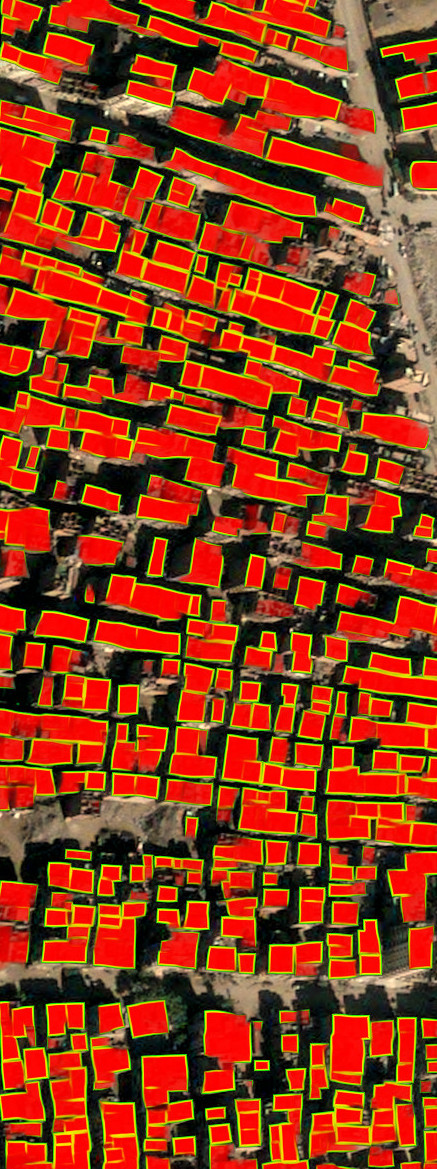}
	\end{subfigure}	
	\caption{Crop results on the ``Egypt'' test image of the private dataset.}
	\label{fig:seg_private_results_Egypt}
\end{figure*}

\begin{figure*}[!htb]
	\begin{subfigure}[t]{0.5\textwidth}
		{U-Net16 (no field)}
		
		\includegraphics[width=\textwidth]{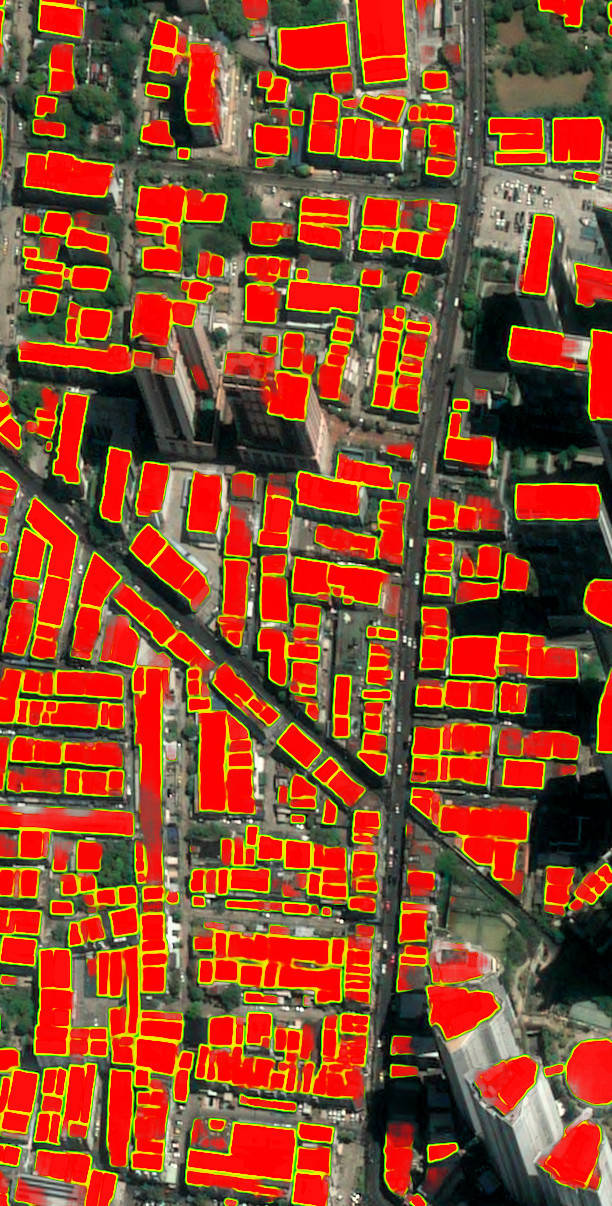}
	\end{subfigure}
	\begin{subfigure}[t]{0.5\textwidth}
		{\textbf{Ours}: U-Net16 (with field)}
		
		\includegraphics[width=\textwidth]{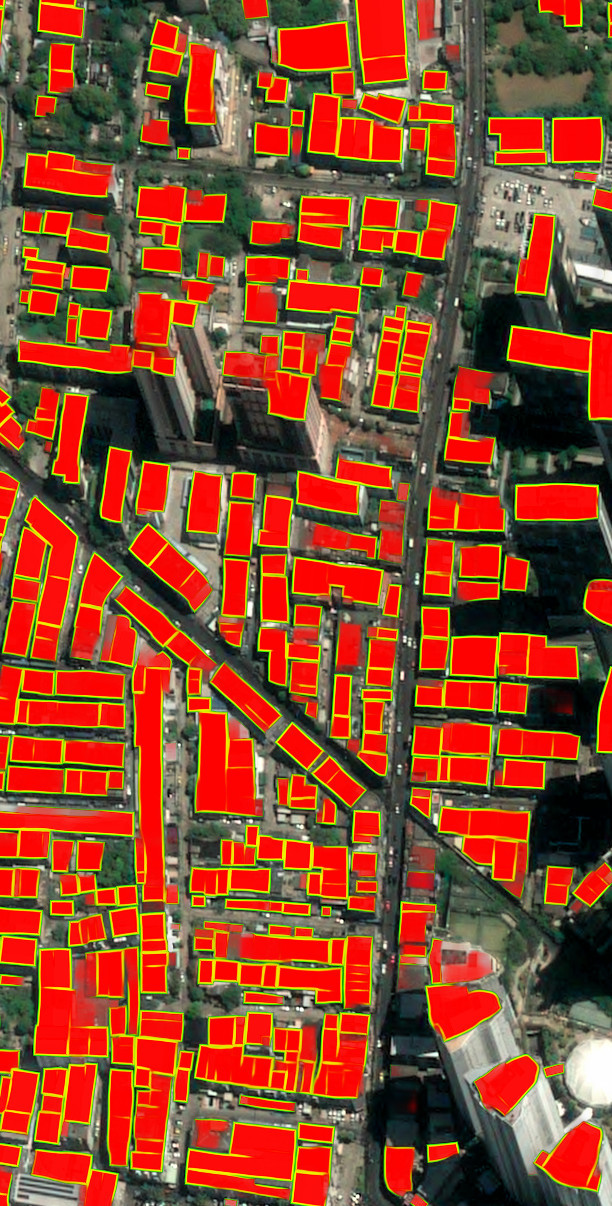}
	\end{subfigure}	
	\caption{Crop results on the ``Bangkok'' test image of the private dataset.}
	\label{fig:seg_private_results_Bangkok}
\end{figure*}

\begin{figure*}[!htb]
	\begin{subfigure}[t]{0.5\textwidth}
		{U-Net16 (no field)}
		
		\includegraphics[width=\textwidth]{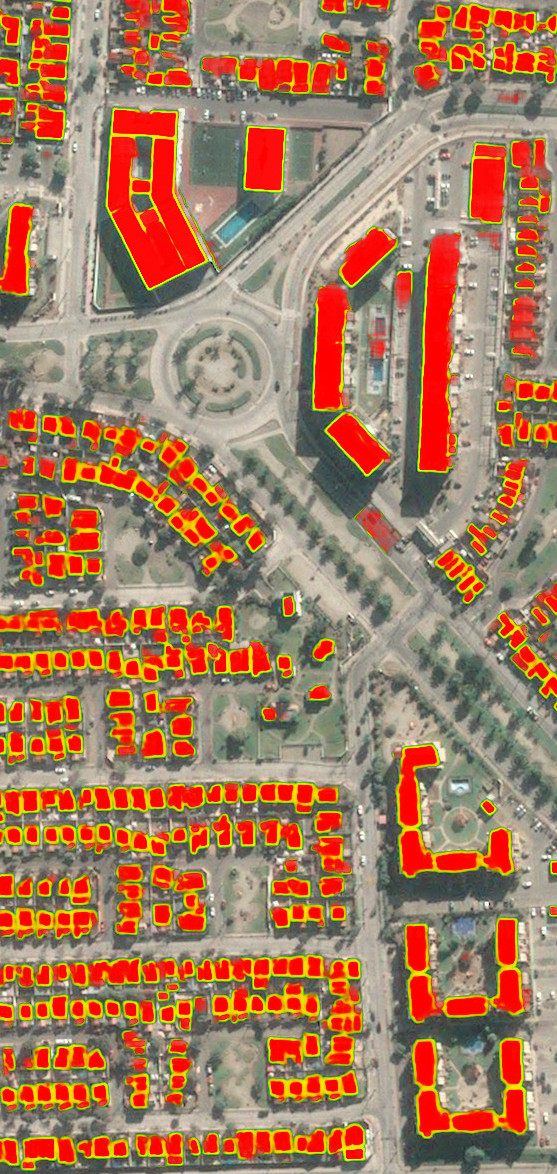}
	\end{subfigure}
	\begin{subfigure}[t]{0.5\textwidth}
		{\textbf{Ours}: U-Net16 (with field)}
		
		\includegraphics[width=\textwidth]{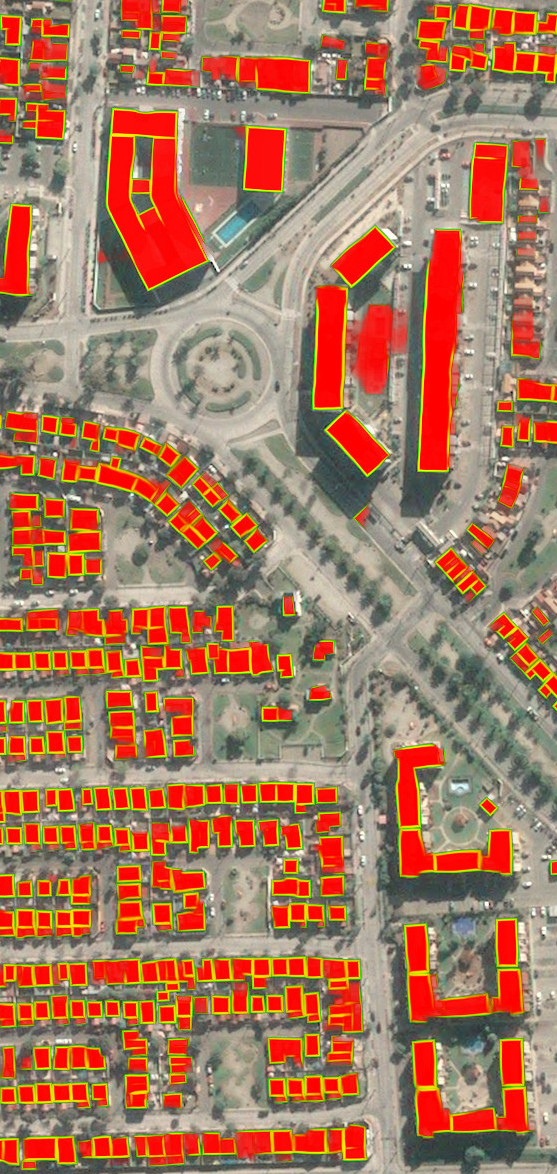}
	\end{subfigure}	
	\caption{Crop results on the ``Chile'' test image of the private dataset.}
	\label{fig:seg_private_results_Chile}
\end{figure*}

Because training on the \emph{private dataset} must be done on a restricted computer with limited access, we only train two models: U-Net16 (full) and U-Net16 (no field) until validation loss converges (around 1500 epochs). First we display segmentation raster outputs in Fig~\ref{fig:seg_private_results_Egypt},~\ref{fig:seg_private_results_Bangkok}~and~\ref{fig:seg_private_results_Chile} and final polygonal buildings in Fig~\ref{fig:private_results_Egypt},~\ref{fig:private_results_Bangkok}~and~\ref{fig:private_results_Chile}. Satellite images being more challenging than aerial images, non-regularized segmentations (no field) appear even more rounded than usual. However, our frame field learning and polygonization (with field) still outputs clean, regular geometry, and separates adjoining buildings.

\begin{figure*}[!htb]
	\begin{subfigure}[t]{0.5\textwidth}
		{U-Net16 (no field), simple poly.}
		
		\includegraphics[width=\textwidth]{{private_unet16_field_off_Egypt.poly_viz.simple.tol_1.crop}.pdf}
	\end{subfigure}
	\begin{subfigure}[t]{0.5\textwidth}
		{\textbf{Ours}: U-Net16 (with field), our poly.}
		
		\includegraphics[width=\textwidth]{{private_unet16_Egypt.poly_viz.asm.tol_1.crop}.pdf}
	\end{subfigure}	
	\caption{Crop results on the ``Egypt'' test image of the \emph{private dataset}.}
	\label{fig:private_results_Egypt}
\end{figure*}

\begin{figure*}[!htb]
	\begin{subfigure}[t]{0.5\textwidth}
		{U-Net16 (no field), simple poly.}
		
		\includegraphics[width=\textwidth]{{private_unet16_field_off_Bangkok.poly_viz.simple.tol_1.crop}.pdf}
	\end{subfigure}
	\begin{subfigure}[t]{0.5\textwidth}
		{\textbf{Ours}: U-Net16 (with field), our poly.}
		
		\includegraphics[width=\textwidth]{{private_unet16_Bangkok.poly_viz.asm.tol_1.crop}.pdf}
	\end{subfigure}	
	\caption{Crop results on the ``Bangkok'' test image of the \emph{private dataset}.}
	\label{fig:private_results_Bangkok}
\end{figure*}

\begin{figure*}[!htb]
	\begin{subfigure}[t]{0.5\textwidth}
		{U-Net16 (no field), simple poly.}
		
		\includegraphics[width=\textwidth]{{private_unet16_field_off_Chile.poly_viz.simple.tol_1.crop}.pdf}
	\end{subfigure}
	\begin{subfigure}[t]{0.5\textwidth}
		{\textbf{Ours}: U-Net16 (with field), our poly.}
		
		\includegraphics[width=\textwidth]{{private_unet16_Chile.poly_viz.asm.tol_1.crop}.pdf}
	\end{subfigure}	
	\caption{Crop results on the ``Chile'' test image of the \emph{private dataset}.}
	\label{fig:private_results_Chile}
\end{figure*}